\documentclass[logoright,numbers]{mhug}

\usepackage{array}
\usepackage{colortbl}
\usepackage{algorithm}
\usepackage{algpseudocode}
\usepackage{listings}
\usepackage{float}
\usepackage{afterpage}
\usepackage{multicol}
\usepackage{environ}
\usepackage{wrapfig}
\usepackage{etoolbox}
\makeatletter\patchcmd{\maketitle}{\vspace{12pt}}{\vspace{3pt}}{}{\PackageWarning{main}{title spacing patch failed}}\makeatother
\makeatletter\newcommand{\merosappendix}{\par\addvspace{26pt plus 40pt}\appendix\noindent{\mhug@titlefont\color{mhugink}Appendix\par}\vspace{4pt}}\makeatother

\newcommand{\meros}{MEROS\xspace}
\makeatletter
\newcommand{\merosnewthm}[2]{%
  \newcounter{#1}%
  \newenvironment{#1}[1][]{%
    \par\addvspace{2pt plus 1pt}\refstepcounter{#1}%
    \noindent{\bfseries #2~\csname the#1\endcsname}%
    \if\relax\detokenize{##1}\relax\else\ (##1)\fi{\bfseries.}\hspace{.5em}\itshape\ignorespaces}%
   {\par\addvspace{2pt plus 1pt}}}
\makeatother
\merosnewthm{theorem}{Theorem}
\merosnewthm{proposition}{Proposition}
\merosnewthm{lemma}{Lemma}
\merosnewthm{corollary}{Corollary}

\newcommand{\E}{\mathbb{E}}
\newcommand{\Loc}{\mathcal{L}}
\newcommand{\Glob}{\mathcal{G}}
\newcommand{\Far}{\mathcal{F}}
\newcommand{\Samp}{\mathcal{J}}
\newcommand{\neff}{n_{\mathrm{eff}}}
\newcommand{\pex}{\pi_{\mathrm{ex}}}
\newcommand{\piF}{\pi_{\Far}}
\newcommand{\bq}{\mathbf{q}}
\newcommand{\bk}{\mathbf{k}}
\newcommand{\bv}{\mathbf{v}}
\newcommand{\bo}{\mathbf{o}}
\newcommand{\bA}{\mathbf{A}}
\newcommand{\bU}{\mathbf{U}}
\newcommand{\bff}{\mathbf{f}}
\newcommand{\ba}{\mathbf{a}}
\newcommand{\bT}{\mathbf{T}}
\definecolor{merosexact}{rgb}{0.21,0.49,0.74}\newcommand{\cex}[1]{{\color{merosexact}#1}}
\definecolor{merosfar}{rgb}{0.80,0.38,0.11}\newcommand{\cfar}[1]{{\color{merosfar}#1}}
\definecolor{merosglob}{rgb}{0.00,0.47,0.37}\newcommand{\cglob}[1]{{\color{merosglob}#1}}
\newcommand{\mainref}[1]{\cref{#1}}

\newcommand{\tabsetup}[1]{\fontsize{#1}{1.18\dimexpr #1pt\relax}\selectfont\setlength{\tabcolsep}{2.6pt}%
  \setlength{\aboverulesep}{0pt}\setlength{\belowrulesep}{0pt}\renewcommand{\arraystretch}{1.13}}
\newcommand{\hcmid}{\noalign{{\color{xhead}\hrule height\cmidrulewidth}\vskip-\cmidrulewidth}}
\newcommand{\rov}{\dimexpr\tabcolsep+0.4pt\relax}
\newlength{\tabw}
\definecolor{xhead}{HTML}{E9EDF3}\definecolor{xlabel}{HTML}{F3F5F8}\definecolor{xsel}{HTML}{EAF2FA}
\definecolor{xsum}{HTML}{EBF5EC}\definecolor{xapx}{HTML}{F3EEF8}\definecolor{xmeros}{HTML}{FCE9E3}
\definecolor{xerr}{HTML}{2F80CC}\definecolor{xspd}{HTML}{1F9D63}
\definecolor{dsuav}{HTML}{F1F3F5}\definecolor{dscity}{HTML}{FFFFFF}\definecolor{dsglobe}{HTML}{F1F3F5}\definecolor{dsurur}{HTML}{FFFFFF}
\definecolor{tabhead}{HTML}{E8ECF2}\definecolor{tabmeros}{HTML}{FBE6E1}\definecolor{tabsel}{HTML}{EAF2F7}
\definecolor{tabsum}{HTML}{EFF5EA}\definecolor{tabapx}{HTML}{F2F0F5}
\newlength{\mcolw}

\newsavebox{\mtabbox}
\newlength{\mtgtw}\newlength{\mtgth}
\newcommand{\mplacebox}[1]{\setlength{\mtgth}{\dimexpr\numexpr\dimexpr\ht\mtabbox+\dp\mtabbox\relax*\dimexpr#1\relax/\wd\mtabbox\relax sp\relax}%
  \ifdim\mtgth>0.88\textheight\resizebox*{!}{0.88\textheight}{\usebox{\mtabbox}}\else\resizebox{#1}{!}{\usebox{\mtabbox}}\fi}
\newcommand{\mtarget}{\mscale\textwidth}
\newenvironment{colwrap}{\begin{lrbox}{\mtabbox}\begin{minipage}{\mcolw}\centering}%
  {\end{minipage}\end{lrbox}\mplacebox{\mtarget}\par}
\NewEnviron{fullwrap}{\sbox{\mtabbox}{\BODY}\setlength{\mtgtw}{\wd\mtabbox}%
  \addtolength{\tabcolsep}{1pt}\sbox{\mtabbox}{\BODY}\setlength{\mtgth}{\dimexpr\wd\mtabbox-\mtgtw\relax}\addtolength{\tabcolsep}{-1pt}%
  \addtolength{\tabcolsep}{\dimexpr(\textwidth*1000/961-\mtgtw)*65536/\number\mtgth\relax}%
  \sbox{\mtabbox}{\BODY}\resizebox{\textwidth}{!}{\usebox{\mtabbox}}\par}
\newenvironment{widewrap}{\begin{lrbox}{\mtabbox}}{\end{lrbox}\ifdim\wd\mtabbox>\textwidth\mplacebox{\textwidth}\else\ifdim\dimexpr\ht\mtabbox+\dp\mtabbox\relax>0.88\textheight\resizebox*{!}{0.88\textheight}{\usebox{\mtabbox}}\else\usebox{\mtabbox}\fi\fi}

\algrenewcommand\algorithmiccomment[1]{\hfill{\color{gray}$\triangleright$ #1}}
\newcommand{\algphase}[1]{{\color{black!60}\textit{#1}}}
\makeatletter\providecommand{\theHALG@line}{}\renewcommand{\theHALG@line}{\thealgorithm.\arabic{ALG@line}}\makeatother
\lstdefinestyle{merospy}{language=Python,basicstyle=\ttfamily\fontsize{7.6}{9.4}\selectfont,
  keywordstyle=\color{merosexact}\bfseries,commentstyle=\color{merosglob},stringstyle=\color{merosfar!90!black},
  emph={sample,meros_attention,sdpa},emphstyle=\bfseries,
  showstringspaces=false,columns=fullflexible,keepspaces=true,
  numbers=left,numberstyle=\fontsize{5.5}{9.4}\selectfont\color{black!40},numbersep=5pt,xleftmargin=11pt,framexleftmargin=11pt,
  frame=lines,framerule=0.5pt,framesep=4pt,aboveskip=4pt,belowskip=2pt,breaklines=true}

\crefname{lstlisting}{Listing}{Listings}

\makeatletter\g@addto@macro\normalsize{\abovedisplayskip=8pt plus 2pt minus 1pt\belowdisplayskip=8pt plus 2pt minus 1pt\abovedisplayshortskip=8pt plus 2pt minus 1pt\belowdisplayshortskip=8pt plus 2pt minus 1pt}\makeatother\normalsize
\AtBeginDocument{\setlength{\parskip}{5pt plus 2pt minus 1pt}}
\makeatletter
\titleformat{\section}{\mhug@display\fontsize{12.6}{15}\selectfont}{\textcolor{mhuglink}{\thesection}}{0.9em}{\color{mhugink}}
\titleformat{\subsection}{\mhug@display\fontsize{10.5}{13}\selectfont}{\textcolor{mhuglink}{\thesubsection}}{0.9em}{\color{mhugink}}
\titleformat{\subsubsection}{\mhug@sans\bfseries\fontsize{9.5}{12}\selectfont}{\textcolor{mhuglink}{\thesubsubsection}}{0.9em}{\color{mhugink}}
\makeatother
\titlespacing*{\section}{0pt}{16pt plus 12pt minus 4pt}{8pt plus 3pt minus 1pt}
\titlespacing*{\subsection}{0pt}{12pt plus 9pt minus 4pt}{5pt plus 2pt minus 1pt}

\renewcommand{\topfraction}{0.92}\renewcommand{\textfraction}{0.06}\renewcommand{\floatpagefraction}{0.85}

\title{MEROS: The More a ViT Sees, the Less Precise It Needs to Attend}

\author[1,3,2]{Hao Yu}
\author[2]{Feng Xue}
\author[1,3]{Hui Wei}
\author[1,3]{Guoying Zhao}

\affiliation[1]{ELLIS Institute Finland}
\affiliation[2]{MHUG, University of Trento}
\affiliation[3]{CMVS, University of Oulu}

\abstract{%
Vision foundation models rely on global self-attention for dense representations, whose quartic cost dominates the compute of high-resolution perception. Existing training-free acceleration methods either select tokens or average token groups, both introducing systematic attention bias. We find that high-resolution attention combines a sharp local peak with a far field that is \textbf{diffuse yet dominant}:
holding most attention mass despite spreading it thinly across distant tokens. For this diffuse far field,
inverse-rate-weighted random sampling yields an unbiased estimate that rivals even the best same-size key selection.
We further prove that its error bound halves as the image side doubles. Building on this finding,
we introduce \meros, a training-free operator for replacing self-attention in pretrained models.
It computes each query's local tile exactly and samples the far field at stratified positions.
Since sampling is input-independent, all queries in a tile share the same far-field sample in one dense pass.
At 2048\,px, each patch query reads $7.2\times$ fewer keys.
Over 17 segmentation settings on seven foundation models,
\meros loses a median of only 0.17 mIoU while speeding up the backbone by over $3.8\times$.
At operator level, \meros runs $4.0\times$ (1024\,px) to $24.8\times$ (8192\,px) faster than well-optimized FlashAttention-2.\parfillskip=0pt plus .26\linewidth
}

\date{\today}
\correspondence{\email{hao.2.yu@oulu.fi}}
\metadata[Project page]{\url{https://cv-ac.github.io/MEROS/}}
\reporttype{Technical Report}

\addlogo[width=6em]{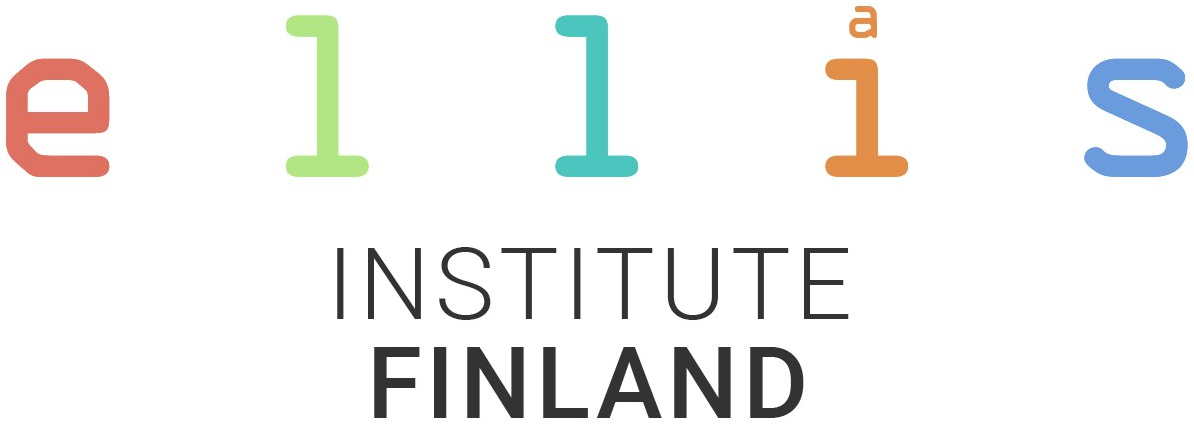}
\addmhuglogo[6em]
\addlogo[width=6em]{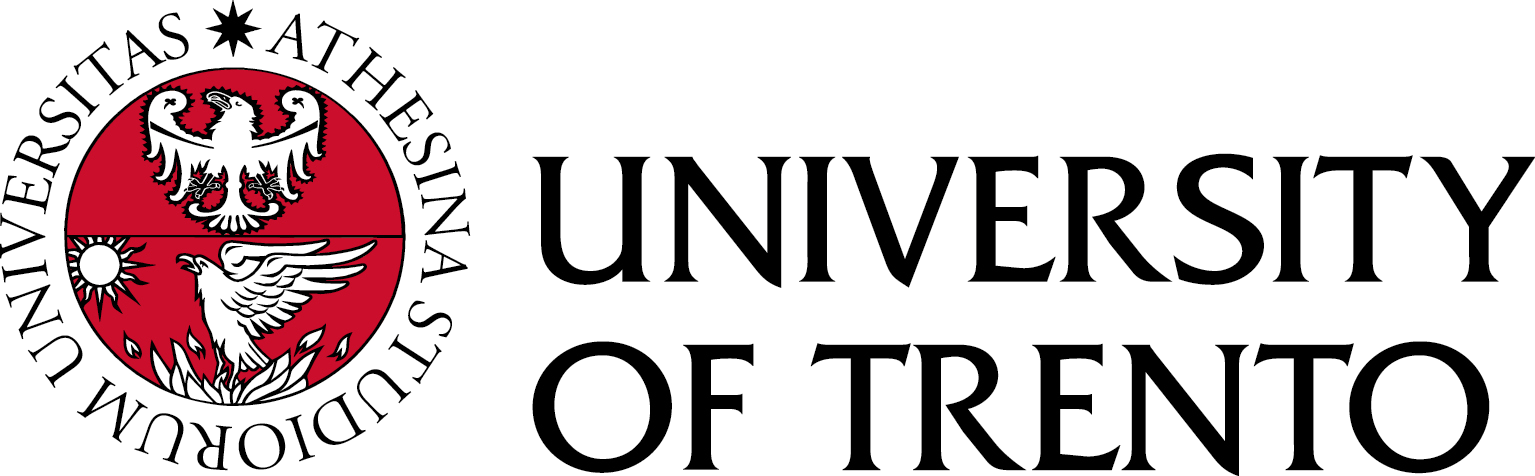}
\addlogo[width=6em]{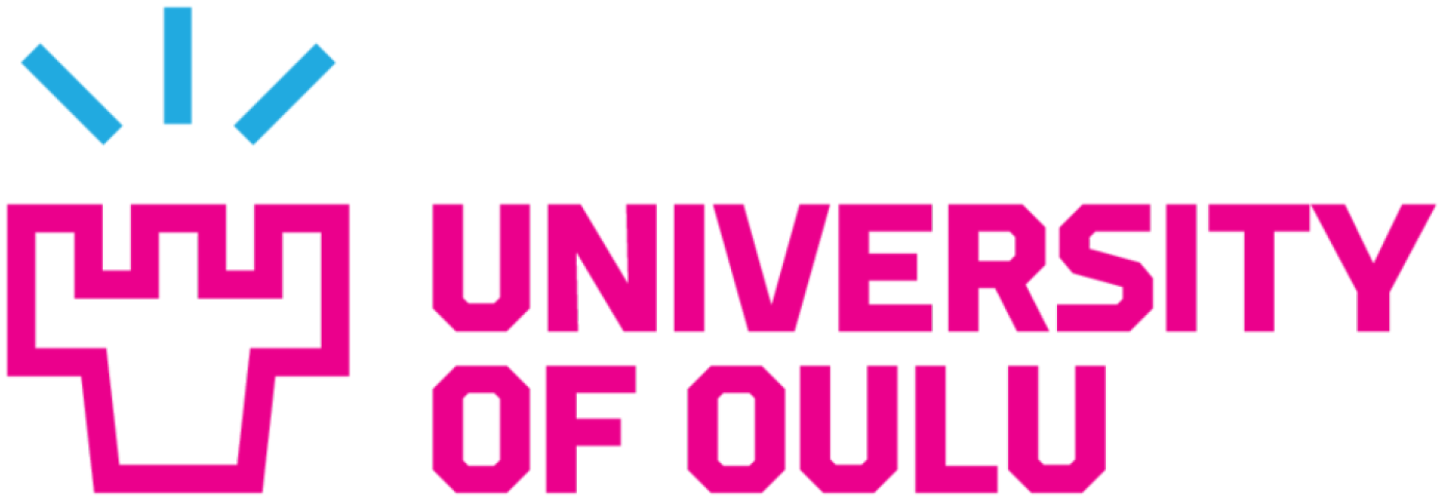}

\newlist{plist}{itemize}{1}\setlist[plist]{label=\textbullet,leftmargin=1.25em,itemsep=2.5pt plus 1pt minus 1pt,topsep=3pt plus 1pt minus 1pt,parsep=0pt}
\newcommand{\PUP}{{\looseness=-1\par}}
\newcommand{\PFILL}{{\parfillskip=0pt plus .22\linewidth\par}}

\begin{document}
\setcounter{topnumber}{2}\setcounter{bottomnumber}{0}\setcounter{totalnumber}{2}\renewcommand{\topfraction}{0.88}\renewcommand{\textfraction}{0.1}\renewcommand{\floatpagefraction}{0.8}
\maketitle
\suppressfloats[t]

\begin{takeaway}[What MEROS offers]
\begin{itemize}[leftmargin=1.2em,itemsep=2pt,topsep=2pt]
  \item \textbf{Training-free, drop-in speedup.} \meros replaces the vanilla attention layers in a pretrained ViT without retraining and keeps every token. One \meros layer runs 67$\times$ faster than vanilla attention at 1024\,px and 145$\times$ at 2048\,px. Compared with FlashAttention-2, it runs 4.0$\times$ to 24.8$\times$ faster from 1024 to 8192\,px.
\item \textbf{Unbiased attention estimation.} The far-field estimate of \meros is unbiased, and its error halves as the image side doubles. Over 17 segmentation settings, \meros loses a median of 0.07~mIoU while running the whole backbone over 3$\times$ faster, and 0.17~mIoU with a more aggressive setting that runs over 3.8$\times$ faster.
  \item \textbf{Whole-model acceleration.} With \meros, whole ViT backbones run up to 20.2$\times$ faster than with FlashAttention-2. Using \meros, four evaluated VLMs produce their first token 3.1$\times$ to 8.8$\times$ sooner.
\end{itemize}
\end{takeaway}

\afterpage{\begin{figure}[!t]
  \centering
  \includegraphics[width=\textwidth]{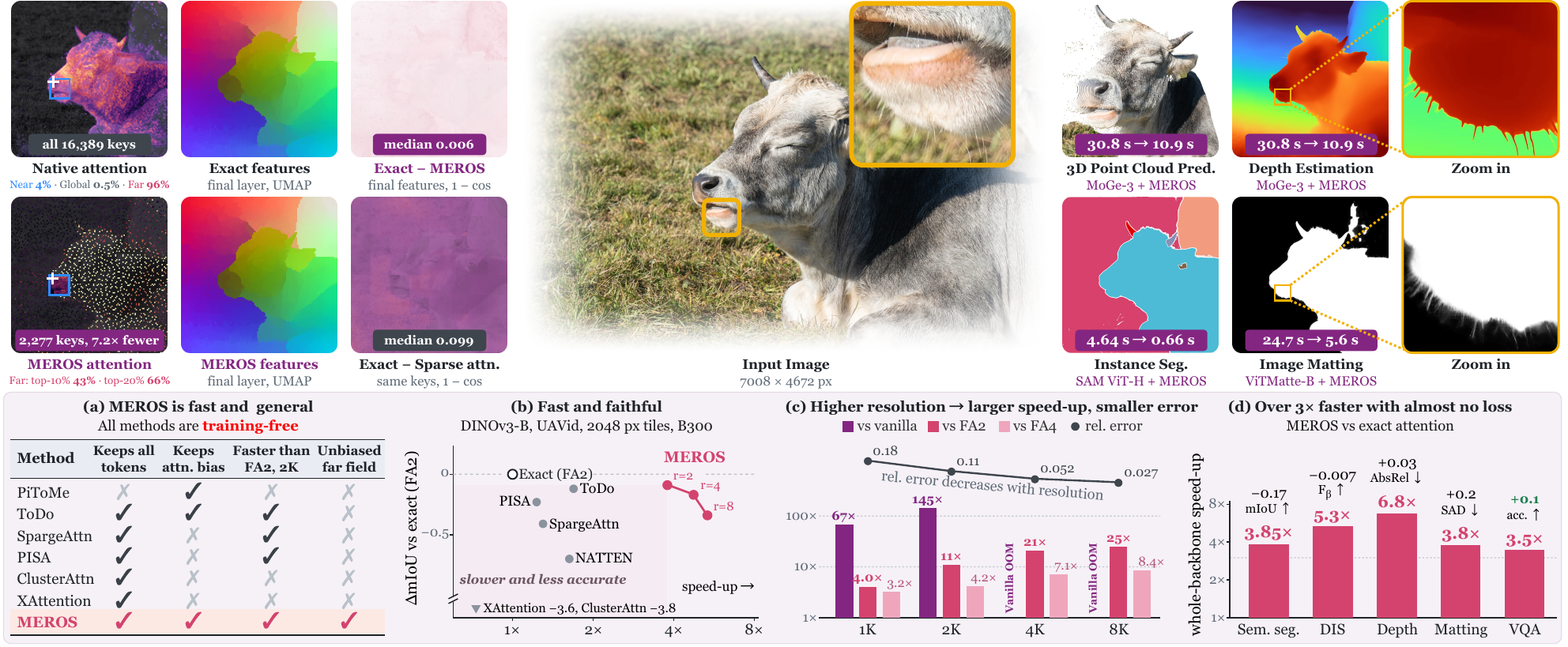}
  \caption{\textbf{MEROS for unbiased acceleration.} In the top row, the exact attention of one query at 2048\,px spans the whole image. \meros computes the local tile exactly and samples the far field at fixed positions, reading $7.2\times$ fewer keys with near-identical features, and speeds up visual perception without training. In the bottom row, (a)~\meros is fast and general. (b)~It beats all the compared training-free baselines in speed and accuracy. (c)~Its speedup over FlashAttention-2 grows and error falls with resolution. (d)~Whole backbones run over 3$\times$ faster with almost no loss in semantic and dichotomous segmentation, depth, matting and VQA.}
  \label{fig:teaser}
\end{figure}}

\section{Introduction}
\label{sec:intro}

Pretrained vision foundation models with global self-attention, such as \mbox{DINOv3}~\cite{simeoni2025dinov3} and SigLIP~2~\cite{tschannen2025siglip2}, are widely used in vision. However, preserving fine details at high resolutions~\cite{liu2024llava15,sun2025llavauhdv3} makes global self-attention expensive. Doubling the image side multiplies query--key interactions by $16\times$. FlashAttention~\cite{zadouri2026flashattention,dao2024flashattention2} speeds up execution but still computes all query--key interactions. Pretrained ViTs thus need faster attention that keeps their features without retraining.\PUP

One way to reduce this cost is to compute only a subset of query--key interactions and approximate the rest. Existing methods follow two strategies.
\textit{Selection}~\cite{hassani2026gnafindings,xu2025xattention,zhang2025sparge} retains local or important keys, and token pruning~\cite{wang2024zerotprune} drops tokens altogether. \textit{Summarization}~\cite{nordenram2026clusterattention,li2026pisa} replaces groups of keys with centroids or first-order expansions around them, and token merging~\cite{bolya2023tome,tran2024pitome} averages tokens. Pruning and merging shorten the whole sequence, so dense prediction must also reconstruct the removed positions.
Unlike local~\cite{liu2021swin} or linear attention~\cite{choromanski2021performer}, which require architectural changes, these methods are more readily applied to pretrained ViTs. However, their accuracy depends on how attention is distributed and which interactions can be approximated.\PFILL

To see how attention is distributed, we examine attention in ViTs and find a consistent pattern.
\textbf{\textit{Attention concentrates sharply around each query (near field),
yet most of its mass is distributed across distant tokens (far field).}} In a frozen DINOv3 at 2048\,px, the near field of a query holds 1.6\% of the keys and a median 6--16\% of the attention mass, while the far field holds 72--90\% at a much smaller weight per key.
This pattern explains the errors of existing approximations.
\begin{itemize}
\item For \textit{Selection}, even keeping the 12.5\% of far keys with the largest true weights drops 5--16\% of the far-field mass, so it systematically underestimates the far field.
\item For \textit{Summarization}, replacing key groups with centroids underestimates summed exponential attention weights, while finer approximations still leave residual errors.
\end{itemize}

These limitations suggest estimating the \textit{total} far-field contribution rather than compressing keys. Randomized attention approximations~\cite{chen2021scatterbrain,zandieh2023kdeformer} take this direction, but their irregular memory access makes them slow on long sequences. This paper shows that the diffuse far field can be estimated without bias by a simple operation that keeps each far-field key with probability $1/r$ and weights it by $r$. This preserves the far-field sums in expectation, and sampling each $4{\times}4$ cell independently lets their errors cancel. Because the sample is fixed in advance, the estimate needs no search, clustering or calibration, and runs as a plain dense attention pass. We further prove an output error bound proportional to $\sqrt{r/n_{\mathrm{eff}}}$, where $n_{\mathrm{eff}}$ is the effective number of cells carrying the far-field mass. For diffuse attention, $n_{\mathrm{eff}}$ grows with the token count, so the bound halves when the image side doubles. Higher resolution thus makes exact attention more expensive but far-field estimation more accurate.\PFILL

Based on this analysis, we introduce a training-free operator, \meros (\cref{fig:teaser}), from the Greek \emph{m\'{e}ros} (``part''). It computes the local tile and the global class and register tokens exactly, where attention concentrates, and samples the diffuse far field. Each $4{\times}4$ far cell keeps center-reflected key pairs, so the sample is stratified and cancels linear trends within the cell exactly. Because this sample depends only on positions, it is fixed in advance and shared by all queries of a tile. A constant $\log r$ logit offset applies the sampling weight $r$, so exact and sampled keys share one dense attention pass per tile. At 2048\,px, each patch query attends to 2,277 of 16,389 keys, reducing query--key interactions by $7.2\times$.

\enlargethispage{\baselineskip}This design needs no change to the pretrained weights. Nothing is trained, no token is removed and every key keeps its own position term, so \meros can directly replace the attention of a released ViT. It also suits the hardware. Each tile is one dense pass, which uses the GPU as fully as FlashAttention-4 (FA4)~\cite{zadouri2026flashattention} does, so fewer keys mean less time.

We evaluate \meros as a drop-in, training-free replacement in seven frozen ViTs from four families (DINOv3~\cite{simeoni2025dinov3}, DeiT III~\cite{touvron2022deit3}, MAE~\cite{he2022mae}, SigLIP 2~\cite{tschannen2025siglip2}) and in four VLMs (Qwen3-VL-4B/8B~\cite{bai2025qwen3vl}, GLM-4.6V-Flash~\cite{glm2025glm45v}, Kimi-VL-A3B~\cite{kimi2025kimivl}).
Its operator error halves as the image side doubles.
On one NVIDIA B300 GPU, one layer of \meros runs up to $24.8\times$ faster than FlashAttention-2 (FA2)~\cite{dao2024flashattention2} and up to $8.4\times$ faster than FA4, the fastest exact kernel we test. On semantic segmentation, \meros loses a median of only 0.17 mIoU while the backbone runs over 3.8$\times$ faster.

\meros also speeds up dichotomous segmentation, matting, depth and point-map estimation, and visual question answering, all with frozen pretrained backbones (\cref{fig:teaser}(d)). One implementation serves all of them, with absolute, rotary or relative position terms and patches of 14 or 16\,px. Our contributions are as follows.\PFILL

\begin{itemize}
  \item We identify a \emph{diffuse yet dominant} far field in high-resolution ViTs, for which the errors of selection and summarization persist. We prove an error bound for its unbiased estimate that halves as the image side doubles.
  \item We introduce \meros, which computes the local tile and global tokens exactly and samples the far field with known inclusion probability for unbiased estimation.
  \item We show that \meros can run as one dense attention pass per tile on NVIDIA Hopper and Blackwell GPUs.
\end{itemize}
\section{Related Work}
\label{sec:related}

\noindent\textbf{Efficient attention.} Windowed~\cite{liu2021swin,hassani2023nat}, pooled~\cite{wang2021pvt,li2022mvitv2}, focal~\cite{yang2021focal} and low-rank or kernel attention~\cite{wang2020linformer,xiong2021nystromformer,choromanski2021performer} change the operator and usually need training. \meros approximates the original softmax attention, so any pretrained ViT with global softmax attention can use it as a drop-in replacement without retraining.\PFILL

\noindent\textbf{Training-free token reduction.} Token merging and pruning~\cite{bolya2023tome,tran2024pitome,wang2024zerotprune,zeng2025tokentransforming} shorten the sequence, so dense prediction must reconstruct the removed positions~\cite{liang2022expediting} or copy them from nearest neighbors~\cite{zeng2025tokentransforming}. ToDo~\cite{smith2024todo} keeps every query but subsamples all keys and values on a fixed grid, which also thins the sharp local peak. \meros keeps every token and computes each query's tile exactly. It samples each far key with inclusion probability $1/r$ and weights it by $r$.

\noindent\textbf{Selection and summarization.} Selection keeps the key blocks predicted to matter, by block scores~\cite{zhang2025sparge} or retrieved representative tokens~\cite{xiao2024infllm}, and loses the mass of the rest. Summarization approximates this mass by cluster centroids~\cite{nordenram2026clusterattention,hooper2025multipole} or a first-order expansion around the block mean~\cite{li2026pisa}. We find that their errors, characterized at a fixed configuration in~\cite{nordenram2026clusterattention}, persist as resolution grows for a diffuse far field. \meros estimates far-field sums without bias and needs no per-query selection, clustering or calibration.\PFILL

\noindent\textbf{Sampling-based attention estimation} can be unbiased for LLM decoding. Several methods combine hashing with sampled~\cite{han2024hyperattention,chen2025magicpig} or low-rank~\cite{chen2021scatterbrain} estimates. Certified Sparse Attention~\cite{li2026certified} and vAttention~\cite{desai2026vattention} draw a fresh sample at every decoding step and locate heavy keys in a one-dimensional sequence. In dense vision, every token is a query and the far field lies on a grid. \meros reads heavy keys near each query exactly, by position, and samples the far field per cell. The sample is input-independent, so each tile runs one dense pass. Its error bound falls as the number of far cells grows with resolution. Isolated heavy patches, such as the high-norm artifacts of register-free models~\cite{darcet2024registers,lu2025artifacts}, are still the hard case for the bound of \cref{cor:resolution}.

\noindent\textbf{Content-based sampling.}
Recent video diffusion methods build input-dependent sampling plans at runtime. They estimate block probabilities by scoring pooled keys against clustered queries~\cite{xu2026abattention} or by running $k$-means at each attention call~\cite{oj2026instagen}, then reweight the sampled blocks to stay unbiased. A video model can spread this cost over many denoising steps~\cite{xu2026abattention}, whereas a perception ViT runs one forward pass per image and would pay for clustering and scoring every time. Because the probabilities depend on input scores, the variance of these estimators is also hard to analyze~\cite{oj2026instagen}. \meros takes a different approach for high-resolution visual perception, where attention peaks near each query and spreads thinly over distant tokens. Instead of searching for important blocks at runtime, it computes the local tile and global tokens exactly to anchor the softmax normalizer and samples every far key with a \textit{fixed, content-independent probability $1/r$ set by position before inference}. To reduce variance without scoring the input, \meros uses spatial symmetry instead. Inside each $4{\times}4$ far cell, center-reflected key pairs cancel linear trends and give lower error than simple random sampling for 86 to over 99\% of queries (\cref{fig:supp_theory}(c)). Because the sample does not depend on the image, it adds no runtime overhead and lets every tile run in one \textit{dense attention pass}. This fixed design also gives an exact variance and an error bound that halves whenever the image side doubles (\cref{cor:resolution}).

\section{Method}
\label{sec:method}

\meros replaces the softmax attention of a frozen ViT without changing its weights. Each patch query attends in full to its near field, the $16\times16$-patch tile that contains it, and to the few global tokens, such as the class token. The rest of the image, the \emph{far field}, often holds most of the attention but spreads it thinly. \meros reads it through a fixed random sample that keeps one key in $r$ and counts each sampled key $r$ times, so the far-field estimates are unbiased. \cref{sec:anatomy,sec:select} show why this split suits high-resolution attention, \cref{sec:analysis} bounds the error of the estimate, and \cref{sec:meros} gives the sample and the dense pass of \meros that make it fast.\PFILL

\begin{figure}[!t]
  \centering
  \includegraphics[width=\textwidth]{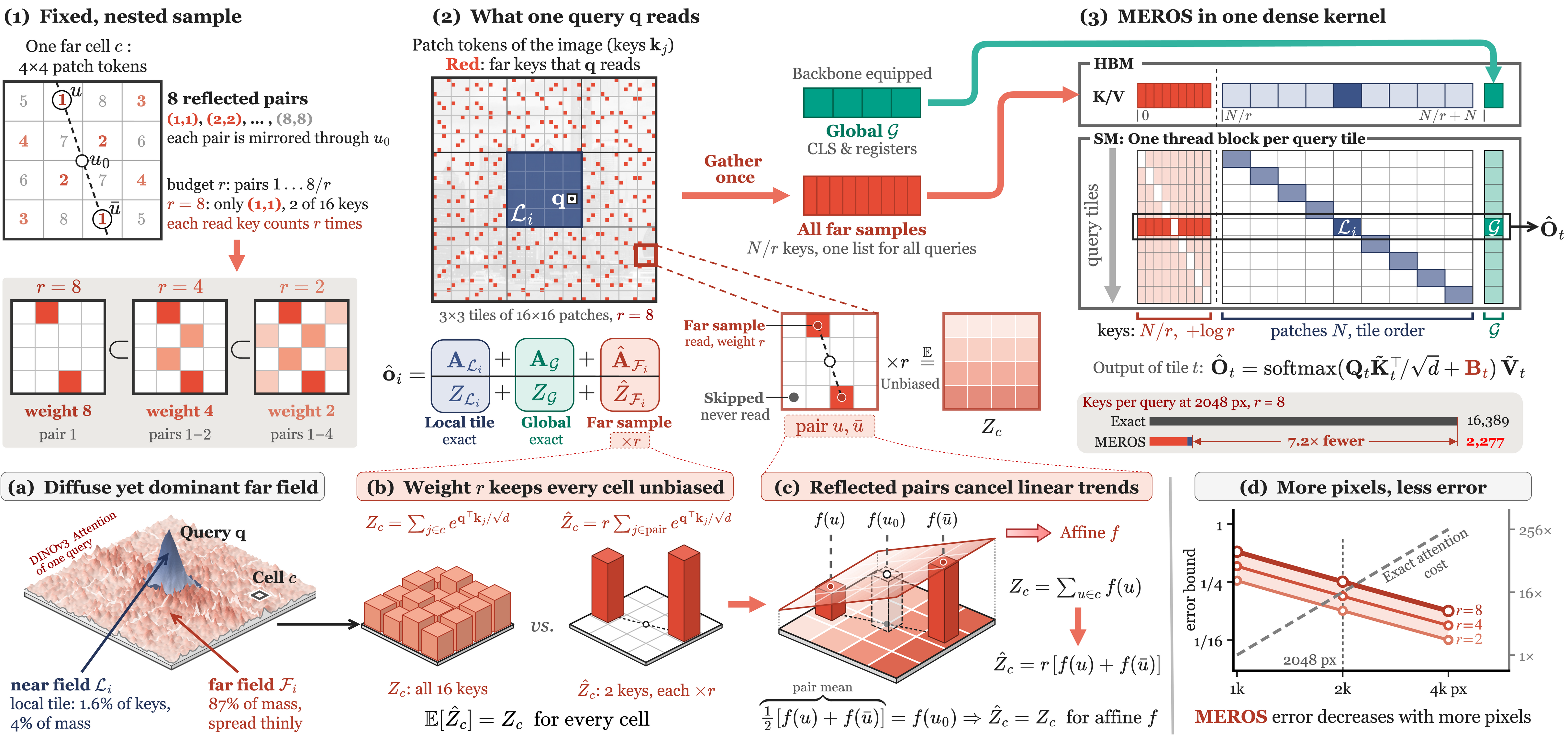}
\caption{\textbf{Overview of \meros.} In (1) and (2), each square is a key $\mathbf{k}_j$ at its patch position on the grid. \textbf{(1)}~Each $4{\times}4$ far cell keeps its first $8/r$ center-reflected pairs, so the samples of all budgets nest. \textbf{(2)}~Query $\mathbf{q}$ reads $\Loc_i$ and $\Glob$ exactly and $\Far_i$ only through samples weighted by $r$, under one normalizer (\cref{eq:meros}). \textbf{(3)}~Rows are query tiles, columns the keys they read. Each tile is one dense attention pass over its local, global and sampled far keys, and adding $\log r$ to the logit of every far key applies the weight $r$. \textbf{(a)--(d)}~The results of \cref{sec:anatomy}, \cref{lem:unbiased}, \cref{prop:pairs} and \cref{cor:resolution} ($\kappa=D=\piF=1$).}
  \label{fig:method}
\end{figure}

\subsection{Where the Attention Mass Lies}
\label{sec:prelim}\label{sec:anatomy}

Consider one attention head on an image of $N$ patch tokens on a grid and a few global tokens $\Glob$, such as the class token and registers~\cite{darcet2024registers}. Let $\bq_i,\bk_j,\bv_j$ be its queries, keys and values after the model's positional encoding, and $s_{ij}=\bq_i^\top\bk_j/\sqrt{d}$ the logit, with head dimension $d$. For a patch query $i$, the keys split into the near field $\Loc_i$ of its tile, the global tokens $\Glob$, and the far field $\Far_i$ of all other patches. With the mass $Z_X=\sum_{j\in X}e^{s_{ij}}$ and the value sum $\bA_X=\sum_{j\in X}e^{s_{ij}}\bv_j$ of a key set $X$, the exact output of query $i$ is
\begin{equation}
  \bo_i = \frac{\cex{\bA_{\Loc_i}}+\cglob{\bA_{\Glob}}+\cfar{\bA_{\Far_i}}}{\cex{Z_{\Loc_i}}+\cglob{Z_{\Glob}}+\cfar{Z_{\Far_i}}},
  \label{eq:decomp}
\end{equation}
whose denominator we call $Z$. At 2048\,px, $|\Loc_i|=256$ and $|\Far_i|=16{,}128$, so the far terms carry nearly all of the cost. All approximations below compute the \cex{local} and \cglob{global} terms exactly and differ only in the \cfar{far terms}.

To see how far mass spreads, we split the grid into $4\times4$ cells. Cells nest in tiles, so $\Far_i$ is a union of cells $c\in\mathcal{C}_i$, of which there are 1{,}008 at 2048\,px. Let $p_j=e^{s_{ij}}/Z$ be the attention weights and $P_c=Z_c/Z$ the cell shares. We drop the index $i$ of $Z$, $p_j$, $P_c$, and from \cref{thm:error} on also of $\bo$ and $\hat\bo$. The far field holds $\piF=\sum_{c\in\mathcal{C}_i}P_c$, the exact terms hold $\pex=1-\piF$, and the effective far cell number is
\begin{equation}
  \neff=\frac{\piF^2}{\sum_{c\in\mathcal{C}_i}P_c^2}\in\bigl[1,\,|\mathcal{C}_i|\bigr].
  \label{eq:neff}
\end{equation}
It is $n$ when $n$ cells share the far mass evenly and 1 when one cell holds it. We call a far field \emph{diffuse yet dominant} when both $\piF$ and $\neff$ are large. In \cref{fig:method}(a), the tile holds 1.6\% of the keys and 4\% of the mass, the global tokens 8\% and the far field 87\%. Such a far field is hard to select but easy to estimate. The pattern is not specific to this query (\cref{tab:supp_anatomy}). On five photographs at 2048\,px, the far field holds a median of 72 to 90\% of the mass in layers 6 and 11, and at least half of it for 74 to 90\% of the query--head pairs. Its median $\neff$ is 46 to 171 cells. A tile key weighs 4.3 to 13.5 times as much as a far key, but the far field has 63 times as many keys.\PFILL

\subsection{Selection and Summarization Are Biased}
\label{sec:select}

\emph{Selection} keeps a subset $\mathcal{K}\subset\Far_i$ of the far keys, such as a window or predicted blocks, and drops the rest. \emph{Summarization} replaces each cell by its mean key $\bar\bk_c$ and value $\bar\bv_c$, counted $|c|$ times as in ToMe's proportional attention~\cite{bolya2023tome}.

\begin{proposition}[Selection and summarization]
\label{prop:bias}
(i)~Selection errs by $\hat\bo_i-\bo_i=\pi_{\mathcal{D}}(\hat\bo_i-\bo_{\mathcal{D}})$, where $\bo_{\mathcal{D}}$ is the attention output over the dropped keys $\mathcal{D}=\Far_i\setminus\mathcal{K}$, of share $\pi_{\mathcal{D}}\ge\piF-|\mathcal{K}|\max_{j\in\Far_i}p_j$. (ii)~Summarization scales the mass of a cell $c$ by $e^{-\Delta_c}\le1$, with $\Delta_c=\log\frac{1}{|c|}\sum_{j\in c}e^{s_{ij}-\bar s_c}$ and $\bar s_c$ the mean logit of $c$, with equality only for equal logits.
\end{proposition}

\noindent\emph{Derivation.} (i)~Selection removes the dropped keys from both sums, so the exact output is the mixture $\bo_i=(1-\pi_{\mathcal{D}})\,\hat\bo_i+\pi_{\mathcal{D}}\,\bo_{\mathcal{D}}$, and rearranging gives the error. The kept far keys hold at most $|\mathcal{K}|\max_{j\in\Far_i}p_j$ of the mass, which bounds $\pi_{\mathcal{D}}$ from below. (ii)~The mean key of a cell has the mean logit $\bar s_c$, so its merged token enters the normalizer with the mass $|c|\,e^{\bar s_c}=Z_c\,e^{-\Delta_c}$, and Jensen's inequality gives $e^{\Delta_c}\ge1$.\PFILL

When no far key stands out, even the best $|\Far_i|/r$ keys miss close to $\piF(1-1/r)$ of the mass. Neither approximation overestimates any far cell, so their errors share one sign and add up instead of canceling. On DINOv3-B at 2048\,px, the eighth of far keys with the largest true weights still misses a median 5--16\% of the far mass and cell centroids miss 17--25\%, while the $r{=}8$ sample of \meros has a mean ratio of 1.00 and an interquartile range within $\pm$4\%.

\begin{algorithm}[!t]
\caption{\meros attention for one head.}
\label{alg:meros}
\small\renewcommand{\baselinestretch}{1.115}\selectfont
\algrenewcommand\algorithmiccomment[1]{\hfill\makebox[0.455\linewidth][l]{\color{black!55}$\triangleright$ #1}}
\begin{algorithmic}[1]
\Statex \algphase{Before inference, drawn once per layer and head}
\State $\cfar{\Samp}\gets$ first $8/r$ reflected pairs per cell \Comment{$\Pr[j\in\Samp]=1/r$, and the budgets $r=8,4,2$ nest}
\Statex \algphase{Forward pass}
\State $\cfar{\mathbf{K}_\Samp,\mathbf{V}_\Samp}\gets$ rows $\cfar{\Samp}$ of $\mathbf{K},\mathbf{V}$ \Comment{gathered once and shared by all tiles}
\For{each $16\times16$ tile $t$ \textbf{in parallel}} \Comment{its 256 queries share every loaded key}
  \State $\tilde{\mathbf{K}}_t,\tilde{\mathbf{V}}_t\gets$ keys and values of $\cfar{\Samp}$, $\cex{t}$ and $\cglob{\Glob}$ \Comment{far samples, the tile and the global tokens}
  \State $\mathbf{B}_t\gets\log r$ on $\cfar{\Samp\setminus t}$, $-\infty$ on $\Samp\cap t$, else $0$ \Comment{weight $r$, with samples inside the tile masked}
  \State $\hat{\mathbf{O}}_t\gets\mathrm{softmax}\bigl(\mathbf{Q}_t\tilde{\mathbf{K}}_t^{\top}\!/\sqrt{d}+\mathbf{B}_t\bigr)\,\tilde{\mathbf{V}}_t$ \Comment{one normalizer for exact and sampled terms}
\EndFor
\State $\cglob{\mathbf{O}_\Glob}\gets$ exact attention of the global queries \Comment{they read all $N+|\Glob|$ keys}
\end{algorithmic}
\end{algorithm}

\subsection{Unbiased Estimation by Design}
\label{sec:analysis}

Sampling removes this bias. Suppose each far key enters a random sample $\Samp$ with a known inclusion probability $1/r$, independently across cells, and is weighted by the \emph{budget} $r$:
\begin{equation}
  \cfar{\hat Z_{\Far_i}}=r\!\!\sum_{j\in\Samp\cap\Far_i}\!\!e^{s_{ij}},\qquad
  \cfar{\hat\bA_{\Far_i}}=r\!\!\sum_{j\in\Samp\cap\Far_i}\!\!e^{s_{ij}}\bv_j .
  \label{eq:farest}
\end{equation}
This is the Horvitz--Thompson estimator~\cite{horvitz1952generalization} (\cref{fig:method}(b)). Its part in a cell $c$ is $\hat Z_c=r\sum_{j\in\Samp\cap c}e^{s_{ij}}$, and $\hat\bo_i$ is \cref{eq:decomp} with these far terms. Expectations are over the draw of $\Samp$ for a fixed input. Proofs are in \cref{sec:supp_proofs}.

\begin{lemma}[Design unbiasedness]
\label{lem:unbiased}
For every input and every far cell $c$, $\E[\hat Z_c]=Z_c$. Hence $\E[\hat Z_{\Far_i}]=Z_{\Far_i}$ and $\E[\hat\bA_{\Far_i}]=\bA_{\Far_i}$ for every query $i$, and $\hat Z_{\Far_i}\ge0$ on every draw of the sample $\Samp$.\PFILL
\end{lemma}

\noindent\emph{Derivation.} Every far key has $\E\bigl[r\,\mathbb{I}[j\in\Samp]\bigr]=1$, so by linearity each weighted sum keeps its mean, for any logits and values. The estimated normalizer is at least $Z_{\Loc_i}+Z_{\Glob}>0$, because the tile and the global tokens are computed exactly.

\begin{theorem}[Error law]
\label{thm:error}
For one query, let $w_j$ be 1 for local and global keys, $r$ for sampled far keys and 0 otherwise. Then $\hat\bo-\bo=\bU/(1+\delta)$ on every draw, with the zero-mean errors $\delta=\sum_jp_j(w_j-1)$ and $\bU=\sum_jp_j(w_j-1)(\bv_j-\bo)$. If $D$ bounds $\|\bv_j-\bo\|$ over all keys, then
\begin{equation}
  \sqrt{\E\|\hat\bo-\bo\|^2}\;\le\;2\sqrt{2}\,\piF D\sqrt{r/\neff},
  \label{eq:law}
\end{equation}
and the bias obeys $\|\E[\hat\bo]-\bo\|\le8\,\piF^2D\,r/\neff$, the square of the bound in \cref{eq:law} divided by $D$.
\end{theorem}

\noindent Each far cell adds to $\bU$ an independent, zero-mean error of root-mean-square size at most $\sqrt r\,DP_c$. Unlike the one-signed errors of \cref{prop:bias}, these errors partly cancel, so the total shrinks as more cells share the far mass, whatever the number of keys. The bias is of second order, at most the square of the bound in \cref{eq:law} divided by $D$.

\noindent\emph{Derivation.} The estimate is a ratio of two unbiased sums, as in self-normalized importance sampling~\cite[Sec.~9.2]{owen2013mcbook}. Its denominator is $\sum_jw_je^{s_j}=Z(1+\delta)$, and because $\sum_jp_j(\bv_j-\bo)=\mathbf{0}$, its numerator exceeds $Z(1+\delta)\,\bo$ by
\begin{equation}
  \textstyle\sum_jw_je^{s_j}(\bv_j-\bo)=Z\sum_jp_j(w_j-1)(\bv_j-\bo)=Z\,\bU .
  \label{eq:identity}
\end{equation}
Dividing the two gives $\hat\bo-\bo=\bU/(1+\delta)$ on every draw. Only far keys have $w_j\neq1$, so $\bU=\sum_c\bU_c$ with one term $\bU_c=\sum_{j\in c}(w_j-1)\,p_j(\bv_j-\bo)$ per far cell. These terms have zero mean and are independent across cells, so their second moments add. No weight exceeds $r$ and $\|\bv_j-\bo\|\le D$, which gives $\E\|\bU_c\|^2\le r\,D^2P_c^2$ and
\begin{equation}
\textstyle\E\|\bU\|^2=\sum_c\E\|\bU_c\|^2\le r\,D^2\sum_cP_c^2=r\,D^2\,\piF^2/\neff .
  \label{eq:second}
\end{equation}
The same steps give $\E[\delta^2]\le r\,\piF^2/\neff$. On the draws with $|\delta|\le\tfrac12$, the identity gives $\|\hat\bo-\bo\|\le2\|\bU\|$. The other draws have probability at most $4\,\E[\delta^2]$ by Chebyshev's inequality, and on them $\|\hat\bo-\bo\|\le D$, because $\hat\bo$ is always a convex combination of values. Adding both parts gives $\E\|\hat\bo-\bo\|^2\le8\,r\,D^2\piF^2/\neff$, the square of \cref{eq:law}. For the bias, the first-order term $\E[\bU]$ vanishes, so $\E[\hat\bo]-\bo=-\E[\bU\delta/(1+\delta)]$. Splitting the draws as above and applying the Cauchy--Schwarz inequality bounds it by a multiple of $\sqrt{\E\|\bU\|^2\,\E[\delta^2]}\le r\,D\,\piF^2/\neff$.

\begin{corollary}[Accuracy improves with resolution]
\label{cor:resolution}
If no far cell holds more than $\kappa$ times its even share, $P_c\le\kappa\,\piF/|\mathcal{C}_i|$, then $\neff\ge|\mathcal{C}_i|/\kappa$. The root-mean-square error is at most $2\sqrt{2}\,\piF D\sqrt{\kappa r/|\mathcal{C}_i|}$, falling as $|\mathcal{C}_i|$ grows.
\end{corollary}

The corollary follows from $\sum_cP_c^2\le(\max_cP_c)\sum_cP_c\le\kappa\,\piF^2/|\mathcal{C}_i|$. With $|\mathcal{C}_i|=(N-256)/16$ and $\piF\le1$, the bound is at most $2\sqrt2\,D\sqrt{16\kappa r/(N-256)}$. For fixed $\kappa$ and $D$, it roughly halves whenever the image side doubles, while exact attention costs $N^2$ (\cref{fig:method}(d)), so higher resolutions allow a larger $r$ at equal error. Far mass concentrated in a few cells, such as on the high-norm artifacts of register-free models~\cite{darcet2024registers}, raises $\kappa$ and loosens the bound of \cref{cor:resolution}.\PUP

\begin{figure}[t]
\centering
\includegraphics[width=\textwidth]{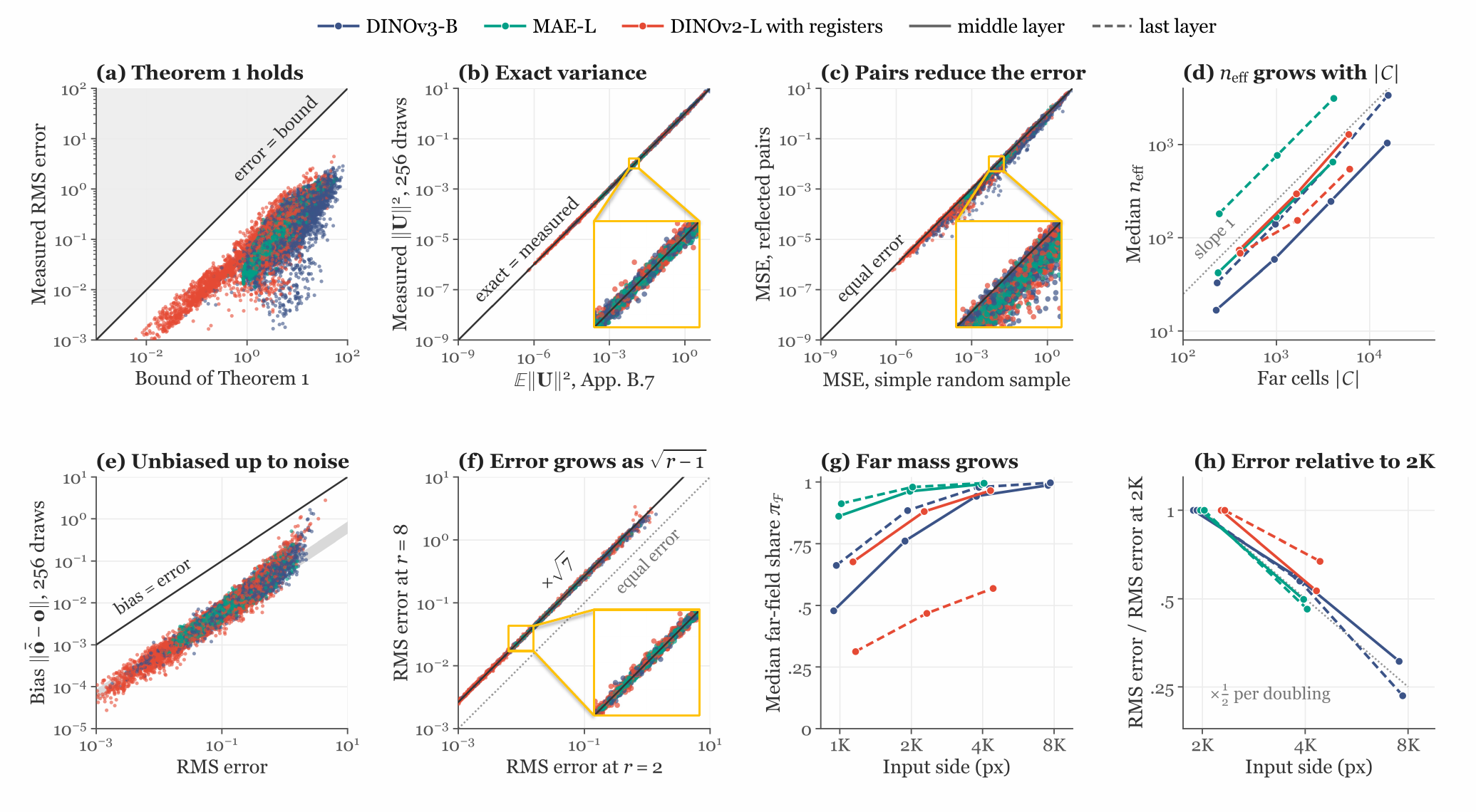}\par\vspace{-10pt}
\caption{\textbf{Measured error of \meros on pretrained ViTs, against the theoretical analysis.} We run \meros on the attention of \mbox{DINOv3-B}, MAE-L and DINOv2-L and compare what we measure with what \cref{sec:analysis} predicts. Each point is one query in one head, whose far-field sample we draw 256 times at $r{=}8$ to measure the error of the output against exact attention. Every model contributes 7{,}000 such points, taken from two layers, four images and all input sizes. Lines are medians per layer. (a)~Measured root-mean-square (RMS) error against the bound of \cref{thm:error}. No point enters the shaded region above the bound. (b)~Measured mean of $\|\bU\|^2$ against its exact value from \cref{sec:supp_variance}, along the diagonal. (c)~Mean squared error of the reflected pairs against a simple random sample with as many keys per cell. Points below the diagonal favor the pairs. (d)~Effective against actual number of far cells, which grow together along the dotted line of slope 1. (e)~Distance of the mean estimate from the exact output, with a gray band of Monte Carlo noise of about $1/\sqrt{256}$ of the error. (f)~Error of each query at $r{=}8$ against that at $r{=}2$, along the line of slope $\sqrt7$ that \cref{sec:supp_variance} predicts. (g)~Far-field share of the attention mass against the input side. (h)~RMS error relative to its value at 2K, with lines as geometric means, close to the dotted line that halves it per doubling. Yellow boxes magnify 0.4 to 0.6 decades per axis.}
\label{fig:supp_theory}
\vspace{-1.0em}
\end{figure}

\noindent\textbf{Check on real attention.} Both properties hold on pretrained models as the analysis states. The estimate shows no measurable bias, and its error shrinks with resolution at the predicted rate. \Cref{fig:supp_theory} tests them on the attention of \mbox{DINOv3-B}, MAE-L and DINOv2-L in 20 settings of model, layer and input size (\cref{sec:supp_check}). The measured error stays below the bound of \cref{thm:error} for every query and head, and their ratio changes by less than a factor of 1.4 across input sizes. The mean of 256 draws lies within Monte Carlo noise of the exact output, so the bias is below what these draws resolve. Its distance from the exact output is 0.056 to 0.065 times the root-mean-square error in the median, where noise alone gives about $1/\sqrt{256}=0.0625$. For 86 to over 99\% of the queries, the reflected pairs of \cref{sec:meros} give a lower error than a simple random sample. The exact variance of \cref{sec:supp_variance} predicts the measured mean squared error to within a median of 2.5\%, and from $r{=}2$ to $r{=}8$ the error grows by a median factor of 2.64, against the $\sqrt7$ of the analysis. The median $\neff$ grows nearly in proportion to the number of far cells, as \cref{cor:resolution} requires. From 2048\,px up, each doubling of the image side multiplies the error by 0.53 on average, close to the $\frac12$ of \cref{cor:resolution}.

\subsection{MEROS for Unbiased Attention Acceleration}
\label{sec:meros}

\noindent\textbf{Exact where the mass concentrates.} The tile of a query holds most of its attention peak, and global tokens often act as attention sinks~\cite{xiao2024streamingllm,lu2025artifacts}. Sampling them would put a few large shares into $\sum_cP_c^2$ and inflate the error, while computing them exactly costs only $256+|\Glob|$ keys per query, about 1.6\% of the keys at 2048\,px.\PFILL

\begin{table*}[t]
\centering
\tabsetup{7.6}
\renewcommand{\arraystretch}{0.98}
\setlength{\tabcolsep}{1.5pt}
\caption{\textbf{Training-free attention acceleration methods on \mbox{DINOv3-B}}~\cite{simeoni2025dinov3}, each in its official variant. Errors vs.\ FP32 exact attention on DIV8K~\cite{gu2019div8k} images. Backbones timed at batch~1 on a random image. Mem.\ is peak at 8K, Cos.\ mean cosine of final patch features to exact attention. Cos.\ and mIoU on 70 UAVid~\cite{lyu2020uavid} and 500 Cityscapes~\cite{cordts2016cityscapes} images. Bold is best apart from exact.}
\begin{widewrap}\begin{tabular}{l*{12}{>{\centering\arraybackslash}p{22.11pt}}>{\centering\arraybackslash}p{26.91pt}*{4}{>{\centering\arraybackslash}p{22.11pt}}}
\toprule
\rowcolor{xhead}[\tabcolsep][\rov] & \multicolumn{8}{c}{\bfseries\boldmath Attention operator, one layer} & \multicolumn{9}{c}{\bfseries\boldmath Full backbone, end to end}\\
\hcmid\cmidrule(lr){2-9}\cmidrule(lr){10-18}
\rowcolor{xhead}[\tabcolsep][\rov] & \multicolumn{4}{c}{\bfseries\boldmath Rel.\ error $\downarrow$} & \multicolumn{4}{c}{\bfseries\boldmath Speedup over FA2 $\uparrow$} & \multicolumn{4}{c}{\bfseries\boldmath Speedup over FA2 $\uparrow$} & \bfseries\boldmath Mem.\,$\downarrow$ & \multicolumn{2}{c}{\bfseries\boldmath Cos.\ $\uparrow$} & \multicolumn{2}{c}{\bfseries\boldmath mIoU $\uparrow$}\\
\hcmid\cmidrule(lr){2-5}\cmidrule(lr){6-9}\cmidrule(lr){10-13}\cmidrule(lr){15-16}\cmidrule(lr){17-18}
\rowcolor{xhead}[\tabcolsep][\rov] \bfseries\boldmath Method & \bfseries\boldmath 1K & \bfseries\boldmath 2K & \bfseries\boldmath 4K & \bfseries\boldmath 8K & \bfseries\boldmath 1K & \bfseries\boldmath 2K & \bfseries\boldmath 4K & \bfseries\boldmath 8K & \bfseries\boldmath 1K & \bfseries\boldmath 2K & \bfseries\boldmath 4K & \bfseries\boldmath 8K & \bfseries\boldmath GB & \bfseries\boldmath UAVid & \bfseries\boldmath City & \bfseries\boldmath UAVid & \bfseries\boldmath City\\
\midrule
Exact (FA2) & .002 & .002 & .002 & .002 & 1.00$\times$ & 1.00$\times$ & 1.00$\times$ & 1.00$\times$ & 1.00$\times$ & 1.00$\times$ & 1.00$\times$ & 1.00$\times$ & 7.42 & 1.00 & 1.00 & 59.13 & 74.84\\
FA4~\cite{zadouri2026flashattention} & .002 & .002 & .002 & .002 & 1.26$\times$ & 2.61$\times$ & 2.97$\times$ & 2.94$\times$ & 1.10$\times$ & 1.77$\times$ & 2.51$\times$ & 2.80$\times$ & 7.42 & -- & -- & 59.13 & 74.84\\
\midrule
\rowcolor{xsel}[\tabcolsep][\rov] \multicolumn{18}{l}{\textit{\textbf{Sparse selection}}}\\
\rowcolor{xsel}[\tabcolsep][\rov] NATTEN~\cite{hassani2026gnafindings} & .153 & .346 & .484 & .685 & 0.25$\times$ & 1.91$\times$ & 9.49$\times$ & 23.9$\times$ & 0.44$\times$ & 1.63$\times$ & 4.95$\times$ & 18.3$\times$ & 7.45 & .846 & .931 & 58.43 & 71.33\\
\rowcolor{xsel}[\tabcolsep][\rov] XAttention~\cite{xu2025xattention} & .493 & .285 & .055 & .047 & 0.22$\times$ & 0.66$\times$ & 1.08$\times$ & 1.58$\times$ & 0.57$\times$ & 0.73$\times$ & 0.98$\times$ & 1.16$\times$ & 7.42 & .867 & .836 & 55.54 & 64.68\\
\rowcolor{xsel}[\tabcolsep][\rov] SpargeAttn~\cite{zhang2025sparge} & .108 & .096 & .063 & .055 & 0.53$\times$ & 1.94$\times$ & 2.76$\times$ & 2.95$\times$ & 0.79$\times$ & 1.30$\times$ & 2.17$\times$ & 2.89$\times$ & 7.44 & .983 & .988 & 58.72 & 74.38\\
\rowcolor{xsel}[\tabcolsep][\rov] ToDo~\cite{smith2024todo} & .445 & .292 & .115 & .073 & 1.01$\times$ & 2.57$\times$ & 3.56$\times$ & 3.89$\times$ & 1.05$\times$ & 1.69$\times$ & 2.76$\times$ & 3.58$\times$ & 7.42 & .945 & .942 & 59.01 & 68.78\\
\rowcolor{xsum}[\tabcolsep][\rov] \multicolumn{18}{l}{\textit{\textbf{Clustering and compensation}}}\\
\rowcolor{xsum}[\tabcolsep][\rov] PISA~\cite{li2026pisa} & \textbf{.054} & \textbf{.021} & \textbf{.017} & .013 & 0.43$\times$ & 1.71$\times$ & 2.55$\times$ & 2.19$\times$ & 0.76$\times$ & 1.23$\times$ & 1.56$\times$ & 1.54$\times$ & 7.42 & .997 & .997 & 58.90 & 73.12\\
\rowcolor{xsum}[\tabcolsep][\rov] ClusterAttn~\cite{nordenram2026clusterattention} & .099 & .066 & .032 & .021 & 0.14$\times$ & 0.64$\times$ & 1.41$\times$ & 1.69$\times$ & 0.32$\times$ & 0.73$\times$ & 1.35$\times$ & 1.67$\times$ & 9.22 & .939 & .993 & 55.29 & 74.27\\
\rowcolor{xsum}[\tabcolsep][\rov] i-clustered~\cite{vyas2020clustered} & .253 & .247 & .219 & .216 & 0.06$\times$ & 0.38$\times$ & 1.85$\times$ & 6.50$\times$ & 0.16$\times$ & 0.46$\times$ & 1.59$\times$ & 4.85$\times$ & 80.9 & .615 & .798 & 26.74 & 50.24\\
\rowcolor{xapx}[\tabcolsep][\rov] \multicolumn{18}{l}{\textit{\textbf{Low-rank and hashing}}}\\
\rowcolor{xapx}[\tabcolsep][\rov] FNA~\cite{lu2025artifacts} & .360 & .399 & .290 & .280 & 0.02$\times$ & 0.12$\times$ & 0.51$\times$ & 2.23$\times$ & 0.37$\times$ & 0.83$\times$ & 1.33$\times$ & 1.59$\times$ & 7.42 & .931 & .966 & 57.33 & 73.78\\
\rowcolor{xapx}[\tabcolsep][\rov] HyperAttention~\cite{han2024hyperattention} & .764 & .541 & .356 & .344 & 0.24$\times$ & 2.01$\times$ & 12.4$\times$ & 23.0$\times$ & 0.80$\times$ & 1.09$\times$ & 1.25$\times$ & 1.31$\times$ & 7.42 & .816 & .858 & 52.97 & 66.81\\
\rowcolor{xapx}[\tabcolsep][\rov] SMYRF~\cite{daras2020smyrf} & .426 & .367 & .195 & .182 & 0.02$\times$ & 0.08$\times$ & 0.32$\times$ & 1.24$\times$ & 0.07$\times$ & 0.12$\times$ & 0.35$\times$ & 1.23$\times$ & 202.5 & .703 & .763 & 35.47 & 54.50\\
\rowcolor{xapx}[\tabcolsep][\rov] Scatterbrain~\cite{chen2021scatterbrain} & .749 & .552 & -- & -- & 0.07$\times$ & 0.16$\times$ & -- & -- & 0.20$\times$ & 0.22$\times$ & -- & OOM & OOM & .446 & .370 & 16.95 & 9.44\\
\midrule
\rowcolor{xmeros}[\tabcolsep][\rov] \meros $r{=}2$ & \cellcolor{xerr!51!white}.073 & \cellcolor{xerr!33!white}.041 & \cellcolor{xerr!14!white}.020 & \cellcolor{xerr!2!white}\textbf{.010} & \cellcolor{xspd!2!white}2.30$\times$ & \cellcolor{xspd!21!white}4.80$\times$ & \cellcolor{xspd!34!white}6.08$\times$ & \cellcolor{xspd!27!white}6.05$\times$ & 2.43$\times$ & 3.79$\times$ & 5.20$\times$ & 5.75$\times$ & 7.19 & \textbf{.998} & \textbf{.998} & \textbf{59.04} & \textbf{74.68}\\
\rowcolor{xmeros}[\tabcolsep][\rov] \meros $r{=}4$ & \cellcolor{xerr!64!white}.123 & \cellcolor{xerr!45!white}.071 & \cellcolor{xerr!27!white}.035 & \cellcolor{xerr!8!white}.018 & \cellcolor{xspd!8!white}3.42$\times$ & \cellcolor{xspd!40!white}7.89$\times$ & \cellcolor{xspd!53!white}11.7$\times$ & \cellcolor{xspd!59!white}12.5$\times$ & 2.75$\times$ & 4.75$\times$ & 8.28$\times$ & 11.2$\times$ & 6.94 & .995 & .994 & 58.96 & 74.38\\
\rowcolor{xmeros}[\tabcolsep][\rov] \meros $r{=}8$ & \cellcolor{xerr!70!white}.180 & \cellcolor{xerr!58!white}.106 & \cellcolor{xerr!39!white}.052 & \cellcolor{xerr!21!white}.027 & \cellcolor{xspd!15!white}\textbf{4.04}$\times$ & \cellcolor{xspd!47!white}\textbf{11.0}$\times$ & \cellcolor{xspd!66!white}\textbf{21.0}$\times$ & \cellcolor{xspd!72!white}\textbf{24.8}$\times$ & \textbf{2.93}$\times$ & \textbf{5.35}$\times$ & \textbf{11.6}$\times$ & \textbf{19.4}$\times$ & \textbf{6.81} & .990 & .987 & 58.79 & 73.73\\
\bottomrule
\end{tabular}\end{widewrap}
\label{tab:tf}
\end{table*}

\noindent\textbf{Stratified, reflected sample.} \cref{lem:unbiased,thm:error} only require inclusion probability $1/r$ and independent cells, so \meros samples each $4\times4$ cell on its own. Within a cell, each position $u=(y,x)$ pairs with its reflection $\bar u=(3-y,3-x)$ through the cell center (\cref{fig:method}(1)). Each cell keeps the first $8/r$ of its eight pairs in a random order, drawn once per layer and head before inference. Every key is thus sampled with probability exactly $1/r$, which no fixed grid achieves, and the samples for $r=8,4,2$ are nested. The sample needs no information about the image, so it is stored once per head and reused for every input. Finer cells can only lower $\sum_cP_c^2$, but $4\times4$ is the finest square that keeps a whole pair at $r=8$, which needs at least 16 positions per cell (\cref{sec:supp_details}).
\begin{proposition}[Reflected pairs]
\label{prop:pairs}
Let $\bff$ be a function on the positions of a far cell $c$, with even and odd parts $\bff^{\pm}(u)=\tfrac12\bigl(\bff(u)\pm\bff(\bar u)\bigr)$, and let $\hat\bT_c=r\sum_{u\in\Samp\cap c}\bff(u)$ estimate $\bT_c=\sum_{u\in c}\bff(u)$. (i)~The error $\hat\bT_c-\bT_c$ depends only on $\bff^{+}$, so the estimate is exact for every draw if $\bff$ is affine on $c$. (ii)~Its mean squared error is $\frac{15}{7}V^{+}$, against $V^{+}+V^{-}$ for simple random sampling of $16/r$ positions of the cell, where $V^{\pm}$ is its mean squared error on $\bff^{\pm}$ alone.
\end{proposition}

\noindent\emph{Derivation.} A sampled pair contributes $\bff(u)+\bff(\bar u)=2\,\bff^{+}(u)$, so the odd part never enters the estimate, and an affine $\bff$ has a constant even part. The cell keeps $8/r$ of its eight pair sums without replacement, a simple random sample from eight units~\cite[Ch.~2]{cochran1977sampling}. Its finite-population divisor is $8-1$ instead of the $16-1$ of single positions, hence the factor $\frac{15}{7}$.

\noindent For $\bff(u)=p_u(\bv_u-\bo)$, the summand of $\bU$, pairs cancel the odd part, which holds every linear trend across the cell (\cref{fig:method}(c)), at the price of a factor $\frac{15}{7}$ on the even part. They pay off when $V^->\frac87V^+$, as for smoothly varying features.\PUP

\noindent\textbf{One normalizer.} \meros puts the exact terms of query $i$ and its far estimates under one normalizer (\cref{fig:method}(2)):
\begin{equation}
  \hat\bo_i=\frac{\cex{\bA_{\Loc_i}}+\cglob{\bA_{\Glob}}+\cfar{\hat\bA_{\Far_i}}}{\cex{Z_{\Loc_i}}+\cglob{Z_{\Glob}}+\cfar{\hat Z_{\Far_i}}} .
  \label{eq:meros}
\end{equation}
Since $r\,e^{s}=e^{s+\log r}$, this is softmax over $\Loc_i\cup\Glob\cup(\Samp\cap\Far_i)$ with $\log r$ added to every far logit, and $r=1$ is exact.

\begin{proposition}[Anchored normalizer]
\label{prop:anchor}
For every draw of $\Samp$, $1+\delta\ge\pex$, and hence $\|\hat\bo-\bo\|\le\|\bU\|/\pex$.
\end{proposition}

\noindent\emph{Derivation.} The estimated normalizer is $Z(1+\delta)=Z_{\Loc_i}+Z_{\Glob}+\hat Z_{\Far_i}$ with $\hat Z_{\Far_i}\ge0$, so $1+\delta\ge(Z_{\Loc_i}+Z_{\Glob})/Z=\pex$, and the identity behind \cref{eq:identity} gives the bound. The exact terms thus put a floor under the normalizer, so even a sample that misses most far mass amplifies $\bU$ by at most $1/\pex$, about $8\times$ in \cref{fig:method}(a). Sampling the peak as well would remove this floor, so \meros never samples the tile. This bound holds on every draw, with no expectation taken.

\noindent\textbf{Why \meros is fast and general.} The sample of \meros is fixed before inference, and the 256 queries of a tile share every loaded key in one dense pass (\cref{alg:meros}). Since every key is read at its own position, position biases and masks apply unchanged, so \meros is compatible with most pretrained ViTs and VLMs with global self-attention (\cref{tab:supp_models}). 

\meros also runs as efficiently on the hardware as the well-optimized FlashAttention. For each query--key pair it needs, the kernel of \meros sustains 51 to 54\% of the dense peak of a B300 GPU, whereas FA4 reaches 49 to 50\% (\cref{sec:practice}). At 4096\,px and above, its speedup over FA4 is therefore close to the reduction in keys.

\meros changes only which far keys a query reads. The sampled keys are weighted so that the far-field sums keep their mean, and the bound on the resulting error tightens with every doubling of the image side. The budget $r$ is its only setting. Raising it from 2 to 8 cuts the far-field cost fourfold and raises the error by about $\sqrt7$ (\cref{fig:supp_theory}(f)).

\begin{figure}[t]
\centering
\includegraphics[width=\textwidth]{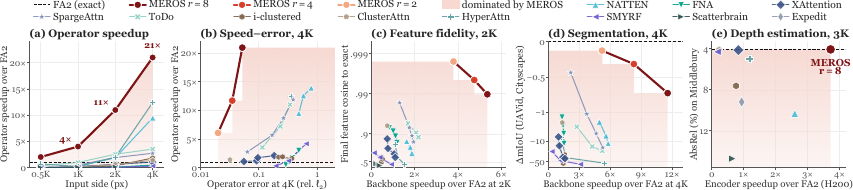}
\caption{\textbf{Comparison with training-free attention acceleration methods.} Shaded areas are slower and less accurate than \meros. (a)~Speedup of a \mbox{DINOv3-B}~\cite{simeoni2025dinov3} attention layer. (b)~Speedup against error at 4K. (c)~Final-feature cosine on UAVid~\cite{lyu2020uavid} and (d)~mean $\Delta$mIoU on UAVid and Cityscapes~\cite{cordts2016cityscapes}, with backbone speedup. (e)~AbsRel of Depth Anything~V2-L~\cite{yang2024dav2} on Middlebury~\cite{scharstein2014middlebury}.}
\label{fig:speed}
\vspace{-1.0em}
\end{figure}

\section{Experiments}
\label{sec:exp}

We compare \meros with attention acceleration methods (\cref{fig:speed}) on semantic and dichotomous segmentation, image matting, depth estimation (\cref{fig:qual}) and visual question answering using VLMs, five tasks that require dense representations. We use FlashAttention-2 as the attention baseline and compute speedups relative to it unless noted.

\noindent\textbf{Setup.} \meros replaces every global attention layer at inference without training. For segmentation backbones, one linear head is trained on frozen exact features and shared by all methods. Released matting and depth models are used as they are. \emph{Exact} is attention through FA2~\cite{dao2024flashattention2}. Baselines run their official code in default settings unless noted, and ClusterAttn keeps the CUDA graphs that its batched eigensolver needs. We use $r\in\{2,4,8\}$ with three sampling seeds. Timings of \meros and all compared methods use one NVIDIA B300, BF16 and batch~1 unless noted otherwise.

\begin{figure}[t]
\centering
\includegraphics[width=\textwidth]{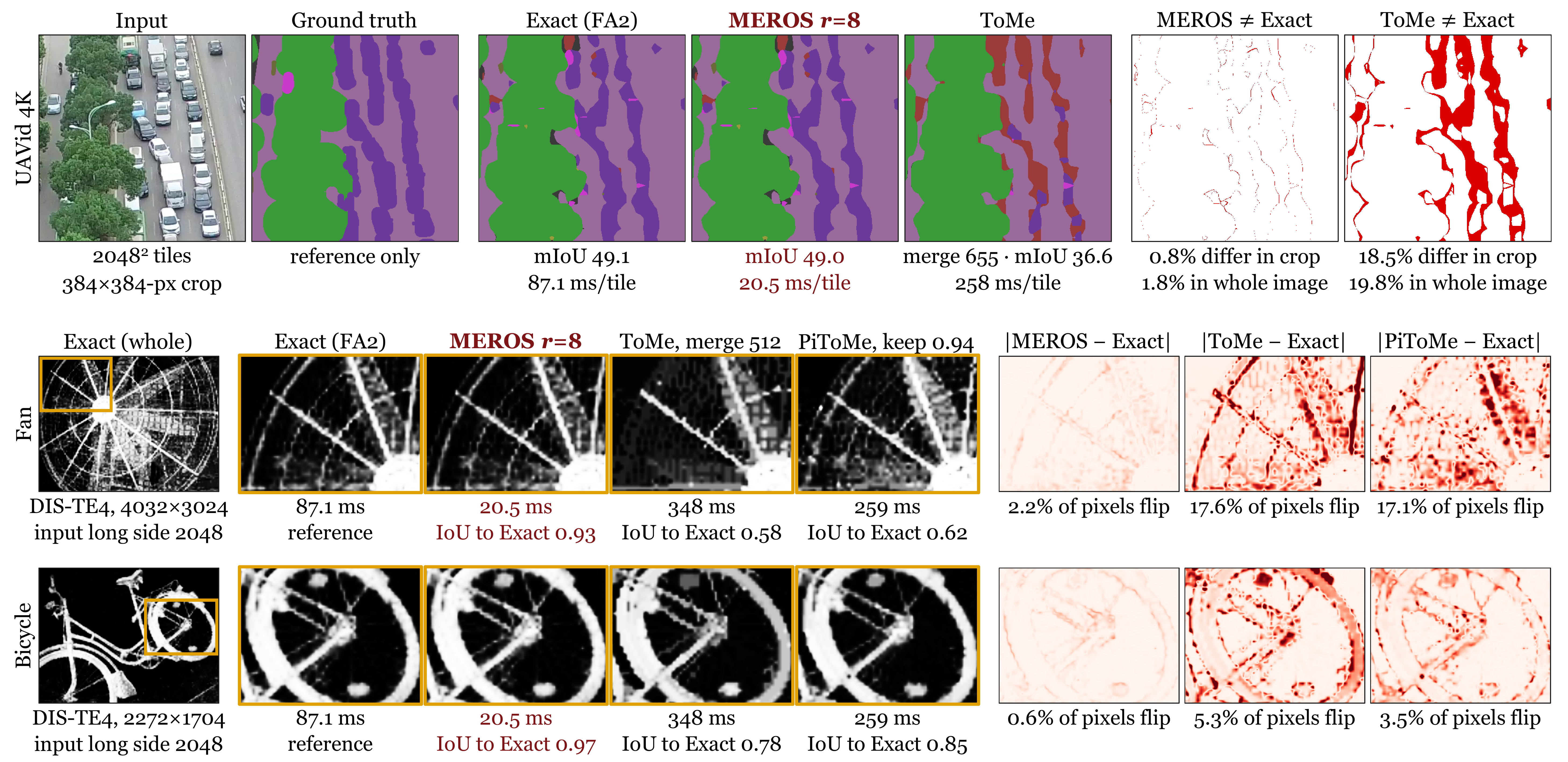}\par\vspace{-1pt}
\caption{\textbf{Comparisons with token merging on MAE-L~\cite{he2022mae}.} We compare exact attention (FA2), \meros $r{=}8$, ToMe~\cite{bolya2023tome} and PiToMe~\cite{tran2024pitome}.
The top row shows a crop of a UAVid~\cite{lyu2020uavid} tile of 2048\,px with the ground truth. Red marks the pixel differences.
ToMe merges 655 tokens per block, the largest merge of our MAE-L sweep in \cref{tab:supp_tome}, which still spends more MACs than \meros $r{=}8$ on UAVid. Labels give the mIoU on UAVid and the backbone time per $2048^2$ tile on one B300. The lower two rows show two DIS-TE4~\cite{qin2022dis} images. ToMe merges 512 tokens per block and PiToMe keeps 0.94 of the tokens, the settings whose MACs are closest to those of $r{=}8$. The maps show the absolute difference in foreground probability from exact attention, on a scale from 0 to 0.5.}
\label{fig:supp_gal_tome}
\vspace{-1.2em}
\end{figure}

\subsection{Comparison with Training-Free Methods}
\label{sec:exp_tf}

We first compare \meros with other training-free acceleration methods, on a single attention layer and on the whole backbone. Specifically, \cref{tab:tf} compares it with eleven attention acceleration methods on \mbox{DINOv3-B}, and \cref{tab:tome} with token merging, which shortens the token sequence instead. At the operator level, $r{=}2$ has the lowest error at 8K, and $r{=}8$ is the fastest at every size. Some budget of \meros is both faster and more accurate than 26 of the 28 baseline variants at 2K and 24 of 25 at 4K. The exceptions, two PISA variants, run 2.2--2.4$\times$ slower than $r{=}2$ at equal error.

On the whole backbone, only SpargeAttn stays within 0.7~mIoU of exact attention on both datasets, with a 2.2$\times$ speedup at 4K. Within the same margin, \meros $r{=}4$ reaches 8.3$\times$, and $r{=}8$ reaches 11.6$\times$ within 1.2~mIoU. NATTEN gains speed only by confining each query to a window, which costs 3.5~mIoU on Cityscapes.\PFILL

\begin{wraptable}{r}{0.5\textwidth}
\vspace{-7pt}
\centering
\tabsetup{7.4}
\renewcommand{\arraystretch}{0.98}
\setlength{\tabcolsep}{2pt}
\caption{\textbf{Token merging} on UAVid~\cite{lyu2020uavid}, per $2048^2$ tile. PiToMe matches \meros $r{=}8$ in MACs and ToMe Exact (FA2) in mIoU.}
\resizebox{\linewidth}{!}{\begin{tabular}{w{l}{44.2pt}*{3}{w{c}{19.4pt}w{c}{15.7pt}w{c}{15.7pt}}}
\toprule
\rowcolor{xhead}[\tabcolsep][\rov] & \multicolumn{3}{c}{\bfseries\boldmath MAE-B~\cite{he2022mae}} & \multicolumn{3}{c}{\bfseries\boldmath MAE-L~\cite{he2022mae}} & \multicolumn{3}{c}{\bfseries\boldmath DeiT\,III-L~\cite{touvron2022deit3}}\\
\hcmid\cmidrule(l{0.6pt}r{2.4pt}){2-4}\cmidrule(l{0.6pt}r{2.4pt}){5-7}\cmidrule(l{0.6pt}r{2.4pt}){8-10}
\rowcolor{xhead}[\tabcolsep][\rov] \bfseries\boldmath Method & \bfseries mIoU$\uparrow$ & \bfseries ms$\downarrow$ & \bfseries GB$\downarrow$ & \bfseries mIoU$\uparrow$ & \bfseries ms$\downarrow$ & \bfseries GB$\downarrow$ & \bfseries mIoU$\uparrow$ & \bfseries ms$\downarrow$ & \bfseries GB$\downarrow$\\
\midrule
Exact (FA2) & 35.43 & 35.2 & 0.51 & 49.11 & 87.1 & 1.02 & 55.95 & 86.4 & 1.06\\
FA4~\cite{zadouri2026flashattention} & -- & 16.7 & 0.51 & -- & 41.0 & 1.02 & -- & 40.3 & 1.06\\
\midrule
\rowcolor{xsum}[\tabcolsep][\rov] PiToMe~\cite{tran2024pitome} & 31.10 & 125 & 13.1 & 38.94 & 215 & 17.3 & 44.87 & 206 & 17.8\\
\rowcolor{xsum}[\tabcolsep][\rov] ToMe~\cite{bolya2023tome} & -- & -- & -- & 49.02 & 711 & 20.7 & \textbf{55.96} & 600 & 19.3\\
\midrule
\rowcolor{xmeros}[\tabcolsep][\rov] \meros $r{=}2$ & 35.44 & 12.4 & 0.72 & \textbf{49.10} & 28.8 & 1.30 & 55.95 & 28.4 & 1.32\\
\rowcolor{xmeros}[\tabcolsep][\rov] \meros $r{=}4$ & \textbf{35.45} & 10.1 & 0.70 & 49.06 & 22.8 & 1.27 & 55.94 & 22.3 & \textbf{1.29}\\
\rowcolor{xmeros}[\tabcolsep][\rov] \meros $r{=}8$ & 35.43 & \textbf{9.08} & \textbf{0.69} & 48.97 & \textbf{20.5} & \textbf{1.24} & 55.91 & \textbf{20.0} & \textbf{1.29}\\
\bottomrule
\end{tabular}}
\label{tab:tome}
\end{wraptable}
Token merging is slower than exact attention in its official code. On MAE-L, PiToMe takes 215\,ms and loses 10.2~mIoU, against 87\,ms for exact attention and 20.5\,ms for \meros, and ToMe needs 711\,ms to match exact attention. A merged token also returns one feature to all the patches it absorbed. In \cref{fig:supp_gal_tome}, ToMe then changes the label of 19.8\% of the pixels of a UAVid image, against 1.8\% for \meros. In the crop, ToMe labels whole rows of cars as buildings, while \meros changes 0.8\% of the labels. On two DIS-TE4 images, ToMe and PiToMe flip 3.5 to 17.6\% of the pixels, whereas \meros flips 0.6 and 2.2\% and runs the backbone over twelve times faster. In the five settings where some merge removes as many MACs as $r{=}8$, $r{=}8$ is 1.1 to 10.0~mIoU more accurate than the best such merge (\cref{tab:supp_tome_macs}).

\afterpage{\begin{table*}[t]
\centering
\tabsetup{7.4}
\renewcommand{\arraystretch}{0.98}
\setlength{\tabcolsep}{2pt}
\caption{\textbf{Training-free acceleration of semantic segmentation}, one frozen linear head per setting. Inputs are tiles (T) or whole images on a canvas (C) or at native size (N). Speed is backbone speedup over Exact (FA2). \meros results average three seeds.}
\begin{widewrap}\begin{tabular}{lll*{7}{>{\centering\arraybackslash}p{25.276pt}}*{6}{>{\cellcolor{xmeros}\centering\arraybackslash}p{25.276pt}}}
\toprule
\rowcolor{xhead}[\tabcolsep][\rov] & & & \bfseries\boldmath Exact & \multicolumn{2}{c}{\bfseries\boldmath SpargeAttn~\cite{zhang2025sparge}} & \multicolumn{2}{c}{\bfseries\boldmath NATTEN~\cite{hassani2026gnafindings}} & \multicolumn{2}{@{}c@{}}{\bfseries\boldmath ClusterAttn~\cite{nordenram2026clusterattention}} & \multicolumn{3}{>{\cellcolor{xmeros}}c}{\bfseries\boldmath \meros, $\Delta$mIoU $\uparrow$} & \multicolumn{3}{>{\cellcolor{xmeros}}c}{\bfseries\boldmath \meros, speed $\uparrow$}\\
\hcmid\cmidrule(lr){5-6}\cmidrule(lr){7-8}\cmidrule(lr){9-10}\cmidrule(lr){11-13}\cmidrule(lr){14-16}
\rowcolor{xhead}[\tabcolsep][\rov] \bfseries\boldmath Dataset & \bfseries\boldmath Backbone & \bfseries\boldmath Input & \makebox[\linewidth][c]{\bfseries\boldmath mIoU\,$\uparrow$} & \bfseries\boldmath $\Delta$mIoU & \bfseries\boldmath Speed & \bfseries\boldmath $\Delta$mIoU & \bfseries\boldmath Speed & \bfseries\boldmath $\Delta$mIoU & \bfseries\boldmath Speed & \bfseries\boldmath $r{=}2$ & \bfseries\boldmath $r{=}4$ & \bfseries\boldmath $r{=}8$ & \bfseries\boldmath $r{=}2$ & \bfseries\boldmath $r{=}4$ & \bfseries\boldmath $r{=}8$\\
\midrule
\rowcolor{dsuav}[\tabcolsep][\rov] UAVid & DINOv3-B~\cite{simeoni2025dinov3} & T\,2048 & 59.13 & $-$0.41 & 1.30$\times$ & $-$0.70 & 1.62$\times$ & $-$3.84 & 0.73$\times$ & $-$0.09 & $-$0.17 & $-$0.33 & 3.79$\times$ & 4.75$\times$ & 5.35$\times$\\
\rowcolor{dsuav}[\tabcolsep][\rov] \cite{lyu2020uavid} & MAE-B~\cite{he2022mae} & T\,2048 & 35.43 & $-$0.61 & 1.54$\times$ & $-$12.29 & 2.00$\times$ & $-$0.49 & 0.70$\times$ & $+$0.02 & $+$0.02 & $+$0.00 & 2.84$\times$ & 3.49$\times$ & 3.87$\times$\\
\rowcolor{dsuav}[\tabcolsep][\rov]  & MAE-L~\cite{he2022mae} & T\,2048 & 49.11 & $-$1.10 & 1.44$\times$ & $-$19.41 & 2.11$\times$ & $-$0.59 & 0.77$\times$ & $-$0.01 & $-$0.04 & $-$0.13 & 3.03$\times$ & 3.82$\times$ & 4.25$\times$\\
\rowcolor{dsuav}[\tabcolsep][\rov]  & DeiT\,III-B~\cite{touvron2022deit3} & T\,2048 & 51.86 & $-$1.43 & 1.56$\times$ & $-$12.18 & 1.93$\times$ & $-$1.00 & 0.71$\times$ & $+$0.01 & $+$0.00 & $+$0.05 & 3.04$\times$ & 3.83$\times$ & 4.33$\times$\\
\rowcolor{dsuav}[\tabcolsep][\rov]  & DeiT\,III-L~\cite{touvron2022deit3} & T\,2048 & 55.95 & $-$0.44 & 1.66$\times$ & $-$3.86 & 1.97$\times$ & $-$0.26 & 0.77$\times$ & $+$0.00 & $-$0.01 & $-$0.05 & 3.04$\times$ & 3.85$\times$ & 4.32$\times$\\
\rowcolor{dsuav}[\tabcolsep][\rov]  & SigLIP2-B~\cite{tschannen2025siglip2} & T\,2048 & 40.72 & $-$5.98 & 1.55$\times$ & $-$13.54 & 2.40$\times$ & $+$0.03 & 0.76$\times$ & $-$0.14 & $-$0.31 & $-$0.36 & 2.63$\times$ & 3.06$\times$ & 3.34$\times$\\
\rowcolor{dsuav}[\tabcolsep][\rov]  & SigLIP2-L~\cite{tschannen2025siglip2} & T\,2048 & 47.43 & $-$4.31 & 1.63$\times$ & $-$15.75 & 2.58$\times$ & $-$0.21 & 0.82$\times$ & $-$0.03 & $-$0.05 & $-$0.33 & 2.78$\times$ & 3.28$\times$ & 3.50$\times$\\
\midrule
\rowcolor{dscity}[\tabcolsep][\rov] Cityscapes & DINOv3-B~\cite{simeoni2025dinov3} & T\,1024 & 74.84 & $-$0.46 & 0.79$\times$ & $-$3.51 & 0.44$\times$ & $-$0.57 & 0.32$\times$ & $-$0.16 & $-$0.45 & $-$1.11 & 2.43$\times$ & 2.75$\times$ & 2.93$\times$\\
\rowcolor{dscity}[\tabcolsep][\rov] \cite{cordts2016cityscapes} & MAE-B~\cite{he2022mae} & T\,1024 & 26.67 & $-$3.92 & 0.45$\times$ & $-$2.47 & 0.53$\times$ & $-$0.81 & 0.31$\times$ & $-$0.01 & $-$0.09 & $-$0.15 & 1.30$\times$ & 1.38$\times$ & 1.46$\times$\\
\rowcolor{dscity}[\tabcolsep][\rov]  & MAE-L~\cite{he2022mae} & T\,1024 & 40.07 & $-$2.21 & 0.38$\times$ & $-$3.11 & 0.56$\times$ & $-$0.22 & 0.28$\times$ & $-$0.09 & $-$0.27 & $-$0.56 & 1.46$\times$ & 1.56$\times$ & 1.65$\times$\\
\midrule
\rowcolor{dsglobe}[\tabcolsep][\rov] DeepGlobe & DINOv3-B~\cite{simeoni2025dinov3} & T\,2048 & 62.66 & $-$0.36 & 1.30$\times$ & $-$3.05 & 1.62$\times$ & $-$6.47 & 0.73$\times$ & $-$0.07 & $-$0.29 & $-$0.57 & 3.79$\times$ & 4.75$\times$ & 5.35$\times$\\
\rowcolor{dsglobe}[\tabcolsep][\rov] \cite{demir2018deepglobe} & DINOv3-B~\cite{simeoni2025dinov3} & C\,2560 & 62.13 & $-$0.14 & 1.51$\times$ & $-$6.55 & 2.28$\times$ & $-$4.98 & 0.97$\times$ & $+$0.07 & $-$0.03 & $-$0.38 & 4.23$\times$ & 5.79$\times$ & 6.95$\times$\\
\rowcolor{dsglobe}[\tabcolsep][\rov]  & DINOv3-B~\cite{simeoni2025dinov3} & N\,2448 & 57.60 & $+$0.12 & 1.51$\times$ & $-$1.01 & 2.28$\times$ & $-$7.74 & 0.97$\times$ & $-$0.08 & $-$0.23 & $-$0.62 & 4.23$\times$ & 5.79$\times$ & 6.95$\times$\\
\midrule
\rowcolor{dsurur}[\tabcolsep][\rov] URUR & DINOv3-B~\cite{simeoni2025dinov3} & T\,2048 & 41.08 & $+$0.17 & 1.30$\times$ & $+$0.17 & 1.62$\times$ & $-$8.47 & 0.73$\times$ & $-$0.21 & $-$0.53 & $-$0.95 & 3.79$\times$ & 4.75$\times$ & 5.35$\times$\\
\rowcolor{dsurur}[\tabcolsep][\rov] \cite{ji2023urur} & DINOv3-B~\cite{simeoni2025dinov3} & T\,4096 & 24.23 & $-$0.68 & 2.17$\times$ & $+$0.11 & 4.95$\times$ & $-$2.02 & 1.35$\times$ & $-$0.10 & $-$0.23 & $-$0.30 & 5.20$\times$ & 8.28$\times$ & 11.6$\times$\\
\rowcolor{dsurur}[\tabcolsep][\rov]  & DINOv3-B~\cite{simeoni2025dinov3} & C\,5120 & 37.20 & $-$0.90 & 2.29$\times$ & $-$1.75 & 7.43$\times$ & $-$1.75 & 1.49$\times$ & $-$0.11 & $-$0.19 & $-$0.41 & 5.53$\times$ & 9.33$\times$ & 14.0$\times$\\
\rowcolor{dsurur}[\tabcolsep][\rov]  & DINOv3-B~\cite{simeoni2025dinov3} & N\,5120 & 22.68 & $-$0.46 & 2.29$\times$ & $+$0.14 & 7.43$\times$ & $-$1.32 & 1.49$\times$ & $-$0.02 & $-$0.09 & $-$0.17 & 5.53$\times$ & 9.33$\times$ & 14.0$\times$\\
\midrule
\rowcolor{xhead}[\tabcolsep][\rov] \multicolumn{3}{l}{\textbf{Median of 17 settings}} & -- & $-$0.61 & 1.51$\times$ & $-$3.11 & 2.00$\times$ & $-$1.00 & 0.76$\times$ & \textbf{$-$0.07} & $-$0.17 & $-$0.33 & 3.04$\times$ & 3.85$\times$ & \textbf{4.33$\times$}\\
\rowcolor{xhead}[\tabcolsep][\rov] \multicolumn{3}{l}{\textbf{Worst of 17 settings}} & -- & $-$5.98 & 0.38$\times$ & $-$19.41 & 0.44$\times$ & $-$8.47 & 0.28$\times$ & \textbf{$-$0.21} & $-$0.53 & $-$1.11 & 1.30$\times$ & 1.38$\times$ & \textbf{1.46$\times$}\\
\bottomrule
\end{tabular}\end{widewrap}
\label{tab:seg}
\vspace{-0.6em}
\end{table*}
\begin{figure}[!t]
\centering
\includegraphics[width=\textwidth]{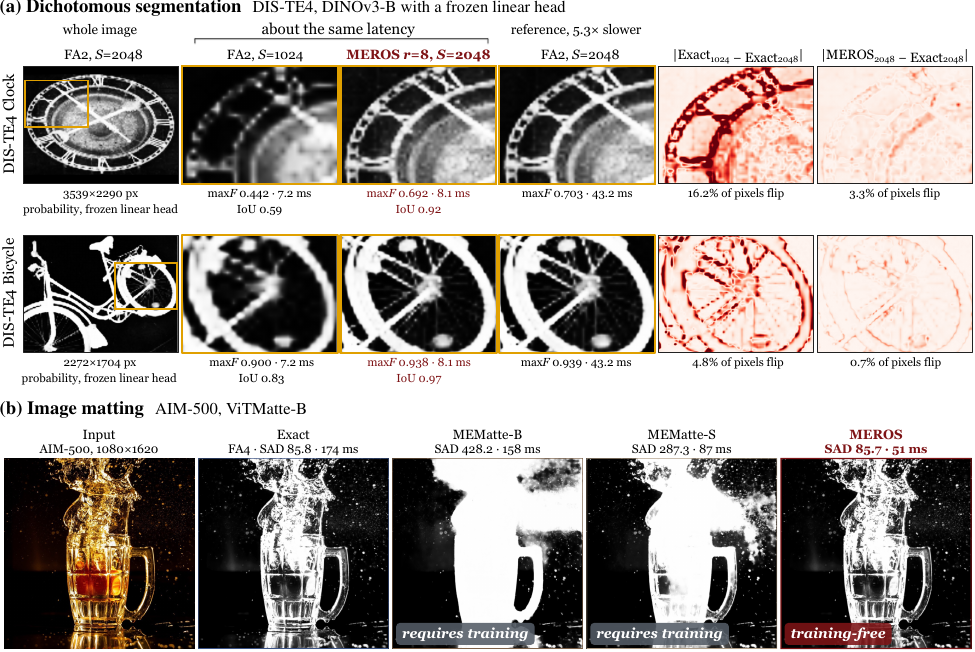}
\caption{\textbf{\boldmath Fine structures in dichotomous segmentation and matting. Higher $F^{\max}_\beta$ and IoU are better. Lower SAD, fewer flipped pixels and shorter times are better.} (a)~Two DIS-TE4 images~\cite{qin2022dis} at about equal latency, with exact attention at 1024\,px (7.2\,ms) and \meros $r{=}8$ at 2048\,px (8.1\,ms). Exact attention (FA2) at 2048\,px (43.2\,ms) is the reference for the difference maps (absolute change in foreground probability, from 0 to 0.5), the IoU and the share of flipped pixels. (b)~An AIM-500 image~\cite{li2021aim} with ViTMatte-B~\cite{yao2024vitmatte}. We compare exact attention (FA4), the trained MEMatte-B and MEMatte-S~\cite{lin2025mematte}, and \meros, which needs no training. Labels give the SAD to the ground truth and the time per image. SAD is the sum of absolute differences between two alpha mattes. In (a), the yellow box in the whole image marks the crop shown in the columns to its right.}
\label{fig:supp_gal_seg}
\vspace{-1.6em}
\end{figure}}
\subsection{Error and Speed across Resolutions}
\label{sec:exp_speed}

We next examine how the error and the speed of one \meros layer change with the input size. From 1K to 8K (\cref{tab:tf}, and 6K in \cref{tab:speed}), each doubling of the image side multiplies the operator error of \meros by 0.52 to 0.53 on average, close to the 1/2 of \cref{cor:resolution}. The error also grows with the budget as $\sqrt{r-1}$, the standard deviation of a stratified sample with inclusion probability $1/r$~\cite{cochran1977sampling}. Divided by this factor, the three budgets agree within 8\% at 1K and 4\% above, so any bias is small next to the sampling noise. Meanwhile, $r{=}8$ runs the layer 4.0--24.8$\times$ faster than FA2 and 3.2--8.4$\times$ faster than FA4. Its speedup rises at every step while the error of every budget falls about sevenfold, so at larger inputs \meros gains more over exact attention and deviates less from it.

\subsection{Semantic Segmentation}
\label{sec:exp_seg}

We evaluate \meros on semantic segmentation of high-resolution images, with frozen backbones and linear heads. Specifically, \cref{tab:seg} covers UAVid~\cite{lyu2020uavid} (aerial video frames), Cityscapes~\cite{cordts2016cityscapes} (street scenes), DeepGlobe~\cite{demir2018deepglobe} ($2448^2$ satellite images) and URUR~\cite{ji2023urur} ($5120^2$ satellite images). Its 17 settings span seven frozen backbones from four families, with tiles of 1024 to 4096\,px or whole images of up to $5120^2$\,px in one call.\PFILL

At $r{=}2$, \meros matches or exceeds exact attention in four of the 17 settings, by up to 0.07~mIoU at a 2.8 to 4.2$\times$ faster backbone, and stays within 0.1~mIoU in 13. Its median loss is 0.07~mIoU and its largest 0.21. At $r{=}8$, the median loss is 0.33~mIoU, with a median backbone speedup of 4.3$\times$. The speedup grows with the input. On URUR, $r{=}8$ gains 5.35$\times$ on 2048-px tiles, 11.6$\times$ on 4096-px tiles and 14.0$\times$ on whole images. A whole $5120^2$ URUR image takes 80 instead of 1{,}119\,ms at $r{=}8$. Others lose up to 6.0 (SpargeAttn), 19.4 (NATTEN) and 8.5~mIoU (ClusterAttn), with median speedups of at most 2.0$\times$. On the three Cityscapes settings, all three run slower than exact attention, at 0.28 to 0.79$\times$, while every budget of \meros is faster, at 1.30 to 2.93$\times$. Confining queries to a window, NATTEN loses 12.2 to 19.4~mIoU on MAE, DeiT\,III-B and SigLIP~2 on UAVid. ClusterAttn is slower than exact attention in 14 of the 17 settings, while the smallest speedup of \meros is 1.30$\times$. On UAVid, $r{=}8$ moves the mIoU of the MAE and DeiT\,III backbones by at most 0.13.\PUP

\subsection{Dichotomous Segmentation}
\label{sec:exp_dis}

\begin{wraptable}{r}{0.5\textwidth}
\vspace{-7pt}
\centering
\tabsetup{7.8}
\renewcommand{\arraystretch}{0.98}
\setlength{\tabcolsep}{1.729pt}
\caption{\textbf{Training-free acceleration of dichotomous segmentation} on DIS-TE~\cite{qin2022dis}. $M$ is the mean absolute error.}
\resizebox{\linewidth}{!}{\begin{tabular}{lccccccc}
\toprule
\rowcolor{xhead}[\tabcolsep][\rov] \bfseries\boldmath Method & \bfseries\boldmath $F^{\max}_\beta{\uparrow}$ & \bfseries\boldmath $F^{w}_\beta{\uparrow}$ & \bfseries\boldmath $M{\downarrow}$ & \bfseries\boldmath $S_\alpha{\uparrow}$ & \bfseries\boldmath $E^m_\phi{\uparrow}$ & \bfseries\boldmath HCE$_\gamma{\downarrow}$ & \bfseries\boldmath ms${\downarrow}$\\
\rowcolor{xlabel}[\tabcolsep][\rov] \multicolumn{8}{l}{\textit{\textbf{DINOv3-B~\cite{simeoni2025dinov3}, 1024\,px}}}\\
Exact & .781 & .606 & .087 & .807 & .814 & 1443 & 7.2\\
\rowcolor{xsel}[\tabcolsep][\rov] SpargeAttn~\cite{zhang2025sparge} & .767 & .601 & .090 & .801 & .810 & 1444 & 9.1\\
\rowcolor{xsel}[\tabcolsep][\rov] XAttention~\cite{xu2025xattention} & .760 & .586 & .093 & .794 & .795 & \textbf{1422} & 12.7\\
\rowcolor{xmeros}[\tabcolsep][\rov] \meros $r{=}2$ & \textbf{.779} & \textbf{.605} & \textbf{.088} & \textbf{.806} & \textbf{.813} & 1441 & 3.0\\
\rowcolor{xmeros}[\tabcolsep][\rov] \meros $r{=}8$ & .775 & .598 & .090 & .803 & .808 & 1438 & \textbf{2.5}\\
\midrule
\rowcolor{xlabel}[\tabcolsep][\rov] \multicolumn{8}{l}{\textit{\textbf{DINOv3-B~\cite{simeoni2025dinov3}, 2048\,px}}}\\
Exact & .816 & .624 & .085 & .823 & .828 & 1514 & 43.2\\
\rowcolor{xsel}[\tabcolsep][\rov] SpargeAttn~\cite{zhang2025sparge} & .813 & .612 & \textbf{.081} & .820 & .824 & 1527 & 33.2\\
\rowcolor{xsel}[\tabcolsep][\rov] NATTEN~\cite{hassani2026gnafindings} & .762 & .561 & .106 & .786 & .784 & 1532 & 26.5\\
\rowcolor{xsel}[\tabcolsep][\rov] XAttention~\cite{xu2025xattention} & .777 & .583 & .101 & .798 & .797 & 1525 & 58.9\\
\rowcolor{xmeros}[\tabcolsep][\rov] \meros $r{=}2$ & \textbf{.815} & \textbf{.622} & .086 & \textbf{.822} & \textbf{.827} & \textbf{1513} & 11.4\\
\rowcolor{xmeros}[\tabcolsep][\rov] \meros $r{=}8$ & .809 & .614 & .088 & .818 & .821 & \textbf{1513} & \textbf{8.1}\\
\midrule
\rowcolor{xlabel}[\tabcolsep][\rov] \multicolumn{8}{l}{\textit{\textbf{MAE-L~\cite{he2022mae}, 2048\,px}}}\\
Exact & .682 & .464 & .147 & .728 & .700 & 1556 & 87.1\\
\rowcolor{xsum}[\tabcolsep][\rov] PiToMe~\cite{tran2024pitome} & .677 & .456 & .150 & .725 & .694 & 1571 & 259\\
\rowcolor{xsel}[\tabcolsep][\rov] SpargeAttn~\cite{zhang2025sparge} & .679 & .462 & .151 & .723 & .696 & 1561 & 60.5\\
\rowcolor{xsel}[\tabcolsep][\rov] NATTEN~\cite{hassani2026gnafindings} & .621 & .383 & .159 & .676 & .624 & 1559 & 40.3\\
\rowcolor{xsel}[\tabcolsep][\rov] XAttention~\cite{xu2025xattention} & .630 & .425 & .165 & .695 & .668 & 1561 & 121\\
\rowcolor{xmeros}[\tabcolsep][\rov] \meros $r{=}2$ & \textbf{.681} & \textbf{.463} & \textbf{.147} & \textbf{.728} & \textbf{.700} & \textbf{1555} & 28.7\\
\rowcolor{xmeros}[\tabcolsep][\rov] \meros $r{=}8$ & .679 & .461 & .149 & .726 & .698 & \textbf{1555} & \textbf{20.5}\\
\bottomrule
\end{tabular}}
\label{tab:dis}
\end{wraptable}
We then evaluate \meros on dichotomous segmentation, which tests whether \meros preserves fine-grained structures. We use DIS5K~\cite{qin2022dis}, whose objects contain thin structures such as wires, spokes and lattices. One linear head, trained on frozen exact features at 2048\,px, serves every method, budget and input size on the 2{,}000 DIS-TE test images. With \mbox{DINOv3-B} at 2048\,px, $r{=}8$ lowers $F^{\max}_\beta$ by 0.007, keeps HCE$_\gamma$ and runs the backbone in 8.1 instead of 43.2\,ms (\cref{tab:dis}). This costs about as much as exact attention at 1024\,px and scores 0.028 higher in $F^{\max}_\beta$. On the two DIS-TE4 images of \cref{fig:supp_gal_seg}(a), exact attention at 1024\,px flips 16.2 and 4.8\% of the pixels of the 2048-px reference, against 3.3 and 0.7\% for $r{=}8$ at 2048\,px. On the clock, exact attention at 1024\,px reaches an $F^{\max}_\beta$ of 0.442, compared with 0.692 for $r{=}8$ and 0.703 for the reference.\PFILL

In all three settings of \cref{tab:dis}, both budgets of \meros need less human correction effort (HCE$_\gamma$) than exact attention, and both are faster than every baseline. $r{=}8$ takes 2.5, 8.1 and 20.5\,ms, and the baselines take 9.1 to 259\,ms. $r{=}2$ is more accurate than SpargeAttn on 17 of the 18 metrics, and SpargeAttn is slower than exact attention at 1024\,px. NATTEN and XAttention lose 0.02 to 0.06 in $F^{\max}_\beta$, and PiToMe on MAE-L is slower than exact attention, at 13 times the latency of $r{=}8$. With MAE-L, $r{=}2$ differs from exact attention by at most 0.001 on the five other metrics, in 28.7 instead of 87.1\,ms. Over both input sizes and MAE-L, no budget of \meros lowers $F^{\max}_\beta$ by more than 0.007, while $r{=}8$ runs the backbone 2.93 to 5.35$\times$ faster (\cref{tab:supp_dis}).\PFILL

\begin{table*}[t]
\centering
\tabsetup{7.6}
\renewcommand{\arraystretch}{0.98}
\setlength{\tabcolsep}{5.626pt}
\caption{\textbf{Image matting at native resolution.} The original ViTMatte-B~\cite{yao2024vitmatte} runs out of memory, so exact attention runs in query chunks. $^\dagger$Needs training. $^\ddagger$Our Blackwell port of a Hopper-only kernel. MSE in $10^{-3}$. Speed is speedup over exact attention on one B300, GB peak memory. w/o pos.\ bias means the kernel cannot add relative position terms. Bold is best below the top two rows.}
\begin{widewrap}\begin{tabular}{lcccccccccccc}
\toprule
\rowcolor{xhead}[\tabcolsep][\rov] & \multicolumn{6}{c}{\bfseries\boldmath PPM-100~\cite{ke2022modnet}, $2875{\times}2997$\,px on average} & \multicolumn{6}{c}{\bfseries\boldmath AIM-500~\cite{li2021aim}, $1260{\times}1397$\,px on average}\\
\hcmid\cmidrule(lr){2-7}\cmidrule(lr){8-13}
\rowcolor{xhead}[\tabcolsep][\rov] \bfseries\boldmath Method & \bfseries\boldmath SAD$\downarrow$ & \bfseries\boldmath MSE$\downarrow$ & \bfseries\boldmath Grad$\downarrow$ & \bfseries\boldmath Conn$\downarrow$ & \bfseries\boldmath Speed\rlap{$\uparrow$}\phantom{$\downarrow$} & \bfseries\boldmath GB$\downarrow$ & \bfseries\boldmath SAD$\downarrow$ & \bfseries\boldmath MSE$\downarrow$ & \bfseries\boldmath Grad$\downarrow$ & \bfseries\boldmath Conn$\downarrow$ & \bfseries\boldmath Speed\rlap{$\uparrow$}\phantom{$\downarrow$} & \bfseries\boldmath GB$\downarrow$\\
\midrule
ViTMatte-B, original & \multicolumn{5}{c}{out of memory on 36 of 100 images} & 260 & 14.4 & 11.69 & 11.1 & 13.7 & 0.90$\times$ & 16.0\\
ViTMatte-B, exact (chunks) & 35.3 & 13.92 & 36.5 & 32.3 & 1.00$\times$ & 22.3 & 14.4 & 11.69 & 11.1 & 13.7 & 1.00$\times$ & 3.7\\
\midrule
\rowcolor{xapx}[\tabcolsep][\rov] MEMatte-B$^\dagger$~\cite{lin2025mematte} & 37.1 & 15.80 & 41.6 & 35.0 & 2.96$\times$ & 33.3 & 16.7 & 15.60 & 13.9 & 16.0 & 1.12$\times$ & \textbf{3.6}\\
\rowcolor{xsum}[\tabcolsep][\rov] PiToMe~\cite{tran2024pitome}, keep 0.5 & 35.5 & 13.96 & 37.1 & 32.9 & 1.27$\times$ & 95.9 & 14.5 & 11.77 & 11.3 & 13.9 & 1.11$\times$ & 7.2\\
\rowcolor{xsum}[\tabcolsep][\rov] PISA~\cite{li2026pisa}, w/o pos.\ bias & 37.8 & 15.06 & 37.6 & 34.9 & 3.06$\times$ & 26.9 & 14.8 & 12.28 & 11.3 & 14.0 & 1.79$\times$ & 3.7\\
\rowcolor{xsel}[\tabcolsep][\rov] SpargeAttn~\cite{zhang2025sparge}, w/o pos.\ bias & 37.4 & 14.72 & 37.2 & 34.4 & 3.11$\times$ & 28.3 & 14.8 & 12.17 & 11.3 & 14.1 & 1.83$\times$ & 3.7\\
\rowcolor{xsum}[\tabcolsep][\rov] ClusterAttn$^\ddagger$~\cite{nordenram2026clusterattention}, w/o pos.\ bias & 38.0 & 15.20 & 37.8 & 35.2 & 2.96$\times$ & 27.3 & 14.8 & 12.35 & 11.4 & 14.1 & 0.68$\times$ & 3.8\\
\midrule
\rowcolor{xmeros}[\tabcolsep][\rov] \meros $r{=}2$ & \textbf{35.2} & \textbf{13.85} & \textbf{36.4} & \textbf{32.2} & 2.23$\times$ & 22.7 & \textbf{14.2} & \textbf{11.52} & \textbf{11.1} & \textbf{13.6} & 1.98$\times$ & 3.7\\
\rowcolor{xmeros}[\tabcolsep][\rov] \meros $r{=}4$ & 35.4 & 13.96 & 36.6 & 32.4 & 2.97$\times$ & \textbf{21.8} & 14.5 & 11.82 & 11.2 & 13.7 & 2.14$\times$ & 3.7\\
\rowcolor{xmeros}[\tabcolsep][\rov] \meros $r{=}8$ & 35.5 & 14.00 & 36.7 & 32.6 & \textbf{3.78}$\times$ & 22.7 & 14.6 & 11.94 & 11.2 & 13.8 & \textbf{2.63}$\times$ & 3.8\\
\bottomrule
\end{tabular}\end{widewrap}
\label{tab:matting}
\vspace{-1.0em}
\end{table*}

\subsection{Image Matting}
\label{sec:exp_matting}

We present image matting experiments in \cref{tab:matting}. We run the official ViTMatte-B~\cite{yao2024vitmatte} at native resolution on PPM-100~\cite{ke2022modnet} and AIM-500~\cite{li2021aim}, and compare with MEMatte~\cite{lin2025mematte}, trained to route tokens for memory-efficient matting. The original ViTMatte-B runs out of memory on 36 of the 100 PPM-100 images even with the 288\,GB of a B300. Exact therefore computes attention in query chunks, with the same output at about a tenth of the peak memory.\PFILL

\begin{figure}[p]
\centering
\includegraphics[width=\textwidth,height=596pt,keepaspectratio]{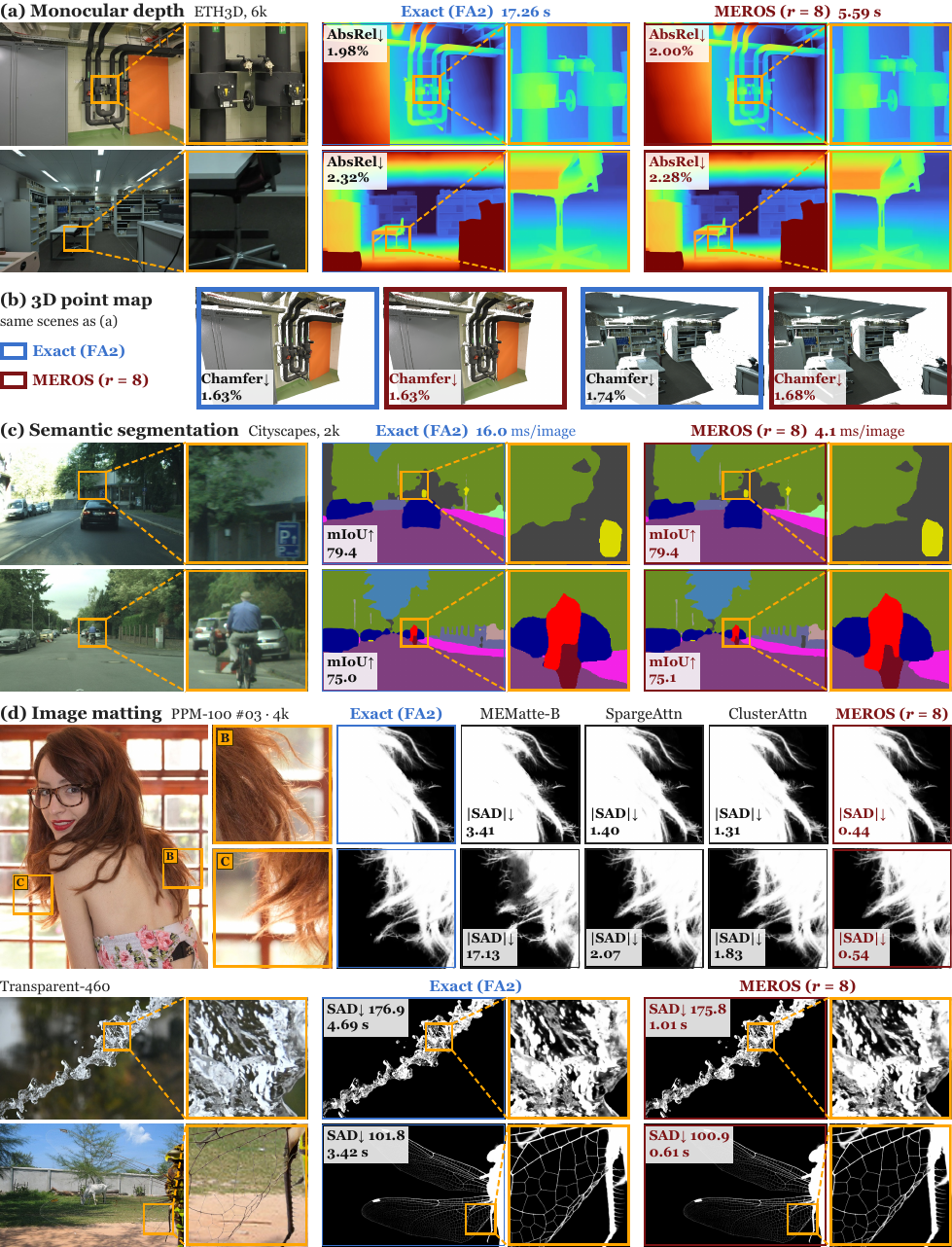}
\caption{\textbf{Results at native resolution. Lower AbsRel, Chamfer distance and SAD are better, and higher mIoU is better.} (a)~Depth and (b)~point maps of MoGe-3~\cite{kong2026moge3} on two ETH3D~\cite{schops2017eth3d} scenes. (c)~\mbox{DINOv3-B} \cite{simeoni2025dinov3} full image segmentation on Cityscapes~\cite{cordts2016cityscapes}. (d)~ViTMatte-B~\cite{yao2024vitmatte} on two hair crops (B, C) of a PPM-100~\cite{ke2022modnet} portrait, with $|$SAD$|$ to exact attention, and on two Transparent-460~\cite{cai2022transmatting} images, with SAD to the ground truth. Blue frames mark exact attention (FA2) and red frames \meros at $r{=}8$. Each zoom shows the yellow box of the picture on its left. The times in (a) are those of the whole depth model, with its decoder.}
\label{fig:qual}
\end{figure}

At $r{=}2$, \meros is even more accurate than exact attention while about twice as fast. Seven of the eight error columns of \cref{tab:matting} are lower, and one is equal. Over all budgets, it runs the whole backbone 2.0--3.8$\times$ faster than exact attention. Without training, it beats the trained MEMatte-B on every error metric of both datasets. Only four of the 12 blocks of ViTMatte-B attend globally, and \meros replaces these alone (\cref{sec:supp_matting}). PiToMe stays close to exact attention, but it is at most 1.27$\times$ faster and needs over four times the memory on PPM-100. PISA, SpargeAttn and ClusterAttn cannot add the relative position bias of ViTMatte in their kernels. Dropping this bias alone raises the SAD of exact attention from 35.3 to 38.0 on PPM-100. \meros keeps the bias (\cref{sec:supp_grid}). On PPM-100, $r{=}8$ is the fastest method at 3.78$\times$, and no baseline has a lower SAD. On AIM-500, it is again the fastest at 2.63$\times$, where the fastest baseline, SpargeAttn, reaches 1.83$\times$ at a higher SAD. The peak memory of \meros is within 0.5\,GB of that of exact attention in chunks. MEMatte-B is trained for matting and still loses 1.8 SAD on PPM-100 and 2.3 on AIM-500 relative to exact attention, whereas $r{=}8$ loses 0.2 on both and runs faster (3.78 vs.\ 2.96$\times$ and 2.63 vs.\ 1.12$\times$). In the hair crops of \cref{fig:qual}(d), \meros is within 0.6 SAD of exact attention, whereas the others differ from it by 1.3 to 17. On two Transparent-460 images of \cref{fig:qual}(d), it runs 4.6 and 5.6$\times$ faster at a lower SAD. Over the 50 test images of this dataset, from 2.8 to 30 megapixels, $r{=}8$ is 4.6$\times$ faster than exact attention in total. On the beer mug of \cref{fig:supp_gal_seg}(b), \meros matches the SAD of exact attention in 51\,ms instead of the 174\,ms of FA4, while the trained MEMatte-B and MEMatte-S reach 428.2 and 287.3.\PUP

\begin{table*}[t]
\centering
\tabsetup{7.2}
\renewcommand{\arraystretch}{0.98}
\setlength{\tabcolsep}{2.287pt}
\caption{\textbf{High-resolution visual question answering} on HR-Bench~\cite{wang2025hrbench} 8K and 4K, accuracy (\%) with the VLMEvalKit. TTFT is seconds to first token on 8K images, Speed its speedup over exact attention. Baselines use official settings. \meros results are means over three seeds, with four runs each. $^\ddagger$Rejects native inputs. Its released attention builds a dense mask that FA4 avoids.}
\begin{widewrap}\begin{tabular}{lcccccccccccccccc}
\toprule
\rowcolor{xhead}[\tabcolsep][\rov] & \multicolumn{4}{c}{\bfseries\boldmath Qwen3-VL-8B~\cite{bai2025qwen3vl}, native} & \multicolumn{4}{c}{\bfseries\boldmath Qwen3-VL-4B~\cite{bai2025qwen3vl}, native} & \multicolumn{4}{c}{\bfseries\boldmath GLM-4.6V-Flash~\cite{glm2025glm45v}, native} & \multicolumn{4}{c}{\bfseries\boldmath Kimi-VL-A3B~\cite{kimi2025kimivl}, 4\,MP$^\ddagger$}\\
\hcmid\cmidrule(lr){2-5}\cmidrule(lr){6-9}\cmidrule(lr){10-13}\cmidrule(lr){14-17}
\rowcolor{xhead}[\tabcolsep][\rov] \bfseries\boldmath Method & \bfseries\boldmath 8K$\uparrow$ & \bfseries\boldmath 4K$\uparrow$ & \bfseries\boldmath TTFT$\downarrow$ & \bfseries\boldmath Speed\rlap{$\uparrow$}\phantom{$\downarrow$} & \bfseries\boldmath 8K$\uparrow$ & \bfseries\boldmath 4K$\uparrow$ & \bfseries\boldmath TTFT$\downarrow$ & \bfseries\boldmath Speed\rlap{$\uparrow$}\phantom{$\downarrow$} & \bfseries\boldmath 8K$\uparrow$ & \bfseries\boldmath 4K$\uparrow$ & \bfseries\boldmath TTFT$\downarrow$ & \bfseries\boldmath Speed\rlap{$\uparrow$}\phantom{$\downarrow$} & \bfseries\boldmath 8K$\uparrow$ & \bfseries\boldmath 4K$\uparrow$ & \bfseries\boldmath TTFT$\downarrow$ & \bfseries\boldmath Speed\rlap{$\uparrow$}\phantom{$\downarrow$}\\
\midrule
Exact (FA2) & 67.75 & 73.38 & 12.60 & 1.00$\times$ & 65.12 & 69.62 & 7.99 & 1.00$\times$ & 65.62 & 71.38 & 19.48 & 1.00$\times$ & 55.25 & 58.38 & 3.26 & 1.00$\times$\\
FA4~\cite{zadouri2026flashattention} & 68.00 & 73.25 & 5.01 & 2.51$\times$ & 65.50 & 69.25 & 3.89 & 2.05$\times$ & 65.38 & 71.50 & 8.10 & 2.41$\times$ & 54.75 & 58.00 & 0.38 & 8.50$\times$\\
\midrule
\rowcolor{xsel}[\tabcolsep][\rov] \multicolumn{17}{l}{\textit{\textbf{Visual-token pruning}}}\\
\rowcolor{xsel}[\tabcolsep][\rov] FastV~\cite{chen2024fastv} & 68.12 & 73.50 & 11.52 & 1.09$\times$ & 65.62 & 68.75 & 7.01 & 1.14$\times$ & 65.75 & 72.38 & 17.43 & 1.12$\times$ & 55.50 & 58.25 & 3.14 & 1.04$\times$\\
\rowcolor{xsel}[\tabcolsep][\rov] PDrop~\cite{xing2024pyramiddrop} & 68.88 & 72.75 & 11.53 & 1.09$\times$ & 64.38 & -- & 7.02 & 1.14$\times$ & 65.88 & 72.25 & 17.47 & 1.12$\times$ & 55.25 & 58.00 & 3.14 & 1.04$\times$\\
\rowcolor{xsel}[\tabcolsep][\rov] SparseVLM~\cite{zhang2025sparsevlm} & 68.25 & 73.88 & 11.30 & 1.11$\times$ & 65.12 & \textbf{69.62} & 6.82 & 1.17$\times$ & \textbf{66.50} & 71.88 & 17.01 & 1.15$\times$ & 54.62 & 57.88 & 3.12 & 1.04$\times$\\
\rowcolor{xsel}[\tabcolsep][\rov] VisionZip~\cite{yang2025visionzip} & 68.88 & 72.62 & 12.58 & 1.00$\times$ & 64.88 & -- & 8.03 & 0.99$\times$ & 65.75 & 71.88 & 18.78 & 1.04$\times$ & 54.50 & 58.38 & 3.15 & 1.03$\times$\\
\rowcolor{xsum}[\tabcolsep][\rov] \multicolumn{17}{l}{\textit{\textbf{Inside the vision encoder}}}\\
\rowcolor{xsum}[\tabcolsep][\rov] ToMe~\cite{bolya2023tome} & 66.62 & 71.25 & 16.98 & 0.74$\times$ & 64.75 & -- & -- & -- & 65.75 & -- & -- & -- & 54.00 & -- & -- & --\\
\rowcolor{xsum}[\tabcolsep][\rov] Window attention & 63.87 & 68.38 & 3.81 & 3.30$\times$ & 51.75 & -- & -- & -- & 60.38 & -- & -- & -- & 51.25 & -- & -- & --\\
\midrule
\rowcolor{xmeros}[\tabcolsep][\rov] \meros $r{=}2$ & \textbf{68.91} & 73.76 & 8.16 & 1.54$\times$ & \textbf{65.71} & 69.15 & 4.94 & 1.62$\times$ & 65.92 & \textbf{72.62} & 10.65 & 1.83$\times$ & \textbf{55.93} & \textbf{58.71} & 0.46 & 7.10$\times$\\
\rowcolor{xmeros}[\tabcolsep][\rov] \meros $r{=}4$ & 68.35 & 72.99 & 5.15 & 2.45$\times$ & 64.44 & 68.72 & 3.39 & 2.35$\times$ & 65.77 & 71.95 & 6.88 & 2.83$\times$ & 55.56 & 58.50 & 0.40 & 8.14$\times$\\
\rowcolor{xmeros}[\tabcolsep][\rov] \meros $r{=}8$ & 67.85 & 72.32 & 3.66 & 3.45$\times$ & 64.23 & 68.54 & 2.62 & 3.05$\times$ & 64.58 & 70.38 & 5.47 & 3.57$\times$ & 55.31 & 57.63 & 0.37 & 8.75$\times$\\
\rowcolor{xmeros}[\tabcolsep][\rov] \meros $r{=}8$ + SparseVLM & 68.89 & \textbf{73.89} & \textbf{2.35} & \textbf{5.36}$\times$ & 65.67 & 68.63 & \textbf{1.44} & \textbf{5.54}$\times$ & 65.71 & 71.91 & \textbf{3.02} & \textbf{6.45}$\times$ & 55.57 & 57.81 & \textbf{0.23} & \textbf{14.2}$\times$\\
\bottomrule
\end{tabular}\end{widewrap}
\label{tab:vlm}
\vspace{-1.5em}
\end{table*}

\subsection{Vision-Language Models}
\label{sec:exp_vlm}

In \cref{tab:vlm}, we replace the attention of the vision encoder in four frozen vision-language models (VLMs) and answer HR-Bench~\cite{wang2025hrbench} questions on 8K and 4K images. Qwen3-VL~\cite{bai2025qwen3vl} and GLM-4.6V-Flash~\cite{glm2025glm45v} see native resolution, while Kimi-VL~\cite{kimi2025kimivl} rejects native inputs and gets 4\,MP. In our experiments, FA2 and FA4 compute the same attention, yet their accuracies differ by up to 0.5 points, since a few flipped answers already move the score. Repeated runs of \meros with the same sampling seed may also differ by a few answers. We therefore report its mean over four runs for each of three sampling seeds. The encoder dominates the time to the first token, so pruning visual tokens after the vision encoder gains at most 1.17$\times$, while $r{=}8$ gains 3.1 to 8.8$\times$. For Qwen3-VL-8B, the encoder takes 10.8 of the 12.6\,s with exact attention, and $r{=}8$ runs it in 1.9\,s (\cref{tab:supp_vlm_time}). Pruning acts on the language model and so combines with \meros in the encoder. SparseVLM cuts the prefill from 1.77 to 0.46\,s, and with it \meros is the fastest on every model, at 5.4 to 14.2$\times$.\PFILL

On HR-Bench 8K, $r{=}2$ matches or exceeds exact attention on all models by up to 1.2 points, while the first token arrives up to 7.1$\times$ sooner. With SparseVLM, $r{=}8$ also matches or exceeds exact attention on all four models, whereas $r{=}8$ alone loses at most 1.0 point. On HR-Bench 4K, $r{=}2$ changes the accuracy by $-0.5$ to $+1.2$ points, and $r{=}8$ lowers it by at most 1.1. With $r{=}8$, the first token arrives sooner than with FA4 on every model, at 3.05 to 8.75$\times$ compared with 2.05 to 8.50$\times$. Window attention loses 3.9--13.4 points and is no faster, with an encoder time of 2.05\,s against 1.88\,s for $r{=}8$ on Qwen3-VL-8B. ToMe inside the encoder runs slower than exact attention, at 0.74$\times$ on the same model.

\begin{table}[t]
\begin{minipage}[t]{0.47\textwidth}%
\renewcommand{\mtarget}{\linewidth}
\centering
\tabsetup{6.7}
\renewcommand{\arraystretch}{1.02}
\setlength{\tabcolsep}{1.22pt}
\caption{\textbf{Relative depth at native resolution} on 23 Middlebury and 10 ETH3D images, after scale-and-shift alignment. AbsRel and $\delta_1$ in \%, RMSE in m. Speed is encoder speedup on one H200. Baselines run in their best setting. Bold is best apart from exact. \textit{OOM} means methods that run out of memory.}
\resizebox{\linewidth}{!}{\begin{tabular}{l*{8}{>{\centering\arraybackslash}p{18.8pt}}}
\toprule
\rowcolor{xhead}[\tabcolsep][\rov] & \multicolumn{4}{c}{\bfseries\boldmath Middlebury~\cite{scharstein2014middlebury}} & \multicolumn{4}{c}{\bfseries\boldmath ETH3D~\cite{schops2017eth3d}}\\
\hcmid\cmidrule(lr){2-5}\cmidrule(lr){6-9}
\rowcolor{xhead}[\tabcolsep][\rov] \bfseries\boldmath Method & \makebox[\linewidth][c]{\bfseries\boldmath\fontsize{5.6}{6.72}\selectfont AbsRel$\downarrow$} & \makebox[\linewidth][c]{\bfseries\boldmath\fontsize{5.6}{6.72}\selectfont RMSE$\downarrow$} & \makebox[\linewidth][c]{\bfseries\boldmath\fontsize{5.6}{6.72}\selectfont $\delta_1\!$\rlap{$\uparrow$}\phantom{$\downarrow$}} & \makebox[\linewidth][c]{\bfseries\boldmath\fontsize{5.6}{6.72}\selectfont Speed\rlap{$\uparrow$}\phantom{$\downarrow$}} & \makebox[\linewidth][c]{\bfseries\boldmath\fontsize{5.6}{6.72}\selectfont AbsRel$\downarrow$} & \makebox[\linewidth][c]{\bfseries\boldmath\fontsize{5.6}{6.72}\selectfont RMSE$\downarrow$} & \makebox[\linewidth][c]{\bfseries\boldmath\fontsize{5.6}{6.72}\selectfont $\delta_1\!$\rlap{$\uparrow$}\phantom{$\downarrow$}} & \makebox[\linewidth][c]{\bfseries\boldmath\fontsize{5.6}{6.72}\selectfont Speed\rlap{$\uparrow$}\phantom{$\downarrow$}}\\
\midrule
\rowcolor{xlabel}[\tabcolsep][\rov] \multicolumn{9}{l}{\textit{\textbf{MoGe-3 ViT-G}~\cite{kong2026moge3}}}\\
Exact & 2.61 & 0.106 & 99.86 & 1.00$\times$ & 3.26 & 0.136 & 99.15 & 1.00$\times$\\
\rowcolor{xsel}[\tabcolsep][\rov] NATTEN~\cite{hassani2026gnafindings} & 7.76 & 0.310 & 95.33 & 2.48$\times$ & 23.12 & 1.151 & 58.09 & \textbf{9.43}$\times$\\
\rowcolor{xsel}[\tabcolsep][\rov] XAttention~\cite{xu2025xattention} & 3.48 & 0.131 & 98.93 & 0.95$\times$ & 3.85 & 0.166 & 98.96 & 1.17$\times$\\
\rowcolor{xsum}[\tabcolsep][\rov] i-clustered~\cite{vyas2020clustered} & 5.02 & 0.216 & 96.73 & 0.84$\times$ & 5.76 & 0.226 & 97.50 & 2.79$\times$\\
\rowcolor{xsum}[\tabcolsep][\rov] Expedit~\cite{liang2022expediting} & 6.46 & 0.269 & 92.92 & 1.26$\times$ & \multicolumn{4}{c}{\textit{OOM}}\\
\rowcolor{xapx}[\tabcolsep][\rov] HyperAttn~\cite{han2024hyperattention} & 2.85 & 0.120 & 99.66 & 1.18$\times$ & 3.29 & 0.163 & \textbf{99.18} & 1.29$\times$\\
\rowcolor{xmeros}[\tabcolsep][\rov] \meros $r{=}2$ & \textbf{2.64} & \textbf{0.107} & \textbf{99.86} & 1.86$\times$ & \textbf{3.26} & \textbf{0.136} & 99.15 & 2.39$\times$\\
\rowcolor{xmeros}[\tabcolsep][\rov] \meros $r{=}8$ & 2.69 & 0.109 & 99.63 & \textbf{3.37}$\times$ & 3.29 & \textbf{0.136} & 99.15 & 6.75$\times$\\
\midrule
\rowcolor{xlabel}[\tabcolsep][\rov] \multicolumn{9}{l}{\textit{\textbf{Depth Anything~V2-L}~\cite{yang2024dav2}}}\\
Exact & 4.07 & 0.161 & 99.26 & 1.00$\times$ & 4.64 & 0.222 & 98.30 & 1.00$\times$\\
\rowcolor{xsel}[\tabcolsep][\rov] NATTEN~\cite{hassani2026gnafindings} & 10.31 & 0.375 & 89.49 & 2.60$\times$ & 25.06 & 1.156 & 58.83 & \textbf{10.3}$\times$\\
\rowcolor{xsel}[\tabcolsep][\rov] XAttention~\cite{xu2025xattention} & 4.12 & 0.166 & 98.54 & 0.83$\times$ & 6.54 & 0.305 & 96.14 & 1.00$\times$\\
\rowcolor{xsum}[\tabcolsep][\rov] i-clustered~\cite{vyas2020clustered} & 7.61 & 0.301 & 93.18 & 0.76$\times$ & 8.64 & 0.377 & 91.83 & 2.61$\times$\\
\rowcolor{xsum}[\tabcolsep][\rov] Expedit~\cite{liang2022expediting} & 9.18 & 0.348 & 90.61 & 0.94$\times$ & \multicolumn{4}{c}{\textit{OOM}}\\
\rowcolor{xapx}[\tabcolsep][\rov] HyperAttn~\cite{han2024hyperattention} & 4.97 & 0.192 & 98.39 & 1.19$\times$ & 6.44 & 0.323 & 96.03 & 1.30$\times$\\
\rowcolor{xmeros}[\tabcolsep][\rov] \meros $r{=}2$ & 4.10 & \textbf{0.161} & 99.22 & 1.94$\times$ & \textbf{4.66} & 0.225 & 98.29 & 2.43$\times$\\
\rowcolor{xmeros}[\tabcolsep][\rov] \meros $r{=}8$ & \textbf{4.05} & \textbf{0.161} & \textbf{99.23} & \textbf{3.72}$\times$ & 4.67 & \textbf{0.224} & \textbf{98.39} & 7.33$\times$\\
\bottomrule
\end{tabular}}
\label{tab:depth}
\end{minipage}\hfill
\begin{minipage}[t]{0.51\textwidth}%
\renewcommand{\mtarget}{\linewidth}
\setlength{\extrarowheight}{0.2pt}
\centering
\tabsetup{6.8}
\renewcommand{\arraystretch}{1.094}
\setlength{\tabcolsep}{0.058pt}
\caption{\textbf{Analysis.} (a)~Each far-field estimate reads as many keys as \meros $r{=}8$, with feature error on 12 images and three seeds and $\Delta$mIoU over \mbox{DINOv3-B} and MAE-L on UAVid and Cityscapes. (b)~One component changes at a time. Agree is the share of pixels labeled as by exact attention (\%). Cos is the cosine to exact features.}
\resizebox{\linewidth}{!}{\begin{tabular}{lc>{\hspace{2.6pt}}ccc>{\hspace{2.6pt}}ccc>{\hspace{3.4pt}}c}
\toprule
\rowcolor{xhead}[\tabcolsep][\rov] & \multicolumn{8}{c}{\bfseries\boldmath (a) Same key budget as \meros $r{=}8$}\\
\hcmid\cmidrule(lr){2-9}
\rowcolor{xhead}[\tabcolsep][\rov] & \multicolumn{3}{c}{\bfseries\boldmath DINOv3-B~\cite{simeoni2025dinov3}, error$\downarrow$} & \multicolumn{3}{c}{\bfseries\boldmath MAE-L~\cite{he2022mae}, error$\downarrow$} & \multicolumn{2}{c}{\bfseries\boldmath $\Delta$mIoU$\uparrow$}\\
\hcmid\cmidrule(lr){2-4}\cmidrule(lr){5-7}\cmidrule(lr){8-9}
\rowcolor{xhead}[\tabcolsep][\rov] \bfseries\boldmath Far field & \bfseries\boldmath 1K & \bfseries\boldmath 2K & \bfseries\boldmath 4K & \bfseries\boldmath 1K & \bfseries\boldmath 2K & \bfseries\boldmath 4K & \bfseries\boldmath mean & \bfseries\boldmath worst\\
\midrule
\rowcolor{xsel}[\tabcolsep][\rov] Nearest keys & .488 & .617 & .854 & .238 & .253 & .269 & $-$11.30 & $-$18.13\\
\rowcolor{xsel}[\tabcolsep][\rov] Larger window & .481 & .614 & .853 & .236 & .296 & .269 & $-$11.33 & $-$18.75\\
Strided subset & .418 & .491 & .871 & .198 & .192 & .143 & $-$5.54 & $-$9.01\\
Unweighted sample & .394 & .479 & .874 & .190 & .183 & .133 & $-$4.77 & $-$7.61\\
Cell centroids & .224 & .321 & .539 & .128 & .105 & .090 & $-$0.70 & $-$1.36\\
Weighted grid & .189 & .181 & .194 & .078 & .043 & .023 & $-$0.98 & $-$2.11\\
\rowcolor{xmeros}[\tabcolsep][\rov] \meros & \textbf{.156} & \textbf{.155} & \textbf{.186} & \textbf{.058} & \textbf{.033} & \textbf{.017} & $\mathbf{-0.53}$ & $\mathbf{-1.11}$\\
\midrule
\rowcolor{xhead}[\tabcolsep][\rov] & \multicolumn{8}{c}{\bfseries\boldmath (b) Ablation, DINOv3-B~\cite{simeoni2025dinov3}, $r{=}8$}\\
\hcmid\cmidrule(lr){2-9}
\rowcolor{xhead}[\tabcolsep][\rov] & \multicolumn{3}{c}{\bfseries\boldmath UAVid~\cite{lyu2020uavid}} & \multicolumn{3}{c}{\bfseries\boldmath Cityscapes~\cite{cordts2016cityscapes}} & \multicolumn{2}{c}{\bfseries\boldmath City}\\
\hcmid\cmidrule(lr){2-4}\cmidrule(lr){5-7}\cmidrule(lr){8-9}
\rowcolor{xhead}[\tabcolsep][\rov] \bfseries\boldmath Variant & \bfseries\boldmath error$\downarrow$ & \bfseries\boldmath cos$\uparrow$ & \bfseries\boldmath Agree$\uparrow$ & \bfseries\boldmath error$\downarrow$ & \bfseries\boldmath cos$\uparrow$ & \bfseries\boldmath Agree$\uparrow$ & \multicolumn{2}{c}{\bfseries\boldmath $\Delta$mIoU$\uparrow$}\\
\midrule
Local tile sampled & .253 & .967 & 95.5 & .411 & .912 & 95.1 & \multicolumn{2}{c}{$-$12.64}\\
No weight $r$ & .439 & .903 & 91.5 & .476 & .882 & 96.1 & \multicolumn{2}{c}{$-$7.61}\\
$8{\times}8$ local tile & .171 & .985 & 97.5 & .216 & .976 & 98.4 & \multicolumn{2}{c}{$-$1.91}\\
Fixed grid & .169 & .985 & 97.2 & .201 & .979 & 98.3 & \multicolumn{2}{c}{$-$2.11}\\
\rowcolor{xmeros}[\tabcolsep][\rov] Full \meros & \textbf{.141} & \textbf{.990} & \textbf{98.0} & \textbf{.162} & \textbf{.987} & \textbf{98.9} & \multicolumn{2}{c}{$\mathbf{-1.11}$}\\
\bottomrule
\end{tabular}}
\label{tab:analysis}
\end{minipage}
\end{table}

\subsection{Depth Estimation}
\label{sec:exp_depth}

Depth estimation is a dense regression task, in which every pixel receives a continuous value. Unlike classification, this task has no decision margin, so a systematic error in the features can shift the prediction significantly. We evaluate MoGe-3 ViT-G~\cite{kong2026moge3} and Depth Anything V2-L~\cite{yang2024dav2} on 23 Middlebury~\cite{scharstein2014middlebury} and 10 ETH3D~\cite{schops2017eth3d} images at native resolution (30k/129k tokens), using per-image scale-and-shift-aligned AbsRel, RMSE and $\delta_1$. We time the depth encoder at batch~1, with one warm-up call and three timed calls per image. \meros raises AbsRel by at most 0.08 and lowers $\delta_1$ by at most 0.23 points (\cref{tab:depth}). Among the compared methods, \meros is the most accurate apart from exact attention in every AbsRel and RMSE metric. The regressed depth thus stays stable from 30k to 129k tokens on both models, as expected from a far-field estimate without systematic bias.\PFILL

With Depth Anything V2 on Middlebury, \meros at $r{=}8$ is even more accurate than exact attention and 3.72$\times$ faster. Its AbsRel is 4.05\%, against 4.07\% for exact attention and 4.12 to 10.31\% for the baselines. At $r{=}8$, it runs the encoder 3.4--3.7$\times$ faster on Middlebury and 6.8--7.3$\times$ faster on ETH3D, where only NATTEN is faster, at five to seven times the AbsRel of exact attention. With MoGe-3 on ETH3D, $r{=}2$ reproduces the AbsRel, RMSE and $\delta_1$ of exact attention to the last printed digit at 2.39$\times$ the encoder speed, and $r{=}8$ changes only the AbsRel, from 3.26 to 3.29\%. On Middlebury, XAttention and i-clustered attention run slower than exact attention on both models, and Expedit and HyperAttention gain at most 1.26$\times$. On all 454 ETH3D images, every budget of \meros stays within 0.03 points of the AbsRel of exact attention (\cref{tab:supp_eth3d_full}). Every baseline has a higher AbsRel there than any budget of \meros, from 4.64\% for HyperAttention to 24.74\% for NATTEN. The $\delta_1$ of \meros differs from that of exact attention by at most 0.05 points and its RMSE by at most 2\,mm. On the ETH3D scenes of \cref{fig:qual}(a,b), \meros keeps depth details such as thin pipes and table legs, and AbsRel and Chamfer distance stay within 0.06 points of exact attention.

\begin{figure}[t]
\centering
\includegraphics[width=\textwidth,height=0.80\textheight,keepaspectratio]{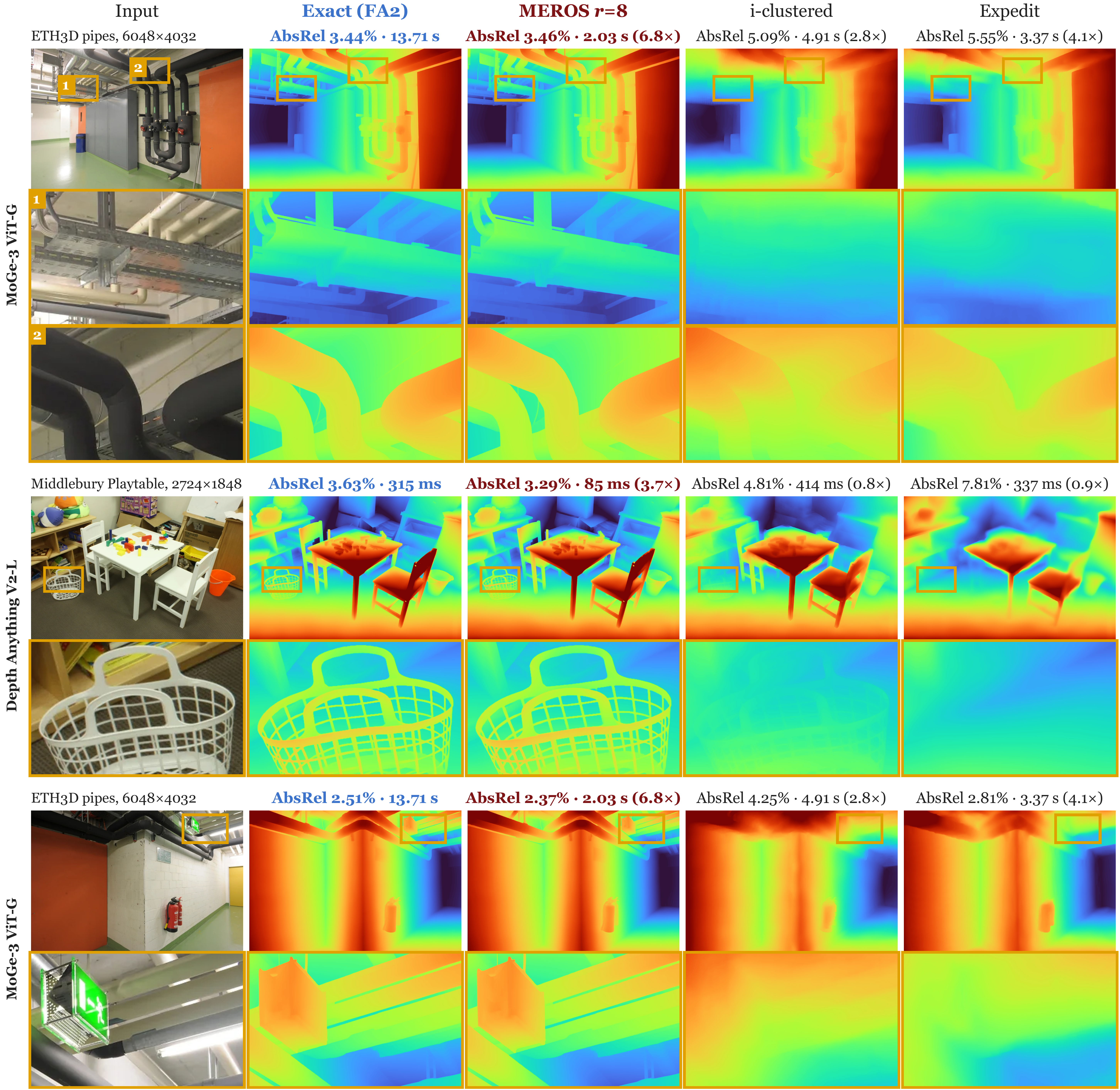}
\caption{\textbf{Monocular depth estimation at native resolution. Lower AbsRel and a shorter time are better.} MoGe-3 ViT-G~\cite{kong2026moge3} on two ETH3D~\cite{schops2017eth3d} frames (top, bottom) of 129k tokens, and Depth Anything V2-L~\cite{yang2024dav2} on a Middlebury image~\cite{scharstein2014middlebury} (middle). Exact attention runs with FlashAttention-2 (FA2). We compare it with \meros $r{=}8$, i-clustered~\cite{vyas2020clustered} in its setting of \cref{tab:depth} and Expedit~\cite{liang2022expediting}, which runs its setting of \cref{tab:depth} on Middlebury and its aggressive setting on ETH3D, where the default runs out of memory. Labels give the AbsRel of the image and the mean encoder time over the images of its dataset, with the speedup over FA2.}
\label{fig:depth_eth3d}
\vspace{-0.4em}
\end{figure}

\begin{figure}[p]
\centering
\includegraphics[width=\textwidth,height=606pt,keepaspectratio]{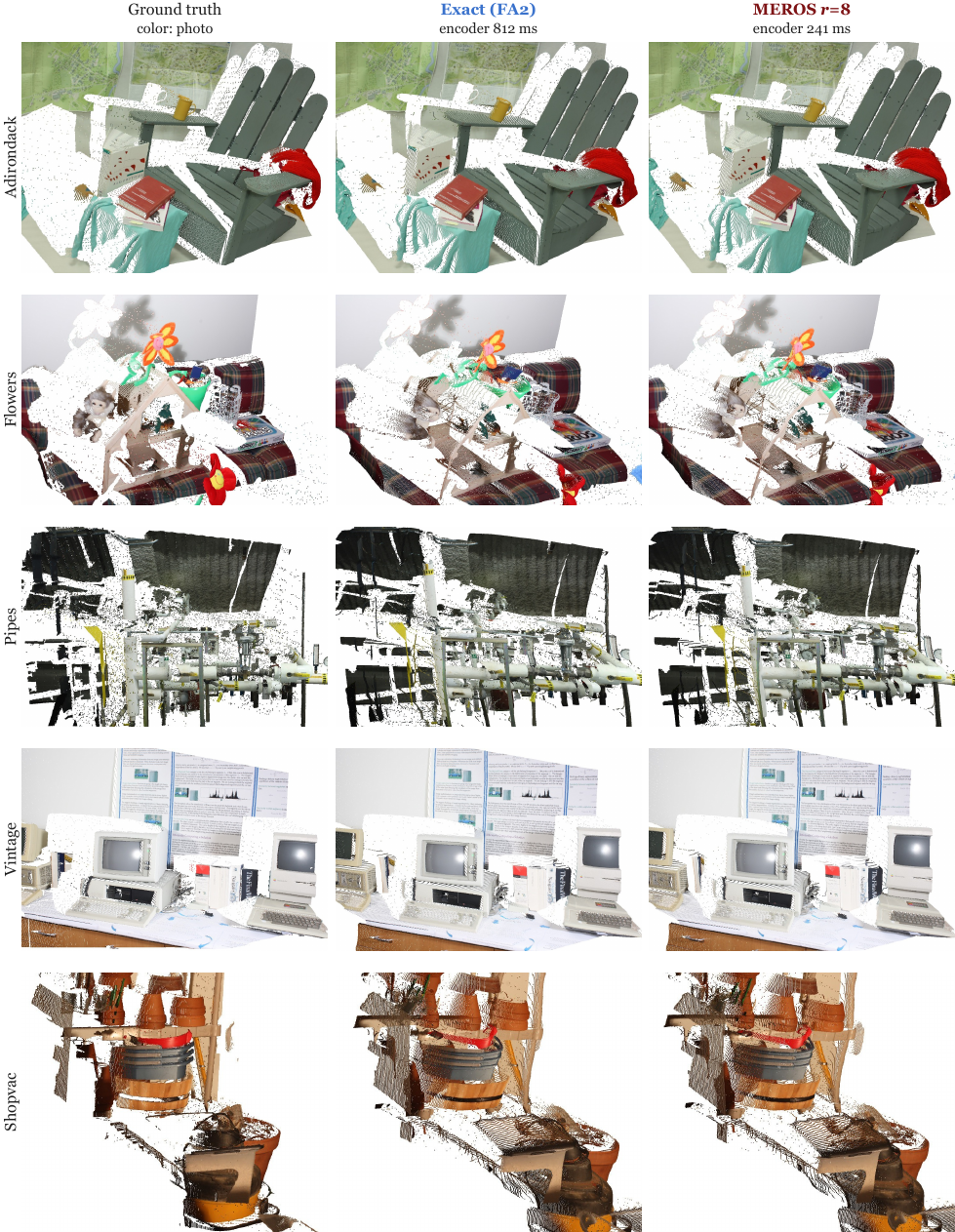}
\caption{\textbf{Point maps of MoGe-3 ViT-G~\cite{kong2026moge3} on Middlebury~\cite{scharstein2014middlebury}. A shorter encoder time is better.} The predicted depth of five scenes is back-projected with the ground-truth intrinsics and viewed obliquely. Each row shows the ground truth, exact attention (FA2) and \meros $r{=}8$ at native resolution, all colored by the photo. The headers give the mean encoder time, as in \cref{fig:depth_eth3d}. With \meros in all 40 encoder layers, walls stay flat and thin parts stay in place, at a 3.4$\times$ shorter encoder time.}
\label{fig:supp_pointmaps}
\end{figure}

\Cref{fig:depth_eth3d} shows three images, and on two of them \meros has a lower AbsRel than exact attention at a 3.7 to 6.8$\times$ faster encoder. On the first ETH3D frame, \meros $r{=}8$ moves the AbsRel of MoGe-3 from 3.44 to 3.46\% at a 6.8$\times$ faster encoder than FA2, and the pipes, brackets and cable trays under the ceiling keep their edges. The i-clustered baseline blurs them at an AbsRel of 5.09\% and a 2.8$\times$ speedup, and the aggressive setting of Expedit at 5.55\% and 4.1$\times$. On the Middlebury image and the second ETH3D frame, \meros keeps the lattice of a basket and an exit sign with the pipes behind it. Both baselines flatten them at a higher AbsRel and a longer encoder time. On the Middlebury image, the encoder of Depth Anything V2 takes 85\,ms with \meros and 315\,ms with FA2, while the two baselines need 414 and 337\,ms. Its AbsRel there is 3.29\% with \meros, below the 3.63\% of exact attention and the 4.81 and 7.81\% of the baselines. On the second ETH3D frame, \meros reaches 2.37\% in 2.03\,s, where exact attention reaches 2.51\% in 13.71\,s and the baselines 4.25 and 2.81\% in 4.91 and 3.37\,s. The gain grows with the token count, from an encoder speedup of 3.7$\times$ at 30k tokens to 6.8$\times$ at 129k. Both baselines are also slower than \meros on every image, by 1.7 to 4.9$\times$. \Cref{fig:supp_pointmaps} back-projects the depth of MoGe-3 into point maps of five Middlebury scenes, where the encoder takes 241 instead of 812\,ms. With \meros in all 40 encoder layers, walls stay flat and thin parts such as pipes and chair legs stay in place.\PUP

\subsection{Analysis}
\label{sec:exp_ana}

We finally analyze which design choices of \meros matter for its accuracy. In \cref{tab:analysis}a, we give every far-field estimate the key budget of $r{=}8$ and change only the estimate, and \cref{tab:analysis}b changes one component at a time. Spending the budget on the nearest keys or a larger window, as selection does, raises the feature error 3.1 to 15.8$\times$ and costs 11.3~mIoU on average. Averaging each cell, dropping the weight $r$ or keeping a strided subset raises it 1.4 to 8.3$\times$. A weighted fixed grid comes closest, yet its worst loss is almost twice that of \meros. Sampling the local tile costs 12.6~mIoU on Cityscapes, and dropping the weight $r$ triples the feature error. Full \meros labels 98 to 99\% of the pixels as exact attention does, and a smaller $8{\times}8$ tile or a fixed grid each loses about one more mIoU on Cityscapes. On MAE-L, its feature error nearly halves with each doubling of the side, from .058 at 1K to .017 at 4K.\PUP

\clearpage
\begin{figure}[t]
\centering
\includegraphics[width=\textwidth]{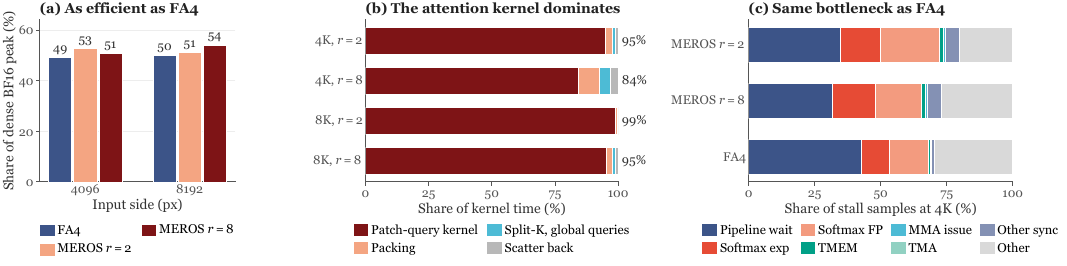}\par\vspace{-3.5pt}
\caption{\textbf{Hardware profile on one B300}, \mbox{DINOv3}-B layer~6 with its real queries, keys and values, batch~1. (a)~Throughput of the patch-query kernel as a share of the dense BF16 peak of 2.25~PFLOP/s, counting $4d$ FLOP for each query--key pair that the operator needs. (b)~GPU time of one \meros call, split by kernel. The numbers on the right give the share of the patch-query kernel. (c)~Warp stall samples of Nsight Compute, grouped by the instruction on which a warp waits, for \meros at $r{=}2$ and $r{=}8$ and for FA4.}
\label{fig:supp_kernel}
\end{figure}

\section{MEROS in Practice}
\label{sec:practice}

We explain three practical questions, namely how \meros enters a ViT, how fast it runs and how accurate it stays.

\subsection{Deployment}
\label{sec:practice_deploy}

\meros needs \textbf{no} training, \textbf{no} calibration data and \textbf{no} change to the weights. The only part of a released model that changes is the call of its global attention layers, in ViTs, ViTMatte, SAM~\cite{kirillov2023sam} and four VLM encoders alike.

\begin{plist}
\item \textbf{Layers.} \meros replaces every global attention layer and leaves windowed layers, such as those of ViTMatte and SAM, unchanged. In ViTMatte, these are four of the 12 blocks. In the four VLMs, only the vision encoder is changed. \Cref{tab:supp_models} lists the layers, position terms and kernel of each model in this report.

\item \textbf{Position terms.} \meros operates on queries and keys after the model's own positional processing, such as the rotary position embeddings (RoPE)~\cite{su2024roformersupp} of \mbox{DINOv3}~\cite{simeoni2025dinov3}. A sampled key keeps the vector computed at its own position, so no position is re-encoded and every value stays paired with its own key. For the decomposed relative bias of ViTMatte and SAM, the kernel gathers the bias of exactly the keys it reads. One implementation covers all four kinds of position terms we meet, which are absolute embeddings, RoPE, 2D RoPE and relative biases.

\item \textbf{Inputs and outputs.} It handles patches of 14 and 16\,px, head dimensions from 64 to 128 and zero to five global tokens. Other grids are padded to a multiple of 16 patches, and the outputs are cropped back to the input size after the head. The operator returns its output in the shape and token order of the attention it replaces, so the linear heads trained on exact features for segmentation and DIS are reused without retraining.\PFILL

\item \textbf{The sample.} The far keys are chosen once per layer and head before inference, as patch positions cached per resolution as $N/r$ integers per head. For $r{=}8$, 4 and 2, the samples are nested in one another.
\end{plist}

As the rest of the model is left untouched, methods that act elsewhere can be added on top of \meros, for example token pruning in the language model. Combined with SparseVLM, $r{=}8$ produces the first token 5.4 to 14.2$\times$ sooner on the four VLMs, and on HR-Bench 8K it matches or exceeds exact attention on all four. On its own, pruning gains at most 1.17$\times$, since the encoder takes 10.8 of the 12.6\,s to the first token of Qwen3-VL-8B.\PFILL

\subsection{Speed}
\label{sec:practice_speed}

\begin{table*}[t]
\centering
\caption{\textbf{Whole-backbone speedup of \meros} on 14 frozen ViTs, timed end to end on one B300 at batch~1. Speedups are over FA2 in (a) and over FA4 in (b). The sizes 1K, 2K, 4K and 8K stand for 1024, 2048, 4096 and 8192\,px for the patch-16 models and 1120, 2240, 4256 and 8288\,px for the patch-14 models. On every backbone, the speedup over FA2 grows with the input size.}
\label{tab:supp_backbones}
\tabsetup{8.2}
\renewcommand{\arraystretch}{1.0}
\setlength{\tabcolsep}{2.0pt}
\setlength{\cmidrulewidth}{\lightrulewidth}
\setlength{\tabw}{\dimexpr\textwidth-2\tabcolsep\relax}%
\begin{widewrap}\begin{tabular}{p{0.125\tabw}>{\centering\arraybackslash}p{0.045\tabw}*{12}{>{\centering\arraybackslash}p{\dimexpr(0.83\tabw-26\tabcolsep)/12\relax}}}
\toprule
\rowcolor{tabhead}[\tabcolsep][\rov] & & \multicolumn{3}{c}{\bfseries 1K} & \multicolumn{3}{c}{\bfseries 2K} & \multicolumn{3}{c}{\bfseries 4K} & \multicolumn{3}{c}{\bfseries 8K}\\
\hcmid\cmidrule(lr){3-5}\cmidrule(lr){6-8}\cmidrule(lr){9-11}\cmidrule(lr){12-14}
\rowcolor{tabhead}[\tabcolsep][\rov] \bfseries Backbone & \bfseries Patch & \bfseries\boldmath $r{=}2$ & \bfseries\boldmath $r{=}4$ & \bfseries\boldmath $r{=}8$ & \bfseries\boldmath $r{=}2$ & \bfseries\boldmath $r{=}4$ & \bfseries\boldmath $r{=}8$ & \bfseries\boldmath $r{=}2$ & \bfseries\boldmath $r{=}4$ & \bfseries\boldmath $r{=}8$ & \bfseries\boldmath $r{=}2$ & \bfseries\boldmath $r{=}4$ & \bfseries\boldmath $r{=}8$\\
\midrule
\rowcolor{tabhead}[\tabcolsep][\rov] \multicolumn{14}{l}{\bfseries\boldmath (a) Speed of \meros over exact attention (FA2)\rlap{$\uparrow$}\phantom{$\downarrow$}}\\
MAE-B~\cite{he2022mae} & 16 & \cellcolor[HTML]{FEEDE0}1.30$\times$ & \cellcolor[HTML]{FEEBDE}1.38$\times$ & \cellcolor[HTML]{FEEADB}1.46$\times$ & \cellcolor[HTML]{FCD4B8}2.84$\times$ & \cellcolor[HTML]{FBCBAA}3.49$\times$ & \cellcolor[HTML]{FBC6A4}3.87$\times$ & \cellcolor[HTML]{F9BA92}4.93$\times$ & \cellcolor[HTML]{F59E69}7.97$\times$ & \cellcolor[HTML]{F0894F}11.2$\times$ & \cellcolor[HTML]{F8B287}5.65$\times$ & \cellcolor[HTML]{F08A50}11.0$\times$ & \cellcolor[HTML]{E1672F}19.4$\times$\\
MAE-L~\cite{he2022mae} & 16 & \cellcolor[HTML]{FEEADB}1.46$\times$ & \cellcolor[HTML]{FEE8D8}1.56$\times$ & \cellcolor[HTML]{FEE7D6}1.65$\times$ & \cellcolor[HTML]{FCD1B4}3.03$\times$ & \cellcolor[HTML]{FBC7A5}3.82$\times$ & \cellcolor[HTML]{FAC29E}4.25$\times$ & \cellcolor[HTML]{F9BB93}4.88$\times$ & \cellcolor[HTML]{F59E6A}7.88$\times$ & \cellcolor[HTML]{F0894F}11.1$\times$ & \cellcolor[HTML]{F8B286}5.71$\times$ & \cellcolor[HTML]{F08A50}11.0$\times$ & \cellcolor[HTML]{E26830}19.1$\times$\\
DINOv3-B~\cite{simeoni2025dinov3} & 16 & \cellcolor[HTML]{FCDAC2}2.43$\times$ & \cellcolor[HTML]{FCD5BA}2.75$\times$ & \cellcolor[HTML]{FCD2B6}2.93$\times$ & \cellcolor[HTML]{FBC7A5}3.79$\times$ & \cellcolor[HTML]{FABD95}4.75$\times$ & \cellcolor[HTML]{F9B68B}5.35$\times$ & \cellcolor[HTML]{F9B78E}5.20$\times$ & \cellcolor[HTML]{F59B66}8.28$\times$ & \cellcolor[HTML]{F0864C}11.6$\times$ & \cellcolor[HTML]{F8B185}5.75$\times$ & \cellcolor[HTML]{F0894F}11.2$\times$ & \cellcolor[HTML]{E1672F}19.4$\times$\\
DINOv3-L~\cite{simeoni2025dinov3} & 16 & \cellcolor[HTML]{FCD7BE}2.59$\times$ & \cellcolor[HTML]{FCD4B9}2.80$\times$ & \cellcolor[HTML]{FCD1B5}2.98$\times$ & \cellcolor[HTML]{FBC7A5}3.77$\times$ & \cellcolor[HTML]{FABD96}4.73$\times$ & \cellcolor[HTML]{F9B68C}5.31$\times$ & \cellcolor[HTML]{F9B98F}5.09$\times$ & \cellcolor[HTML]{F59D68}8.13$\times$ & \cellcolor[HTML]{F0884E}11.3$\times$ & \cellcolor[HTML]{F8B185}5.79$\times$ & \cellcolor[HTML]{F0894F}11.1$\times$ & \cellcolor[HTML]{E26830}19.1$\times$\\
DINOv2-B~\cite{oquab2024dinov2} & 14 & \cellcolor[HTML]{FEE9DA}1.50$\times$ & \cellcolor[HTML]{FEE7D7}1.60$\times$ & \cellcolor[HTML]{FEE7D6}1.65$\times$ & \cellcolor[HTML]{FBCBAB}3.47$\times$ & \cellcolor[HTML]{FABF99}4.57$\times$ & \cellcolor[HTML]{F9B58B}5.37$\times$ & \cellcolor[HTML]{F9B990}5.07$\times$ & \cellcolor[HTML]{F59C67}8.24$\times$ & \cellcolor[HTML]{EF8348}12.3$\times$ & \cellcolor[HTML]{F8B287}5.65$\times$ & \cellcolor[HTML]{F08A50}11.0$\times$ & \cellcolor[HTML]{E0652D}20.2$\times$\\
DINOv2-L~\cite{oquab2024dinov2} & 14 & \cellcolor[HTML]{FEE6D5}1.67$\times$ & \cellcolor[HTML]{FDE4D2}1.82$\times$ & \cellcolor[HTML]{FDE3D0}1.89$\times$ & \cellcolor[HTML]{FBCAAA}3.52$\times$ & \cellcolor[HTML]{FABD96}4.73$\times$ & \cellcolor[HTML]{F8B387}5.62$\times$ & \cellcolor[HTML]{F9B88F}5.11$\times$ & \cellcolor[HTML]{F59B66}8.32$\times$ & \cellcolor[HTML]{EF8247}12.4$\times$ & \cellcolor[HTML]{F8B286}5.67$\times$ & \cellcolor[HTML]{F0894F}11.2$\times$ & \cellcolor[HTML]{E0652D}20.1$\times$\\
DeiT\,III-B~\cite{touvron2022deit3} & 16 & \cellcolor[HTML]{FEE8D9}1.55$\times$ & \cellcolor[HTML]{FDE5D3}1.78$\times$ & \cellcolor[HTML]{FDE3D0}1.91$\times$ & \cellcolor[HTML]{FCD1B3}3.04$\times$ & \cellcolor[HTML]{FBC7A4}3.83$\times$ & \cellcolor[HTML]{FAC29C}4.33$\times$ & \cellcolor[HTML]{F9BC95}4.78$\times$ & \cellcolor[HTML]{F6A06D}7.62$\times$ & \cellcolor[HTML]{F18C53}10.6$\times$ & \cellcolor[HTML]{F8B388}5.57$\times$ & \cellcolor[HTML]{F18B52}10.8$\times$ & \cellcolor[HTML]{E26A31}18.7$\times$\\
DeiT\,III-L~\cite{touvron2022deit3} & 16 & \cellcolor[HTML]{FEE6D5}1.67$\times$ & \cellcolor[HTML]{FDE3D1}1.86$\times$ & \cellcolor[HTML]{FDE2CE}1.97$\times$ & \cellcolor[HTML]{FCD1B3}3.04$\times$ & \cellcolor[HTML]{FBC7A4}3.85$\times$ & \cellcolor[HTML]{FAC29D}4.32$\times$ & \cellcolor[HTML]{FABD96}4.72$\times$ & \cellcolor[HTML]{F6A16E}7.58$\times$ & \cellcolor[HTML]{F18C53}10.6$\times$ & \cellcolor[HTML]{F8B185}5.77$\times$ & \cellcolor[HTML]{F0894F}11.1$\times$ & \cellcolor[HTML]{E26830}19.2$\times$\\
EVA-02-B~\cite{fang2024eva02} & 14 & \cellcolor[HTML]{FEE6D5}1.70$\times$ & \cellcolor[HTML]{FDE4D1}1.83$\times$ & \cellcolor[HTML]{FDE3D0}1.90$\times$ & \cellcolor[HTML]{FCD3B6}2.91$\times$ & \cellcolor[HTML]{FBCAA9}3.56$\times$ & \cellcolor[HTML]{FAC5A2}3.98$\times$ & \cellcolor[HTML]{FAC09A}4.51$\times$ & \cellcolor[HTML]{F7A878}6.72$\times$ & \cellcolor[HTML]{F49660}9.03$\times$ & \cellcolor[HTML]{F8B58B}5.39$\times$ & \cellcolor[HTML]{F29058}10.0$\times$ & \cellcolor[HTML]{E67036}16.9$\times$\\
EVA-02-L~\cite{fang2024eva02} & 14 & \cellcolor[HTML]{FEEBDD}1.40$\times$ & \cellcolor[HTML]{FEEADB}1.47$\times$ & \cellcolor[HTML]{FEE9DA}1.50$\times$ & \cellcolor[HTML]{FDDDC6}2.28$\times$ & \cellcolor[HTML]{FCD8BE}2.58$\times$ & \cellcolor[HTML]{FCD5BA}2.74$\times$ & \cellcolor[HTML]{FBC6A3}3.89$\times$ & \cellcolor[HTML]{F9B68C}5.30$\times$ & \cellcolor[HTML]{F7AB7C}6.41$\times$ & \cellcolor[HTML]{F9B78E}5.20$\times$ & \cellcolor[HTML]{F3955F}9.14$\times$ & \cellcolor[HTML]{EC7B40}13.9$\times$\\
CLIP-B~\cite{radford2021clip} & 16 & \cellcolor[HTML]{FEEDE0}1.30$\times$ & \cellcolor[HTML]{FEEBDE}1.39$\times$ & \cellcolor[HTML]{FEEADC}1.45$\times$ & \cellcolor[HTML]{FCD5BA}2.77$\times$ & \cellcolor[HTML]{FBCCAD}3.38$\times$ & \cellcolor[HTML]{FBC8A6}3.73$\times$ & \cellcolor[HTML]{FABD96}4.72$\times$ & \cellcolor[HTML]{F6A16F}7.49$\times$ & \cellcolor[HTML]{F18D55}10.4$\times$ & \cellcolor[HTML]{F8B388}5.54$\times$ & \cellcolor[HTML]{F18B52}10.8$\times$ & \cellcolor[HTML]{E26A31}18.7$\times$\\
CLIP-L~\cite{radford2021clip} & 14 & \cellcolor[HTML]{FDE5D3}1.76$\times$ & \cellcolor[HTML]{FDE2CF}1.94$\times$ & \cellcolor[HTML]{FDE1CD}2.03$\times$ & \cellcolor[HTML]{FBCBAB}3.48$\times$ & \cellcolor[HTML]{FABE97}4.67$\times$ & \cellcolor[HTML]{F8B388}5.56$\times$ & \cellcolor[HTML]{F9B990}5.06$\times$ & \cellcolor[HTML]{F59C67}8.25$\times$ & \cellcolor[HTML]{EF8348}12.2$\times$ & \cellcolor[HTML]{F8B387}5.63$\times$ & \cellcolor[HTML]{F0894F}11.1$\times$ & \cellcolor[HTML]{E1662E}19.8$\times$\\
SigLIP2-B~\cite{tschannen2025siglip2} & 16 & \cellcolor[HTML]{FEEDE1}1.28$\times$ & \cellcolor[HTML]{FEECDF}1.35$\times$ & \cellcolor[HTML]{FEEBDD}1.40$\times$ & \cellcolor[HTML]{FCD7BD}2.63$\times$ & \cellcolor[HTML]{FCD0B3}3.06$\times$ & \cellcolor[HTML]{FBCDAD}3.34$\times$ & \cellcolor[HTML]{FABE97}4.68$\times$ & \cellcolor[HTML]{F6A573}7.11$\times$ & \cellcolor[HTML]{F3925B}9.58$\times$ & \cellcolor[HTML]{F8B489}5.52$\times$ & \cellcolor[HTML]{F18D54}10.5$\times$ & \cellcolor[HTML]{E46C33}18.0$\times$\\
SigLIP2-L~\cite{tschannen2025siglip2} & 16 & \cellcolor[HTML]{FEEBDD}1.40$\times$ & \cellcolor[HTML]{FEEADB}1.48$\times$ & \cellcolor[HTML]{FEE9D9}1.53$\times$ & \cellcolor[HTML]{FCD4B9}2.78$\times$ & \cellcolor[HTML]{FBCDAF}3.28$\times$ & \cellcolor[HTML]{FBCBAA}3.50$\times$ & \cellcolor[HTML]{FABD96}4.72$\times$ & \cellcolor[HTML]{F6A472}7.19$\times$ & \cellcolor[HTML]{F2915A}9.75$\times$ & \cellcolor[HTML]{F8B287}5.65$\times$ & \cellcolor[HTML]{F18B52}10.7$\times$ & \cellcolor[HTML]{E36C33}18.1$\times$\\
\midrule
\rowcolor{tabhead}[\tabcolsep][\rov] \multicolumn{14}{l}{\bfseries\boldmath (b) Speed of \meros over FA4~\cite{zadouri2026flashattention}\rlap{$\uparrow$}\phantom{$\downarrow$}}\\
MAE-B~\cite{he2022mae} & 16 & \cellcolor[HTML]{FFF2E9}1.07$\times$ & \cellcolor[HTML]{FFF1E7}1.13$\times$ & \cellcolor[HTML]{FFEFE4}1.19$\times$ & \cellcolor[HTML]{FEECDF}1.35$\times$ & \cellcolor[HTML]{FEE7D6}1.66$\times$ & \cellcolor[HTML]{FDE4D1}1.84$\times$ & \cellcolor[HTML]{FDE5D3}1.78$\times$ & \cellcolor[HTML]{FCD3B7}2.88$\times$ & \cellcolor[HTML]{FAC4A1}4.05$\times$ & \cellcolor[HTML]{FDE2CE}1.97$\times$ & \cellcolor[HTML]{FBC7A4}3.84$\times$ & \cellcolor[HTML]{F7A777}6.77$\times$\\
MAE-L~\cite{he2022mae} & 16 & \cellcolor[HTML]{FFF3EA}1.05$\times$ & \cellcolor[HTML]{FFF1E7}1.12$\times$ & \cellcolor[HTML]{FFEFE4}1.19$\times$ & \cellcolor[HTML]{FEEADC}1.43$\times$ & \cellcolor[HTML]{FDE4D2}1.80$\times$ & \cellcolor[HTML]{FDE2CE}2.00$\times$ & \cellcolor[HTML]{FDE5D2}1.79$\times$ & \cellcolor[HTML]{FCD3B7}2.89$\times$ & \cellcolor[HTML]{FAC4A1}4.06$\times$ & \cellcolor[HTML]{FDE2CE}1.99$\times$ & \cellcolor[HTML]{FBC7A4}3.85$\times$ & \cellcolor[HTML]{F7A879}6.67$\times$\\
DINOv3-B~\cite{simeoni2025dinov3} & 16 & \cellcolor[HTML]{FDDEC8}2.21$\times$ & \cellcolor[HTML]{FCD9C1}2.49$\times$ & \cellcolor[HTML]{FCD6BC}2.65$\times$ & \cellcolor[HTML]{FDE0CA}2.14$\times$ & \cellcolor[HTML]{FCD6BC}2.68$\times$ & \cellcolor[HTML]{FCD1B4}3.03$\times$ & \cellcolor[HTML]{FDE1CC}2.07$\times$ & \cellcolor[HTML]{FBCDAE}3.30$\times$ & \cellcolor[HTML]{FABF98}4.60$\times$ & \cellcolor[HTML]{FDE1CC}2.05$\times$ & \cellcolor[HTML]{FAC5A2}3.98$\times$ & \cellcolor[HTML]{F6A675}6.92$\times$\\
DINOv3-L~\cite{simeoni2025dinov3} & 16 & \cellcolor[HTML]{FDE1CD}2.02$\times$ & \cellcolor[HTML]{FDDFC9}2.18$\times$ & \cellcolor[HTML]{FDDCC5}2.32$\times$ & \cellcolor[HTML]{FDDFC9}2.17$\times$ & \cellcolor[HTML]{FCD5BB}2.72$\times$ & \cellcolor[HTML]{FCD1B3}3.05$\times$ & \cellcolor[HTML]{FDE1CD}2.04$\times$ & \cellcolor[HTML]{FBCEAF}3.25$\times$ & \cellcolor[HTML]{FAC09A}4.52$\times$ & \cellcolor[HTML]{FDE1CC}2.08$\times$ & \cellcolor[HTML]{FAC5A2}4.00$\times$ & \cellcolor[HTML]{F6A776}6.86$\times$\\
DINOv2-B~\cite{oquab2024dinov2} & 14 & \cellcolor[HTML]{FFF1E8}1.10$\times$ & \cellcolor[HTML]{FFF0E5}1.17$\times$ & \cellcolor[HTML]{FEEFE4}1.21$\times$ & \cellcolor[HTML]{FEE9DA}1.49$\times$ & \cellcolor[HTML]{FDE2CE}1.97$\times$ & \cellcolor[HTML]{FDDCC5}2.31$\times$ & \cellcolor[HTML]{FDE3D0}1.88$\times$ & \cellcolor[HTML]{FCD0B3}3.06$\times$ & \cellcolor[HTML]{FABF99}4.56$\times$ & \cellcolor[HTML]{FDE2CE}1.98$\times$ & \cellcolor[HTML]{FBC6A4}3.86$\times$ & \cellcolor[HTML]{F6A574}7.07$\times$\\
DINOv2-L~\cite{oquab2024dinov2} & 14 & \cellcolor[HTML]{FFF2E9}1.08$\times$ & \cellcolor[HTML]{FFF0E5}1.18$\times$ & \cellcolor[HTML]{FEEEE3}1.23$\times$ & \cellcolor[HTML]{FEEADB}1.48$\times$ & \cellcolor[HTML]{FDE2CE}1.99$\times$ & \cellcolor[HTML]{FDDBC4}2.36$\times$ & \cellcolor[HTML]{FDE3D0}1.88$\times$ & \cellcolor[HTML]{FCD0B3}3.06$\times$ & \cellcolor[HTML]{FABF98}4.58$\times$ & \cellcolor[HTML]{FDE2CE}2.00$\times$ & \cellcolor[HTML]{FBC6A3}3.94$\times$ & \cellcolor[HTML]{F6A573}7.09$\times$\\
DeiT\,III-B~\cite{touvron2022deit3} & 16 & \cellcolor[HTML]{FFF0E5}1.17$\times$ & \cellcolor[HTML]{FEECDF}1.34$\times$ & \cellcolor[HTML]{FEEADC}1.44$\times$ & \cellcolor[HTML]{FEEBDE}1.39$\times$ & \cellcolor[HTML]{FDE5D3}1.76$\times$ & \cellcolor[HTML]{FDE2CE}1.98$\times$ & \cellcolor[HTML]{FDE5D2}1.79$\times$ & \cellcolor[HTML]{FCD3B8}2.85$\times$ & \cellcolor[HTML]{FAC5A2}3.98$\times$ & \cellcolor[HTML]{FDE2CE}1.96$\times$ & \cellcolor[HTML]{FBC7A5}3.81$\times$ & \cellcolor[HTML]{F7A979}6.61$\times$\\
DeiT\,III-L~\cite{touvron2022deit3} & 16 & \cellcolor[HTML]{FFF2E9}1.08$\times$ & \cellcolor[HTML]{FEEFE4}1.21$\times$ & \cellcolor[HTML]{FEEEE2}1.27$\times$ & \cellcolor[HTML]{FEEBDD}1.42$\times$ & \cellcolor[HTML]{FDE4D2}1.80$\times$ & \cellcolor[HTML]{FDE1CD}2.02$\times$ & \cellcolor[HTML]{FDE5D3}1.77$\times$ & \cellcolor[HTML]{FCD3B8}2.85$\times$ & \cellcolor[HTML]{FAC5A2}3.97$\times$ & \cellcolor[HTML]{FDE2CE}2.00$\times$ & \cellcolor[HTML]{FBC6A4}3.86$\times$ & \cellcolor[HTML]{F7A979}6.64$\times$\\
EVA-02-B~\cite{fang2024eva02} & 14 & \cellcolor[HTML]{FFEFE4}1.20$\times$ & \cellcolor[HTML]{FEEDE1}1.29$\times$ & \cellcolor[HTML]{FEECDF}1.34$\times$ & \cellcolor[HTML]{FEEBDE}1.39$\times$ & \cellcolor[HTML]{FEE6D5}1.70$\times$ & \cellcolor[HTML]{FDE3D0}1.89$\times$ & \cellcolor[HTML]{FDE5D3}1.76$\times$ & \cellcolor[HTML]{FCD7BD}2.63$\times$ & \cellcolor[HTML]{FBCAAA}3.53$\times$ & \cellcolor[HTML]{FDE2CF}1.94$\times$ & \cellcolor[HTML]{FBC9A8}3.60$\times$ & \cellcolor[HTML]{F7AE80}6.08$\times$\\
EVA-02-L~\cite{fang2024eva02} & 14 & \cellcolor[HTML]{FFF2E9}1.07$\times$ & \cellcolor[HTML]{FFF1E7}1.12$\times$ & \cellcolor[HTML]{FFF1E6}1.14$\times$ & \cellcolor[HTML]{FEEEE3}1.23$\times$ & \cellcolor[HTML]{FEEBDE}1.39$\times$ & \cellcolor[HTML]{FEEADB}1.48$\times$ & \cellcolor[HTML]{FEE7D7}1.61$\times$ & \cellcolor[HTML]{FDDEC9}2.20$\times$ & \cellcolor[HTML]{FCD6BC}2.65$\times$ & \cellcolor[HTML]{FDE3D1}1.86$\times$ & \cellcolor[HTML]{FBCEAF}3.27$\times$ & \cellcolor[HTML]{F9BA91}4.99$\times$\\
CLIP-B~\cite{radford2021clip} & 16 & \cellcolor[HTML]{FFF2E9}1.07$\times$ & \cellcolor[HTML]{FFF1E6}1.14$\times$ & \cellcolor[HTML]{FFEFE4}1.19$\times$ & \cellcolor[HTML]{FEECDF}1.34$\times$ & \cellcolor[HTML]{FEE7D6}1.64$\times$ & \cellcolor[HTML]{FDE4D2}1.81$\times$ & \cellcolor[HTML]{FDE5D3}1.78$\times$ & \cellcolor[HTML]{FCD4B8}2.82$\times$ & \cellcolor[HTML]{FBC6A3}3.93$\times$ & \cellcolor[HTML]{FDE2CE}1.97$\times$ & \cellcolor[HTML]{FBC7A4}3.83$\times$ & \cellcolor[HTML]{F7A879}6.66$\times$\\
CLIP-L~\cite{radford2021clip} & 14 & \cellcolor[HTML]{FFF2E8}1.09$\times$ & \cellcolor[HTML]{FEEFE4}1.21$\times$ & \cellcolor[HTML]{FEEEE2}1.26$\times$ & \cellcolor[HTML]{FEEADB}1.48$\times$ & \cellcolor[HTML]{FDE2CE}1.99$\times$ & \cellcolor[HTML]{FDDBC4}2.37$\times$ & \cellcolor[HTML]{FDE3D1}1.87$\times$ & \cellcolor[HTML]{FCD1B3}3.05$\times$ & \cellcolor[HTML]{FAC09A}4.52$\times$ & \cellcolor[HTML]{FDE2CE}1.99$\times$ & \cellcolor[HTML]{FBC6A3}3.91$\times$ & \cellcolor[HTML]{F6A574}7.00$\times$\\
SigLIP2-B~\cite{tschannen2025siglip2} & 16 & \cellcolor[HTML]{FFF2E9}1.08$\times$ & \cellcolor[HTML]{FFF1E7}1.13$\times$ & \cellcolor[HTML]{FFF0E5}1.18$\times$ & \cellcolor[HTML]{FEECDF}1.33$\times$ & \cellcolor[HTML]{FEE8D9}1.55$\times$ & \cellcolor[HTML]{FEE6D5}1.69$\times$ & \cellcolor[HTML]{FDE5D3}1.78$\times$ & \cellcolor[HTML]{FCD6BB}2.71$\times$ & \cellcolor[HTML]{FBC9A8}3.65$\times$ & \cellcolor[HTML]{FDE2CF}1.95$\times$ & \cellcolor[HTML]{FBC8A7}3.71$\times$ & \cellcolor[HTML]{F7AB7D}6.34$\times$\\
SigLIP2-L~\cite{tschannen2025siglip2} & 16 & \cellcolor[HTML]{FFF2E9}1.06$\times$ & \cellcolor[HTML]{FFF1E7}1.12$\times$ & \cellcolor[HTML]{FFF0E5}1.16$\times$ & \cellcolor[HTML]{FEEDE0}1.32$\times$ & \cellcolor[HTML]{FEE8D8}1.56$\times$ & \cellcolor[HTML]{FEE6D5}1.67$\times$ & \cellcolor[HTML]{FDE5D2}1.79$\times$ & \cellcolor[HTML]{FCD5BB}2.72$\times$ & \cellcolor[HTML]{FBC8A7}3.69$\times$ & \cellcolor[HTML]{FDE2CE}1.98$\times$ & \cellcolor[HTML]{FBC8A6}3.76$\times$ & \cellcolor[HTML]{F7AB7D}6.36$\times$\\
\bottomrule
\end{tabular}\end{widewrap}
\end{table*}

In the following, we explain where the speed advantage of \meros comes from.

\begin{plist}
\item \textbf{One layer of \meros} runs 4.0 to 24.8$\times$ faster than FA2 from 1K to 8K and 3.2 to 8.4$\times$ faster than FA4, the fastest exact kernel we test. Among the eleven training-free attention acceleration methods we compare, \meros $r{=}8$ is the fastest at every image size from 1K to 8K and $r{=}2$ has the lowest operator error of all at 8K.\PFILL

\item \textbf{Hardware efficiency.} The kernel of \meros sustains 51 to 54\% of the dense BF16 peak of a B300, against 49 to 50\% for FA4 (\cref{fig:supp_kernel}). This kernel takes 84 to 99\% of the GPU time of a call, and packing the samples, the global queries and the scatter of the outputs take the rest. Its tiles share one packed buffer of sampled keys, so 90 to 98\% of their reads hit the L2 cache. As in FA4, most of its stalls come from the softmax and from the barriers of its pipeline. Because the sample is fixed before inference, every launch has static shapes and never waits for the host, so the whole forward pass fits in a single CUDA graph.\PFILL

\item \textbf{Every backbone we time.} \Cref{tab:supp_backbones} times 14 frozen ViTs from seven families end to end. On every backbone and at every budget, the speedup over FA2 grows with the image side. With $r{=}8$, it ranges from 1.40 to 2.98$\times$ at 1K, 2.74 to 5.62$\times$ at 2K, 6.41 to 12.4$\times$ at 4K and 13.9 to 20.2$\times$ at 8K. At 4K, ten of the 14 backbones run more than ten times faster than with FA2. All budgets are also faster than FA4 on all backbones and sizes, by up to 7.09$\times$ for $r{=}8$ on \mbox{DINOv2-L} at 8K. Over FA4, $r{=}8$ gains 1.48 to 3.05$\times$ at 2K and 2.65 to 4.60$\times$ at 4K. At 8K, even $r{=}2$ runs every backbone 5.20 to 5.79$\times$ faster than FA2 and 1.86 to 2.08$\times$ faster than FA4.\PFILL

\item \textbf{GPU memory usage.} At 8K, the backbone of \mbox{DINOv3-B} peaks at 6.81 to 7.19\,GB with \meros, slightly below the 7.42\,GB of FA2. On UAVid tiles, token merging instead reserves over 16 times the peak memory of exact attention (FA2). Even with the 288\,GB of a B300, the original ViTMatte-B runs out of memory on 38 of the 50 Transparent-460 images, while exact attention in query chunks and \meros never exceed 29\,GB of peak memory on any of them.\PFILL
\end{plist}

Two things make \meros fast. Fewer keys are read, and the retained keys are read in a dense pass at least as efficiently as FA4 reads all of them. At 4096 and 8192\,px, the speedup over FA4 is therefore close to the reduction in keys. We provide a Triton kernel for Hopper and Blackwell GPUs, a CuTe kernel that adds speed on Blackwell and a generic kernel for any head dimension and relative position biases. \Cref{lst:supp_reference} gives a PyTorch reference of one attention layer.\PFILL

\subsection{Accuracy}
\label{sec:practice_accuracy}

Because the far-field estimate is unbiased, its error is noise that averages out across cells. Our measurements on \mbox{DINOv3-B}, MAE-L and DINOv2-L agree with the analysis. For every query and head we sampled, the error stays below the proven bound, and the mean of 256 draws lies within Monte Carlo noise of the exact output. The exact variance of the reflected design predicts the measured mean squared error to within a median of 2.5\%, and for 86 to over 99\% of the queries the reflected pairs give a lower error than a simple random sample. From 2048\,px up, every doubling of the image side multiplies the error by 0.53 on average, close to the predicted $1/2$.\PFILL

\begin{plist}

\item \textbf{The budget $r$} is the one setting to choose, and its effect is predictable. The operator error grows as $\sqrt{r-1}$. Over the 17 segmentation settings, the median mIoU change is $-0.07$, $-0.17$ and $-0.33$ at $r{=}2$, 4 and 8, for median backbone speedups of 3.04, 3.85 and 4.33$\times$. $r{=}2$ suits tasks that must track exact attention closely, and $r{=}8$ gives the most speed. Larger inputs in turn allow a larger budget at the same error.

\begin{table*}[t]
\centering
\caption{\textbf{\meros at $r{=}8$ on every task}, with the values printed in \cref{tab:seg,tab:dis,tab:matting,tab:vlm,tab:depth}. Speed is that of the backbone for segmentation and matting, of the encoder for depth, and of the first token for visual question answering (VQA).}
\label{tab:practice_r8}
\tabsetup{7.6}
\renewcommand{\arraystretch}{1.1}
\setlength{\tabcolsep}{2.5pt}
\setlength{\tabw}{\dimexpr\textwidth-2\tabcolsep\relax}%
\begin{tabular}{p{0.175\tabw}p{0.20\tabw}p{0.195\tabw}>{\centering\arraybackslash}p{0.115\tabw}>{\centering\arraybackslash}p{0.09\tabw}>{\centering\arraybackslash}p{0.105\tabw}>{\centering\arraybackslash}p{\dimexpr0.12\tabw-12\tabcolsep\relax}}
\toprule
\rowcolor{tabhead}[\tabcolsep][\rov] \bfseries Task & \bfseries Model & \bfseries Data & \bfseries Metric & \bfseries Exact (FA2) & \bfseries\boldmath \meros $r{=}8$ & \bfseries Speed\rlap{$\uparrow$}\\
\midrule
Semantic segmentation &  Seven ViTs~\cite{simeoni2025dinov3,touvron2022deit3,he2022mae,tschannen2025siglip2} & 17 settings~\cite{lyu2020uavid,cordts2016cityscapes,demir2018deepglobe,ji2023urur} & median $\Delta$mIoU & 0.00 & $-0.33$ & 4.33$\times$\\
Dichotomous segmentation & \mbox{DINOv3-B}~\cite{simeoni2025dinov3} & DIS-TE~\cite{qin2022dis}, 2048\,px & $F^{\max}_\beta\uparrow$ & .816 & .809 & 5.35$\times$\\
Image matting & ViTMatte-B~\cite{yao2024vitmatte} & PPM-100~\cite{ke2022modnet} & SAD$\downarrow$ & 35.3 & 35.5 & 3.78$\times$\\
 & ViTMatte-B~\cite{yao2024vitmatte} & AIM-500~\cite{li2021aim} & SAD$\downarrow$ & 14.4 & 14.6 & 2.63$\times$\\
Depth estimation & MoGe-3 ViT-G~\cite{kong2026moge3} & ETH3D~\cite{schops2017eth3d} & AbsRel (\%)$\downarrow$ & 3.26 & 3.29 & 6.75$\times$\\
 & Depth Anything V2-L~\cite{yang2024dav2} & Middlebury~\cite{scharstein2014middlebury} & AbsRel (\%)$\downarrow$ & 4.07 & 4.05 & 3.72$\times$\\
Question answering & Qwen3-VL-8B~\cite{bai2025qwen3vl} & HR-Bench 8K~\cite{wang2025hrbench} & Accuracy (\%)$\uparrow$ & 67.75 & 67.85 & 3.45$\times$\\
\bottomrule
\end{tabular}
\end{table*}

\item \textbf{A small cost on every task.} \Cref{tab:practice_r8} puts $r{=}8$ beside exact attention per task. \meros loses a median of 0.33~mIoU in segmentation, at most 0.007 in the $F^{\max}_\beta$ of DIS, 0.2 SAD in matting and at most 1.0 point on HR-Bench 8K. With MoGe-3 on all 454 ETH3D images, every budget stays within 0.03 points of the AbsRel of exact attention.

\item \textbf{At times more accurate than exact attention.} On HR-Bench 8K, $r{=}2$ matches or exceeds exact attention on all four VLMs, by up to 1.2 points, and answers up to 7.1$\times$ sooner. In matting, it lowers seven of eight errors at about twice the speed. In semantic segmentation, it matches or exceeds exact attention in four of the 17 settings, by up to 0.07~mIoU. On DIS, both budgets need less human correction effort than exact attention in all three settings. With Depth Anything V2, $r{=}8$ has the lower AbsRel on the Middlebury images at 3.72$\times$ the encoder speed.\PFILL

\item \textbf{Other far-field estimates.} At the key budget of $r{=}8$, spending it on the nearest keys or on a larger window instead raises the feature error 3.1 to 15.8$\times$ and costs 11.3~mIoU on average over both backbones. A weighted fixed grid comes closest in feature error, yet its worst mIoU loss is almost twice that of \meros.

\item \textbf{Seeds.} The \meros sampling seed fixes the sample of every layer and head before inference, so a model reads the same keys for every input. In dense prediction, the final representations are largely insensitive to the sampling seed, with the final feature error at $r{=}8$ exhibiting a standard deviation bounded by $0.0024$ across different seeds, as shown in \cref{tab:supp_equal_budget}. Over three sampling seeds, the mIoU of the 17 segmentation settings has a median standard deviation of 0.04 and at most 0.26 (\cref{tab:supp_seg_seeds}), and the $F^{\max}_\beta$ of DIS one of at most 0.001. \end{plist}

In addition, \meros removes no token. Since every patch keeps its own feature in every layer, dense heads read the model as they were trained to, with nothing to unmerge or fill in. Token merging instead returns a single feature to all the patches that a merged token absorbed. Thin structures gain the most from this. In the DIS, matting and depth examples of \cref{sec:exp}, spokes, hair and pipes survive at $r{=}8$, whereas token merging flips 3.5 to 17.6\% of the pixels of two DIS images. For \meros, these shares are 0.6 and 2.2\%, at under a twelfth of their backbone time per image.\PFILL

By design, \meros is therefore unbiased and grows more accurate as images grow, and without any training it runs faster than the fastest exact kernels in measured time.
Of the training-free methods we compare, it is the only one that keeps every token and its position biases, is unbiased and is faster than FA2 at 2K (\cref{fig:teaser}(a)). 

\section{Conclusion}
\label{sec:conclusion}

This paper studies training-free acceleration of self-attention at high resolution. We find that attention concentrates near each query, while most of its mass spreads thinly over distant tokens. Sampling this far field with known inclusion probabilities gives an unbiased estimate of the far-field sums, whose error bound halves each time the image side doubles. \meros computes the local tile and global tokens exactly and samples the far field at fixed positions, so it runs as one dense attention pass that keeps masks and position biases. On a B300 GPU, its attention runs 4.0--24.8$\times$ faster than FA2 from 1K to 8K. Over 17 segmentation settings, $r{=}2$ speeds up the whole backbone by a median of 3.0$\times$ at a median loss of 0.07~mIoU. Without training, \meros at $r{=}8$ also makes image matting, depth estimation and visual question answering 2.6 to 8.8$\times$ faster with little loss in accuracy.

\setlength{\premulticols}{84pt}
\begingroup\renewcommand{\bibsection}{}\let\newpage\pagebreak
\begin{multicols}{2}[\section*{References}]
\bibliographystyle{ieeenat_fullname}
\bibliography{refs}
\end{multicols}\endgroup

\clearpage
\setcounter{topnumber}{2}\setcounter{bottomnumber}{0}\setcounter{totalnumber}{2}\renewcommand{\topfraction}{0.92}\renewcommand{\floatpagefraction}{0.85}
\merosappendix

This appendix proves the results of \cref{sec:method} and gives the kernels, protocols and full results behind the experiments of \cref{sec:exp}. We also locate the attention mass on real samples and check the theory against measurements.

\section{Notation and Setting}
\label{sec:supp_notation}

Throughout the paper, we consider one attention head and one patch query $i$, whose index we drop when no confusion arises.
Let $\bq$ be the query vector of patch $i$, and $\bk_j$ and $\bv_j$ the key and value vectors of token $j$ in the same head, all taken after the positional encoding.
With head dimension $d$, the logits $s_j=\bq^\top\bk_j/\sqrt{d}$ give the total mass $Z=\sum_je^{s_j}$, the attention weights $p_j=e^{s_j}/Z$ and the exact output $\bo=\sum_jp_j\bv_j$.
As in \mainref{sec:anatomy}, the keys of the query fall into three groups.
The near field $\Loc$ is the $16\times16$-patch tile that contains the query, and the far field $\Far$ holds all other patches.
The rest are a few global tokens $\Glob$, such as the class token and registers~\cite{darcet2024registers}.
The $4\times4$ blocks of patches outside the tile form the set $\mathcal{C}$ of far cells, so every far key lies in exactly one cell.
Sums over $j$ without a range run over all keys, and sums over $c$ run over $\mathcal{C}$.
A key set $X$ has the mass $Z_X$ and the value sum $\bA_X$, and a cell $c$ holds the share $P_c=Z_c/Z$ of the total mass.
The far field and the exactly computed keys hold the shares $\piF$ and $\pex$, and $\neff$ is the effective number of far cells:
\begin{equation}
\begin{gathered}
  Z_X=\textstyle\sum_{j\in X}e^{s_j},\qquad \bA_X=\textstyle\sum_{j\in X}e^{s_j}\bv_j,\\
  \piF=\textstyle\sum_cP_c,\quad \pex=1-\piF,\quad \neff=\dfrac{\piF^2}{\sum_cP_c^2}.
\end{gathered}
\label{eq:supp_shares}
\end{equation}
Since $\sum_cP_c^2\le\piF^2\le|\mathcal{C}|\sum_cP_c^2$, the second step by Cauchy--Schwarz, $\neff$ lies in $[1,|\mathcal{C}|]$ as stated in \mainref{eq:neff}.
It equals $|\mathcal{C}|$ when all far cells hold the same mass and 1 when a single cell holds all of the far mass.

The far field is estimated from a random sample $\Samp$ that keeps one far key in $r$, for a budget $r\ge1$, and \meros uses $r\in\{1,2,4,8\}$ (\cref{sec:supp_details}).
The sample is a set of patch positions, and query $i$ reads the far keys and values only at these positions.
Key $j$ has the weight $w_j$, which is $1$ on $\Loc\cup\Glob$ and $r\,\mathbb{I}[j\in\Samp]$ on $\Far$.
The output of \mainref{eq:meros} is then $\hat\bo=\sum_jw_je^{s_j}\bv_j\big/\sum_jw_je^{s_j}$, with the far parts $\hat Z_{\Far}=\sum_{j\in\Far}w_je^{s_j}$ and $\hat\bA_{\Far}=\sum_{j\in\Far}w_je^{s_j}\bv_j$.
We describe its error by the normalizer error $\delta=\sum_jp_j(w_j-1)$ and the numerator error $\bU=\sum_jp_j(w_j-1)(\bv_j-\bo)$.
The estimated softmax denominator is $Z(1+\delta)$, and the estimated numerator is $Z(1+\delta)\,\bo+Z\bU$.
A common factor $1+\delta$ alone would leave the output exact, so $\bU$ is the part of the error that moves it.
Let $D\ge\max_j\|\bv_j-\bo\|$, as in \mainref{thm:error}.
Every average of the values with non-negative weights, $\hat\bo$ included, then lies within $D$ of $\bo$.
Expectations are over draws of $\Samp$ for a fixed input.
We call the rule that draws $\Samp$ the sampling design.
\meros draws reflected pairs of patches in every far cell, once per layer and head before inference (\mainref{sec:meros}).
The general results of \mainref{sec:analysis}, however, hold for any design with the following two properties.\PFILL

\noindent\textbf{(P1)}~Every far key belongs to $\Samp$ with probability $1/r$. With its weight $r$, each far key then counts once in expectation.

\noindent\textbf{(P2)}~The samples drawn in different cells are independent, so the squared errors of different cells add up in expectation.

\noindent The design of \meros has both properties (\cref{sec:supp_inclusion}).
\Cref{lem:unbiased} needs only (P1), whereas \cref{thm:error} needs both.

\section{Proofs}
\label{sec:supp_proofs}

We first verify that the reflected design of \meros (\mainref{sec:meros}) has both properties, then prove the results of \cref{sec:method}.
Only \cref{prop:pairs} and the exact variance of \cref{sec:supp_variance} use the reflection, whereas all other proofs need only (P1) and (P2).

\subsection{Properties of the Reflected Design}
\label{sec:supp_inclusion}

A far cell has 16 positions $u=(y,x)$ with $y,x\in\{0,1,2,3\}$, which form eight pairs $\{u,\bar u\}$ with $\bar u=(3-y,3-x)$.
The reflection maps the cell onto itself and fixes no position, since $3-y=y$ has no integer solution.
The eight pairs are therefore disjoint and cover the cell.
A cell orders its pairs uniformly at random and keeps the first $k=8/r$ of them, so query $i$ reads the keys and values of $16/r$ of its patches.
Every pair is equally likely to take any place in the order, so it is kept with probability $k/8=1/r$.
A key is sampled exactly when its pair is, which gives (P1).
The orders of different cells are drawn independently, which gives (P2), and so are those of different layers and heads.
Since every budget keeps a prefix of the same order, the samples $\Samp^{(r)}$ of the budgets $r=8,4,2$ are nested, $\Samp^{(8)}\subset\Samp^{(4)}\subset\Samp^{(2)}$, for every layer and head.\PUP

\subsection{Proof of \texorpdfstring{\cref{prop:bias}}{Proposition 1}}

\emph{(i) Selection.}
Selection keeps a set $\mathcal{K}\subset\Far$ of far keys (\mainref{sec:select}). In this proof only, $\hat\bo$ denotes the output of selection.
Let $\mathcal{D}=\Far\setminus\mathcal{K}$ be the dropped keys and $\pi_{\mathcal{D}}=Z_{\mathcal{D}}/Z$ their share.
Selection removes $\mathcal{D}$ from both sums, so the exact output is the mixture $\bo=(1-\pi_{\mathcal{D}})\,\hat\bo+\pi_{\mathcal{D}}\,\bo_{\mathcal{D}}$ with $\bo_{\mathcal{D}}=\bA_{\mathcal{D}}/Z_{\mathcal{D}}$.
Rearranging gives $\hat\bo-\bo=\pi_{\mathcal{D}}(\hat\bo-\bo_{\mathcal{D}})$.
The kept mass is $1-\pi_{\mathcal{D}}=\pex+\sum_{j\in\mathcal{K}}p_j\le\pex+|\mathcal{K}|\max_{j\in\Far}p_j$, so the lost mass satisfies $\pi_{\mathcal{D}}\ge\piF-|\mathcal{K}|\max_{j\in\Far}p_j$.
For nearly uniform far weights, $\max_{j\in\Far}p_j\approx\piF/|\Far|$, and keeping $|\mathcal{K}|=|\Far|/r$ keys loses about $\piF(1-1/r)$ of the mass.\PUP

\emph{(ii) Summarization.}
Under proportional attention~\cite{bolya2023tome}, a cell $c$ is replaced by its mean key $\bar\bk_c$ and mean value $\bar\bv_c$, counted $|c|$ times.
The query thus attends to one merged token per far cell instead of its 16 keys.
The logit of the mean key is the mean logit, $\bq^\top\bar\bk_c/\sqrt{d}=\bar s_c$, so the cell enters the normalizer with the mass $|c|\,e^{\bar s_c}$.
With $\Delta_c=\log\frac{1}{|c|}\sum_{j\in c}e^{s_j-\bar s_c}$ as in \cref{prop:bias}, its exact mass is
\begin{equation}
  Z_c=\sum_{j\in c}e^{s_j}=|c|\,e^{\bar s_c}\cdot\frac{1}{|c|}\sum_{j\in c}e^{s_j-\bar s_c}=|c|\,e^{\bar s_c}\,e^{\Delta_c}.
\end{equation}
The summarized mass is therefore $Z_ce^{-\Delta_c}$, and Jensen's inequality gives
\begin{equation}
  e^{\Delta_c}=\frac{1}{|c|}\sum_{j\in c}e^{s_j-\bar s_c}\ge\exp\Bigl(\frac{1}{|c|}\sum_{j\in c}(s_j-\bar s_c)\Bigr)=1,
\end{equation}
with equality if and only if all logits in the cell are equal.
For a small spread of the logits in the cell, $\Delta_c$ is about half their variance.
Summed over the far field, the relative error of the far mass is
\begin{equation}
  \frac{\sum_cZ_c(e^{-\Delta_c}-1)}{Z_{\Far}}=\sum_c\frac{P_c}{\piF}\,(e^{-\Delta_c}-1)\le0 .
\end{equation}

\subsection{Proof of \texorpdfstring{\cref{lem:unbiased}}{Lemma 1}}

The lemma is the unbiasedness of the Horvitz--Thompson estimator~\cite{horvitz1952generalization}.
By (P1), $\E\bigl[r\,\mathbb{I}[j\in\Samp]\bigr]=1$ for every far key $j$.
For a far cell $c$, $\hat Z_c=\sum_{j\in c}r\,\mathbb{I}[j\in\Samp]\,e^{s_j}$, so linearity of expectation gives $\E[\hat Z_c]=Z_c$.
Summing over the cells, we obtain $\E[\hat Z_{\Far}]=Z_{\Far}$.
The same argument, applied to $e^{s_j}\bv_j$, gives $\E[\hat\bA_{\Far}]=\bA_{\Far}$.
All terms of $\hat Z_{\Far}$ are non-negative, so $\hat Z_{\Far}\ge0$ on every draw.
The estimated normalizer is therefore at least $Z_{\Loc}+Z_{\Glob}>0$, because the query computes its tile and global tokens exactly.
The proof uses only (P1), so the lemma holds for any logits and values.

\subsection{Proof of \texorpdfstring{\cref{thm:error}}{Theorem 1}}

We first write the output error in terms of $\delta$ and $\bU$ and bound their second moments with (P1) and (P2).
We then treat the draws on which the normalizer is poorly estimated, and finally bound the bias.\PFILL

\emph{Error identity.}
As in self-normalized importance sampling~\cite[Sec.~9.2]{owen2013mcbook}, $\hat\bo$ is a ratio of two unbiased estimates, the estimated numerator $\sum_jw_je^{s_j}\bv_j$ and the estimated denominator $\sum_jw_je^{s_j}$ of the softmax.
Since $\sum_jp_jw_j=1+\delta$, the denominator is $Z(1+\delta)$, and the numerator exceeds $Z(1+\delta)\,\bo$ by
\begin{equation}
  \textstyle\sum_jw_je^{s_j}(\bv_j-\bo)=Z\sum_jp_jw_j(\bv_j-\bo)=Z\,\bU,
\end{equation}
where the last step uses $\sum_jp_j(\bv_j-\bo)=\mathbf{0}$.
The denominator is at least $Z_{\Loc}+Z_{\Glob}>0$ (proof of \cref{lem:unbiased}), so dividing the numerator by it gives $\hat\bo-\bo=\bU/(1+\delta)$ on every draw.
Both $\delta$ and $\bU$ are linear in the weights.
Every key has $\E[w_j]=1$, by (P1) for far keys and by construction for the tile and global keys, so $\E[\delta]=0$ and $\E[\bU]=\mathbf{0}$.

\emph{Second moments.}
Only far keys have $w_j\neq1$, so $\bU=\sum_{c\in\mathcal{C}}\bU_c$ with $\bU_c=\sum_{j\in c}(w_j-1)\ba_j$ and $\ba_j=p_j(\bv_j-\bo)$.
We call $\ba_j$ the pull of key $j$ on the output, and $\bU_c$ is the error that the sample of cell $c$ adds to $\bU$.
By (P1) the $\bU_c$ have zero mean, and by (P2) they are independent, so $\E\|\bU\|^2=\sum_c\E\|\bU_c\|^2$.
To bound one cell, let $M_c=\sum_{j\in c}\|\ba_j\|$ be its total pull and $\hat M_c=\sum_{j\in c}w_j\|\ba_j\|$ the estimate of $M_c$ from its sampled keys.
Since $\bU_c$ is $\sum_{j\in c}w_j\ba_j$ minus its mean $\sum_{j\in c}\ba_j$, its second moment is a variance, and the triangle inequality gives $\|\sum_{j\in c}w_j\ba_j\|\le\hat M_c$, so
\begin{equation}
  \E\|\bU_c\|^2=\E\Bigl\|\sum_{j\in c}w_j\ba_j\Bigr\|^2-\Bigl\|\sum_{j\in c}\ba_j\Bigr\|^2\le\E[\hat M_c^2] .
\end{equation}
No weight exceeds $r$, so $0\le\hat M_c\le rM_c$ on every draw, while $\E[\hat M_c]=M_c$ by (P1).
Hence $\E[\hat M_c^2]\le rM_c\,\E[\hat M_c]=rM_c^2$.
The bound $\|\bv_j-\bo\|\le D$ gives $\|\ba_j\|\le p_jD$ and $M_c\le DP_c$, and summing over the cells, we obtain $\E\|\bU\|^2\le rD^2\sum_cP_c^2$.
The same steps with the scalar weights $p_j$ in place of $\ba_j$ give $\E[\delta^2]\le r\sum_cP_c^2$.

\emph{Output error.}
The identity alone does not bound the output error, since $1+\delta$ can be small on some draws.
Let $\mathcal{E}$ be the event $|\delta|\le\tfrac12$, on which $\|\hat\bo-\bo\|=\|\bU\|/(1+\delta)\le2\|\bU\|$.
On every draw, $\hat\bo$ is a convex combination of the values of the keys with $w_j>0$, so $\|\hat\bo-\bo\|\le D$.
Chebyshev's inequality gives $\Pr[\mathcal{E}^{\complement}]\le4\,\E[\delta^2]$.
Splitting the expectation over $\mathcal{E}$ and its complement then gives
\begin{align}
  \E\|\hat\bo-\bo\|^2&\le4\,\E\|\bU\|^2+4D^2\,\E[\delta^2]\nonumber\\
  &\le8\,rD^2\textstyle\sum_cP_c^2 .
  \label{eq:supp_mse}
\end{align}
Substituting $\sum_cP_c^2=\piF^2/\neff$ into \cref{eq:supp_mse} and taking square roots gives the bound of \mainref{eq:law}.

\emph{Bias.}
Let $\epsilon=8r\sum_cP_c^2=8r\piF^2/\neff$, so that $\E\|\bU\|^2\le\epsilon D^2/8$, $\E[\delta^2]\le\epsilon/8$ and $\Pr[\mathcal{E}^{\complement}]\le\epsilon/2$.
We split $\E[\hat\bo-\bo]$ over $\mathcal{E}$ and its complement.
On $\mathcal{E}^{\complement}$, $\|\hat\bo-\bo\|\le D$, which contributes at most $D\Pr[\mathcal{E}^{\complement}]\le\epsilon D/2$.
On $\mathcal{E}$, $\hat\bo-\bo=\bU-\bU\delta/(1+\delta)$ with $|\delta|/(1+\delta)\le2|\delta|$.
Because $\E[\bU]=\mathbf{0}$, we also have $\E\bigl[\bU\,\mathbb{I}[\mathcal{E}]\bigr]=-\E\bigl[\bU\,\mathbb{I}[\mathcal{E}^{\complement}]\bigr]$.
The Cauchy--Schwarz inequality bounds the two remaining terms,
\begin{align}
  \bigl\|\E\bigl[\bU\,\mathbb{I}[\mathcal{E}^{\complement}]\bigr]\bigr\|&\le\sqrt{\E\|\bU\|^2\,\Pr[\mathcal{E}^{\complement}]}\le\epsilon D/4,\nonumber\\
  2\,\E\bigl[\|\bU\|\,|\delta|\bigr]&\le2\sqrt{\E\|\bU\|^2\,\E[\delta^2]}\le\epsilon D/4 .
\end{align}
Adding the three contributions gives $\|\E[\hat\bo]-\bo\|\le\epsilon D=8\,\piF^2D\,r/\neff$, which is the square of the bound in \mainref{eq:law} divided by $D$.
The identity and $\E[\bU]=\mathbf{0}$ also give $\E[\hat\bo]-\bo=-\E[\bU\delta/(1+\delta)]$, and \cref{prop:anchor} bounds $1+\delta$ from below by $\pex$.
Combining $1+\delta\ge\pex$ with the Cauchy--Schwarz inequality, we obtain $\|\E[\hat\bo]-\bo\|\le\piF^2D\,r/(\pex\neff)$.
This bound is tighter whenever $\pex>1/8$, whereas the first does not depend on $\pex$.\PFILL

\subsection{Proof of \texorpdfstring{\cref{cor:resolution}}{Corollary 1}}

The even share of a far cell is $\piF/|\mathcal{C}|$, the mass it would hold if the query spread its far mass equally over all far cells.
Suppose that no far cell holds more than $\kappa$ times its even share, so that $P_c\le\kappa\,\piF/|\mathcal{C}|$ for every cell. Then
\begin{equation}
  \textstyle\sum_cP_c^2\le\bigl(\max_cP_c\bigr)\sum_cP_c\le\kappa\,\piF^2/|\mathcal{C}| .
\end{equation}
Hence $\neff\ge|\mathcal{C}|/\kappa$, and substituting this into \mainref{eq:law} gives the stated bound.
For an image of $N$ patches, $|\mathcal{C}|=(N-256)/16$. With $\piF\le1$, the stated bound is at most
\begin{equation}
  2\sqrt{2}\,D\,\sqrt{\frac{16\,\kappa\, r}{N-256}},
\end{equation}
the form used after \cref{cor:resolution}.
Doubling the side scales it by $\sqrt{(N-256)/(4N-256)}$, which tends to $\frac12$ from below.\PUP

\subsection{Proof of \texorpdfstring{\cref{prop:pairs}}{Proposition 2}}

As in \cref{prop:pairs}, $\bff$ assigns a term to each position $u$ of a far cell $c$, such as $p_u$ or $\ba_u$.
The sample estimates the cell total $\bT_c=\sum_{u\in c}\bff(u)$ by $\hat\bT_c=r\sum_{u\in\Samp\cap c}\bff(u)$, and for these two choices of $\bff$ the error $\hat\bT_c-\bT_c$ is the part of $\delta$ or $\bU$ that comes from cell $c$.
The even part $\bff^{+}(u)=\frac12\bigl(\bff(u)+\bff(\bar u)\bigr)$ is the mean term of the pair $\{u,\bar u\}$, and the odd part $\bff^{-}(u)=\frac12\bigl(\bff(u)-\bff(\bar u)\bigr)$ is half the difference within it.
Since $\bff^{+}(\bar u)=\bff^{+}(u)$ and $\bff^{-}(\bar u)=-\bff^{-}(u)$, the odd part sums to zero over every pair and over the cell.
The cell mean $\bar\bff=\frac{1}{16}\sum_{u\in c}\bff(u)$ is therefore also the mean of $\bff^{+}$.
The cell keeps $k=8/r$ of its pairs.
In Monte Carlo terms, these pairs are antithetic samples~\cite[Sec.~8.2]{owen2013mcbook} within a stratified sample with one stratum per far cell~\cite[Sec.~8.4]{owen2013mcbook}.\PFILL

\emph{(i)}
A sampled pair $\{u,\bar u\}$ contributes $\bff(u)+\bff(\bar u)=\bff^{+}(u)+\bff^{+}(\bar u)$, so $\hat\bT_c=r\sum_{u\in\Samp\cap c}\bff^{+}(u)$, while $\bT_c=\sum_{u\in c}\bff^{+}(u)$.
Both depend on $\bff$ only through $\bff^{+}$.
Now let $\bff$ be affine on the cell, $\bff(u)=\bff_0+\boldsymbol{\Gamma}(u-u_0)$ with a constant matrix $\boldsymbol{\Gamma}$ and cell center $u_0$, so each term changes linearly across the cell.
Then $\tfrac12(u+\bar u)=u_0$ gives $\bff^{+}(u)=\bff_0=\bar\bff$ for every $u$, so $\hat\bT_c=r\cdot\frac{16}{r}\,\bar\bff=\bT_c$ for every draw of the sample.\PFILL

\emph{(ii)}
Let $\mathbf{y}_1,\dots,\mathbf{y}_8$ be the eight pair sums, with mean $\bar{\mathbf{y}}=2\bar\bff$. A sampled pair adds its sum to the cell total before the weight $r$.
The estimator draws $k$ of the eight pair sums without replacement and multiplies their sum by $8/k=r$.
Let $S_y^2=\frac17\sum_{l=1}^{8}\|\mathbf{y}_l-\bar{\mathbf{y}}\|^2$ be the variance of the pair sums.
Under simple random sampling from a finite population~\cite[Ch.~2]{cochran1977sampling}, the variance of the estimated total is $64\,(1-k/8)\,S_y^2/k=8(r-1)\,S_y^2$.
The estimate is unbiased, so this variance is its mean squared error.
Each pair sum equals $2\bff^{+}(u)$ for both of its positions, so $\sum_l\|\mathbf{y}_l-\bar{\mathbf{y}}\|^2=2\sum_{u\in c}\|\bff^{+}(u)-\bar\bff\|^2$ and
\begin{equation}
  \mathrm{MSE}_{\mathrm{pair}}=\frac{16(r-1)}{7}\sum_{u\in c}\|\bff^{+}(u)-\bar\bff\|^2 .
  \label{eq:supp_msepair}
\end{equation}
For comparison, a simple random sample draws $16/r$ keys of the cell without regard to their reflections, as in the ablation with independent positions (\cref{tab:supp_components}b).
Applied to a simple random sample of $16/r$ of the 16 positions and a function $\mathbf{g}$ with cell mean $\bar{\mathbf{g}}$, the same formula gives
\begin{equation}
  V(\mathbf{g})=\frac{16(r-1)}{15}\sum_{u\in c}\|\mathbf{g}(u)-\bar{\mathbf{g}}\|^2 .
\end{equation}
With $V^{\pm}=V(\bff^{\pm})$, the two formulas share the same sum of squares, so $\mathrm{MSE}_{\mathrm{pair}}=\frac{15}{7}V^{+}$, where $\frac{15}{7}$ is the ratio of the finite-population divisors $16-1$ and $8-1$.
For simple random sampling of $\bff$ itself, the cross term $\sum_{u\in c}\langle\bff^{+}(u)-\bar\bff,\,\bff^{-}(u)\rangle$ vanishes.
Re-indexing this sum by $\bar u$ leaves its value unchanged, but it keeps the first factor and flips the sign of the second, so the sum equals its own negative.
Hence $\sum_u\|\bff(u)-\bar\bff\|^2=\sum_u\|\bff^{+}(u)-\bar\bff\|^2+\sum_u\|\bff^{-}(u)\|^2$.
Since $\bff^{-}$ has mean zero over the cell, the simple random sample has $\mathrm{MSE}_{\mathrm{simple}}=V^{+}+V^{-}$.
Pairs therefore have the smaller mean squared error exactly when $\frac{15}{7}V^{+}<V^{+}+V^{-}$, that is, when $V^{-}$ exceeds $\frac87V^{+}$.

\subsection{Exact Variance of the Reflected Design}
\label{sec:supp_variance}

The proof of \cref{thm:error} bounds $\E\|\bU\|^2$ for every design with (P1) and (P2), whereas \cref{eq:supp_msepair} gives it exactly for the reflected design.
With $\bff(u)=\ba_u$, the error $\hat\bT_c-\bT_c$ equals $\bU_c$.
Summing over the independent cells gives
\begin{equation}
  \E\|\bU\|^2=\frac{16(r-1)}{7}\sum_{c\in\mathcal{C}}\sum_{u\in c}\|\ba^{+}_u-\bar\ba_c\|^2,
\end{equation}
where $\ba^{+}_u=\frac12(\ba_u+\ba_{\bar u})$ is the mean of the pair of $u$, and $\bar\ba_c$ is the mean of $\ba$ over cell $c$.
Thus $\E\|\bU\|^2$ grows linearly with $r-1$ and vanishes at $r=1$, which matches the $\sqrt{r-1}$ growth of the operator error in \mainref{sec:exp_speed}.
It depends only on how the pair means vary inside each cell, so a cell whose eight pairs have equal means adds nothing, however large its mass.
The same formula with the scalars $p_u$ in place of $\ba_u$ gives $\E[\delta^2]$.\PFILL

\subsection{Proof of \texorpdfstring{\cref{prop:anchor}}{Proposition 3}}

The estimated normalizer is $Z_{\Loc}+Z_{\Glob}+\hat Z_{\Far}=Z(1+\delta)$.
By \cref{lem:unbiased}, $\hat Z_{\Far}\ge0$ on every draw.
Dividing by $Z$ gives $1+\delta\ge(Z_{\Loc}+Z_{\Glob})/Z=\pex$.
The identity in the proof of \cref{thm:error} then gives the bound $\|\hat\bo-\bo\|\le\|\bU\|/\pex$.

\subsection{The Softmax Form and Additive Terms}
\label{sec:supp_softmax}

As noted below \mainref{eq:meros}, $r\,e^{s_j}=e^{s_j+\log r}$ for a sampled far key, so the operator is exactly the softmax over $\Loc\cup\Glob\cup(\Samp\cap\Far)$ with $\log r$ added to every sampled far logit.
All other keys receive weight zero.
For $r=1$, every far key is sampled with weight one, so \mainref{eq:meros} reduces to the original attention, which our code then calls directly.
If a layer adds a term $b_j$ to each logit, such as a relative position bias or a mask with $b_j=-\infty$, exact attention uses the masses $e^{s_j+b_j}$.
\meros adds $b_j$ to the logit of every key it reads, whether exact or sampled, and so estimates sums of these masses.
\Cref{lem:unbiased,thm:error,prop:anchor} hold for any non-negative masses with $Z_{\Loc}+Z_{\Glob}>0$, so they apply with $s_j+b_j$ in place of $s_j$.\PFILL

\section{Design and Implementation Details}
\label{sec:supp_details}

\Cref{eq:meros} is written per query, but a layer evaluates all queries of a head with one shared far-field sample.

\subsection{One Sample for All Queries}

For a fixed query, the results of \cref{sec:analysis} depend only on the distribution of $\Samp$.
This distribution is the same whether $\Samp$ is drawn for that query alone or shared by all queries of the head.
Sharing thus leaves every statement unchanged for each query and only correlates the errors of different queries within a head.
In exchange, all queries of a tile read the same far keys from a sample fixed before the forward pass, which makes the dense kernel of \cref{sec:supp_kernel} possible.

\subsection{Constructing the Sample}

As in \cref{sec:supp_inclusion}, every head and cell orders its eight pairs uniformly at random, independently of all others, and seeded pseudo-random generators draw these orders.
Budget $r$ keeps the first $8/r$ pairs of each order, so the budgets differ only in how many pairs they keep.
The sampled patch indices depend only on the seed, layer, head, number of heads and grid size.
We compute them once per layer and resolution and cache them as $N/r$ integers per head, a negligible amount next to the memory of the keys and values.\PFILL

\subsection{Cell Size and Budgets}

Refining the cells can only tighten the bound of \cref{thm:error}.
If a cell $c$ splits into sub-cells $c_1,\dots,c_m$, for example a $4\times4$ cell into four $2\times2$ blocks, then $\sum_lP_{c_l}^2\le\bigl(\sum_lP_{c_l}\bigr)^2=P_c^2$, so $\sum_cP_c^2$ cannot increase.
At budget $r$, a cell $c$ keeps $|c|/r$ of its positions.
A whole pair needs at least two, so $|c|\ge2r$.
At $r=8$, this requires $|c|\ge16$, which makes the $4\times4$ cell the finest square that keeps a whole pair at every budget $r\in\{2,4,8\}$, with nested samples.
For an integer budget, the number $8/r$ of kept pairs per cell is an integer only for $r\in\{1,2,4,8\}$, where $r=1$ recovers exact attention.\PUP

\subsection{Grid Alignment and Position Terms}
\label{sec:supp_grid}

The tiles require grid sides that are multiples of 16 patches.
For other grids, we pad the image on the right and bottom to the next multiple of 16 patches before the backbone.
The padded patches take part in attention as ordinary tokens, and the dense prediction protocol crops the outputs back to the original size.
The depth models and Kimi-VL, which resize their inputs anyway, resize them to such a multiple instead.
\meros operates on queries and keys after the model's own positional processing, such as the rotary position embeddings (RoPE)~\cite{su2024roformersupp} of \mbox{DINOv3}~\cite{simeoni2025dinov3}.
A sampled key keeps the vector computed at its own position, so no position is re-encoded and every value stays paired with its own key.
The ViTDet-style layers~\cite{li2022vitdet} of ViTMatte and SAM add a decomposed relative position bias, $b_{ij}=\bq_i^\top\mathbf{R}^{\mathrm{h}}_{y_i-y_j}+\bq_i^\top\mathbf{R}^{\mathrm{w}}_{x_i-x_j}$.
Its two terms index the model's learned tables $\mathbf{R}^{\mathrm{h}}$ and $\mathbf{R}^{\mathrm{w}}$ by the offsets in grid row $y$ and grid column $x$ between patches $i$ and $j$.
Our generic Triton kernel gathers this bias for exactly the keys it reads and adds it to their logits (\cref{sec:supp_softmax}).\PUP

\section{The Dense Kernel}
\label{sec:supp_kernel}

This section derives the dense computation that \cref{sec:meros} summarizes, gives a PyTorch reference and describes the three kernels we use.
The few global queries of the class and register tokens attend to all keys exactly, so the three kernels serve the patch queries.
The first is a Triton~\cite{tillet2019triton} kernel for Hopper (H200) and Blackwell (B300) GPUs.
It loads its operands through the tensor memory accelerator (TMA) of these GPUs, which copies whole blocks of a tensor into on-chip memory.
On the B300, a second kernel is usually faster.
We derived it from the fused multi-head attention example of CUTLASS~\cite{nvidia2026cutlasssw} and wrote it in its CuTe Python DSL.
For each backbone marked CuTe/Triton in \cref{tab:supp_models}, a short timed run on the B300 picks the fastest of up to four paths at each image size and budget.
These are the Triton kernel, its variant for $r{=}8$, and the CuTe kernel with key blocks of 128 or 64.
The third, a generic Triton kernel, runs the encoders of the vision-language models (VLMs), ViTMatte and SAM~\cite{kirillov2023sam}.
It supports any head dimension and adds the relative position biases of ViTDet-style layers (\cref{sec:supp_grid}).\PFILL

\subsection{Why the Kernel Must Stay Dense}

FlashAttention~\cite{dao2022flashattention} is fast in part because it reuses every key it loads.
A key and its value hold $2d$ numbers, or $4d$ bytes in BF16.
Each query that attends to them spends $4d$ floating-point operations, $2d$ on the logit and $2d$ on the weighted value.
When the 256 queries of a tile share their keys, one load serves all of them, and their logits form one matrix product on the tensor cores.
The arithmetic intensity, in operations per byte loaded from memory, is then
\begin{equation}
  \frac{256\cdot 4d\ \text{FLOP}}{4d\ \text{bytes}} = 256\ \text{FLOP/byte}.
\end{equation}
If every query reads its own keys, each load serves one query, the products become matrix-vector products and the intensity falls to 1~FLOP/byte.
An H200, for example, has a dense BF16 peak of 989~TFLOP/s and a memory bandwidth of 4.8~TB/s.
To keep its tensor cores busy, it needs about 206~FLOP/byte.
Without cache reuse, per-query keys leave the kernel two orders of magnitude below this point, so memory traffic alone sets its speed.
In practice, approximate attention also spends time choosing, clustering or hashing keys and reads them irregularly.
As a result, it often runs slower than exact attention and has seen little adoption~\cite{dao2022flashattention}.
In \mainref{tab:tf}, ten of the eleven training-free methods, in their official code, trail FA2 at 1024\,px.
Foundation models such as \mbox{DINOv3}~\cite{simeoni2025dinov3} and SAM~\cite{kirillov2023sam} still use dense softmax attention.
\meros fixes its sample in advance and keeps every tile dense (\cref{sec:supp_details}), so it runs on dense tensor-core kernels like exact attention.
For our CuTe kernel on the B300, \cref{sec:supp_profile} measures this intensity and how often its reads hit in the L2 cache.\PUP

\subsection{One Masked Softmax per Tile}

Let $t$ be a tile with patches $\Loc_t$, and let $\Samp_t=\Samp\cap\Loc_t$ be the samples drawn from it.
Every query $i$ in $t$ has $\Loc_i=\Loc_t$, and its sampled far keys are $\Samp\cap\Far_i=\Samp\setminus\Samp_t$.
All 256 queries of $t$ can therefore read the same keys and values,
\begin{equation}
  \tilde{\mathbf{K}}_t = \bigl[\mathbf{K}_{\Samp};\,\mathbf{K}_{\Loc_t};\,\mathbf{K}_{\Glob}\bigr], \quad
  \tilde{\mathbf{V}}_t = \bigl[\mathbf{V}_{\Samp};\,\mathbf{V}_{\Loc_t};\,\mathbf{V}_{\Glob}\bigr],
\end{equation}
with a logit offset $\boldsymbol{\lambda}_t$ that is zero on the last two ranges and, on the samples,
\begin{equation}
  \lambda_{t,j} = \begin{cases}
    \log r, & j\in\Samp\setminus\Samp_t,\\
    -\infty, & j\in\Samp_t.
  \end{cases}
\end{equation}
If the layer has its own bias $b_{ij}$ or mask, this term is added on top.
The offset multiplies the weight $e^{s_{ij}}$ of every far sample by $e^{\log r}=r$ and sets the weight of the samples of $t$ to zero.
It leaves the weights of the local and global keys unchanged.
By \cref{sec:supp_softmax}, the softmax of the offset logits over this sequence is therefore \mainref{eq:meros} for every query in $t$.
The whole tile computes
\begin{equation}
  \hat{\mathbf{O}}_t = \mathrm{softmax}\bigl(\mathbf{Q}_t\tilde{\mathbf{K}}_t^{\top}/\sqrt{d} + \mathbf{1}\boldsymbol{\lambda}_t^{\top}\bigr)\,\tilde{\mathbf{V}}_t
\end{equation}
in one dense call, as in \cref{alg:meros} with $\mathbf{B}_t=\mathbf{1}\boldsymbol{\lambda}_t^{\top}$.
We mask the samples of $t$ rather than drop them, so the far range of the key sequence, the $N/r$ samples, is the same for every tile.
In the Triton and CuTe kernels, masking wastes $256/r$ key slots per query.
For \mbox{DINOv3}-B at $r=8$ and 2048\,px, these are 32 of 2{,}309 key slots, and the other 2{,}277 are the keys counted in \cref{sec:supp_cost}.
The generic kernel instead skips every key block that holds only samples of its own tile.
The kernels load keys in key blocks of $b_k$ consecutive keys, one block per tensor-core product.
Padding the five global tokens to one such block wastes 123 more keys in the CuTe kernel with blocks of 128, and 59 more with blocks of 64.
The Triton kernel pads the global tokens to 16 keys, wasting 11 key slots.\PFILL

\subsection{The Weight in the Online Softmax}

FlashAttention~\cite{dao2022flashattention} extends the online softmax~\cite{milakov2018online} to attention and reads the key sequence in blocks.
For each query it keeps a running maximum $m$, a running sum $\ell$ and an accumulator $\mathbf{O}$, initialized to $-\infty$, $0$ and $\mathbf{0}$.
The maximum $m$ keeps the exponentials in range, while $\ell$ grows into the softmax denominator and $\mathbf{O}$ into the unnormalized output.
A block $\mathcal{B}$ of keys updates them to
\begin{align}
  m' &= \max\bigl(m,\ \textstyle\max_{j\in \mathcal{B}} s_j\bigr), \nonumber\\
  \ell' &= e^{m-m'}\,\ell + \textstyle\sum_{j\in \mathcal{B}} e^{s_j-m'}, \\
  \mathbf{O}' &= e^{m-m'}\,\mathbf{O} + \textstyle\sum_{j\in \mathcal{B}} e^{s_j-m'}\,\bv_j, \nonumber
\end{align}
and the output is $\mathbf{O}/\ell$ after the last block. A block whose keys are all masked leaves the state unchanged.
By induction over the blocks, the state always satisfies
\begin{equation}
  e^{m}\ell = \textstyle\sum_{j} e^{s_j}, \qquad e^{m}\mathbf{O} = \textstyle\sum_{j} e^{s_j}\,\bv_j,
  \label{eq:supp_invariant}
\end{equation}
where both sums run over all of the keys read so far, so that $\mathbf{O}/\ell$ is the softmax average of their values.

Our CuTe kernel reads the far range first, with weight one and the samples of $t$ masked.
After the far range, \cref{eq:supp_invariant} gives $e^{m}\ell=Z_{\Samp\setminus\Samp_t}$ and $e^{m}\mathbf{O}=\bA_{\Samp\setminus\Samp_t}$.
Replacing $m$ by $m+\log r$ and keeping $\ell$ and $\mathbf{O}$ multiplies both sums by $r$, which gives $\hat Z_{\Far_i}$ and $\hat\bA_{\Far_i}$.
The local and global ranges then add $Z_{\Loc_i}+Z_{\Glob}$ and $\bA_{\Loc_i}+\bA_{\Glob}$ with weight one, so the final $\mathbf{O}/\ell$ is $\hat\bo_i$ of \mainref{eq:meros}.
After the shift, $m$ exceeds every logit read so far, so it is no longer a running maximum.
The recurrence, however, relies only on \cref{eq:supp_invariant}, and every exponent stays non-positive.
Because the three terms share one normalizer, a single scan of the sequence suffices and no separately normalized parts need merging.
The shift costs one addition per query, while adding $\log r$ to every far logit costs one per far key.
Our Triton kernels take the second route and add the offset through the same vector $\boldsymbol{\lambda}$ that masks the far logits.
They compute exponentials in base 2, so the offset $\log_2 r\in\{1,2,3\}$ is exact in floating-point arithmetic.\PFILL

\subsection{A Worked Example of the Cost}
\label{sec:supp_cost}

\begin{table}[!tp]
\begin{minipage}[t]{0.47\textwidth}%
\renewcommand{\mtarget}{\linewidth}
\setlength{\extrarowheight}{1.0pt}
\setlength{\mcolw}{\dimexpr\linewidth*1000/961\relax}
\centering
\caption{\textbf{Keys per patch query} for \mbox{DINOv3}-B ($|\Loc_i|{=}256$, $|\Glob|{=}5$), from \cref{eq:supp_keys}. Parentheses give the reduction from the $N+|\Glob|$ keys that exact attention reads.}
\begin{colwrap}
\label{tab:supp_keys}
{\footnotesize\setlength{\tabcolsep}{3.3pt}\renewcommand{\arraystretch}{1.12}%
\setlength{\aboverulesep}{0pt}\setlength{\belowrulesep}{0pt}%
\setlength{\tabw}{\dimexpr\linewidth-2\tabcolsep\relax}%
\begin{tabular}{p{0.2\tabw}*{3}{>{\centering\arraybackslash}p{\dimexpr(0.8\tabw-6\tabcolsep)/3\relax}}}
\toprule
\rowcolor{tabhead}[\tabcolsep][\rov] Pixels & 2048 & 4096 & 8192 \\
\rowcolor{tabhead}[\tabcolsep][\rov] Patches $N$ & 16{,}384 & 65{,}536 & 262{,}144 \\
\midrule
Exact & 16{,}389 & 65{,}541 & 262{,}149 \\
\rowcolor{tabmeros}[\tabcolsep][\rov] \meros $r{=}2$ & 8{,}325\,(1.97$\times$) & 32{,}901\,(1.99$\times$) & 131{,}205\,(2.00$\times$) \\
\rowcolor{tabmeros}[\tabcolsep][\rov] \meros $r{=}4$ & 4{,}293\,(3.82$\times$) & 16{,}581\,(3.95$\times$) & 65{,}733\,(3.99$\times$) \\
\rowcolor{tabmeros}[\tabcolsep][\rov] \meros $r{=}8$ & 2{,}277\,(7.20$\times$) & 8{,}421\,(7.78$\times$) & 32{,}997\,(7.94$\times$) \\
\bottomrule
\end{tabular}}
\end{colwrap}
\end{minipage}\hfill
\begin{minipage}[t]{0.5\textwidth}%
\renewcommand{\mtarget}{\linewidth}
\setlength{\mcolw}{\dimexpr\linewidth*1000/961\relax}
\centering
\caption{\textbf{Efficiency} at batch~1 on one B300. Operator rows use layer~6 of \mbox{DINOv3}-B~\cite{simeoni2025dinov3}, or at 4K the range over nine layers of \mbox{DINOv3}-B/L and MAE-L~\cite{he2022mae}. The backbone ranges cover the 14 ViTs of \cref{sec:supp_backbones}, with patch-14 models at 1120, 2240, 4256 and 8288\,px. The largest speedup in each row is printed in bold.}
\begin{colwrap}
\label{tab:eff}
\tabsetup{7.2}
\setlength{\tabcolsep}{1.0pt}
\setlength{\tabw}{\dimexpr\linewidth-2\tabcolsep\relax}%
\begin{tabular}{>{\raggedright\arraybackslash}p{0.27\tabw}*{4}{>{\raggedleft\arraybackslash}p{\dimexpr(0.73\tabw-8\tabcolsep)/4\relax}}}
\toprule
\rowcolor{tabhead}[\tabcolsep][\rov]
\meros $r{=}8$ speedup & 1K & 2K & 4K & 8K \\
\midrule
Operator over FA2 & 4.0$\times$ & 11.0$\times$ & 20.6--21.0$\times$ & \textbf{24.8}$\times$ \\
Operator over FA4 & 3.2$\times$ & 4.2$\times$ & 6.8--7.1$\times$ & \textbf{8.4}$\times$ \\
Backbone over FA2 & 1.4--3.0$\times$ & 2.7--5.6$\times$ & 6.4--12.4$\times$ & \textbf{13.9--20.2}$\times$ \\
Backbone over FA4 & 1.1--2.7$\times$ & 1.5--3.1$\times$ & 2.7--4.6$\times$ & \textbf{5.0--7.1}$\times$ \\
\bottomrule
\end{tabular}
\end{colwrap}
\end{minipage}
\end{table}

Take \mbox{DINOv3}-B~\cite{simeoni2025dinov3}, with 16-px patches, at 2048\,px and $r=8$.
The image gives $128\times128 = 16{,}384$ patches, which form 64 tiles of 256 patches and 1{,}024 cells of 16, of which 1{,}008 lie outside a query's tile.
With the class token and four registers, $|\Glob| = 5$, so exact attention reads 16{,}389 keys per query.
Each cell keeps $8/r = 1$ pair, so each head samples $N/r = 2{,}048$ patches, 32 from each tile.
A patch query of tile $t$ reads these samples with the 32 from $t$ masked, then the 256 patches of $t$ and the five global tokens.
In total, it attends to $2{,}016+256+5 = 2{,}277$ keys, $7.2\times$ fewer than exact attention.
In general, a patch query attends to
\begin{equation}
  n_{\mathrm{key}}(N,r) = |\Loc_i| + |\Glob| + \frac{N-|\Loc_i|}{r}
  \label{eq:supp_keys}
\end{equation}
keys.
The ratio $(N+|\Glob|)/n_{\mathrm{key}}(N,r)$ approaches $r$ as $N$ grows, since the 261 exact keys become negligible next to the far samples (\cref{tab:supp_keys}).
By \cref{eq:supp_keys}, attention costs $O(N^2/r)$, while packing and scattering are linear in $N$.
The five global queries add only $5\times16{,}389$ logits, 0.2\% of the $16{,}384\times2{,}277$ logits of the patch queries.
Per head, the CuTe kernel copies $N(1+1/r)$ key and value rows, $N$ query rows and $N$ output rows.
The Triton kernel copies only the $N/r$ sampled key and value rows into a buffer.\PFILL

\subsection{A PyTorch Reference Implementation}
\label{sec:supp_reference}

\begin{figure}[!t]
\captionsetup{type=lstlisting}
\caption{\textbf{PyTorch reference of \meros} for one attention layer, runnable as printed. \texttt{sample} returns the sampled patches of one head, and \texttt{meros\_attention} evaluates \cref{eq:meros} for every patch query and runs the global queries exactly.}
\label{lst:supp_reference}
\vspace{4pt}
\begin{minipage}[t]{0.60\textwidth}\vspace{0pt}
\begin{lstlisting}[style=merospy,aboveskip=0pt,belowskip=0pt,basicstyle=\ttfamily\fontsize{7.6}{10.25}\selectfont]
import hashlib, math, random
import torch
import torch.nn.functional as F
sdpa = F.scaled_dot_product_attention

def sample(rows, cols, r, layer, head, seed=17):
    """First 8/r reflected pairs of every 4x4 cell."""
    tag, idx = f"meros-cell-v1|{seed}|{layer}|{head}", []
    for c in range(rows * cols // 16):
        cy, cx = divmod(c, cols // 4)
        key = f"{tag}|{cy}|{cx}|pair".encode()
        h = hashlib.sha256(key).digest()
        rng = random.Random(int.from_bytes(h, "little"))
        pairs = list(range(8))    # pair j: j and 15-j
        rng.shuffle(pairs)        # uniform order per cell
        for j in pairs[:8 // r]:  # prefixes: budgets nest
            for u in (j, 15 - j): # (y,x) and (3-y,3-x)
                idx.append((4*cy+u//4)*cols + 4*cx+u%4)
    return torch.tensor(idx)

def meros_attention(q, k, v, rows, cols, r, idx, g):
    """q, k, v: [B, H, g+N, d] after position encoding,
    g global tokens first, then N = rows*cols patches in
    raster order. idx: [H, N/r] samples of each head."""
    B, H, _, d = q.shape; idx = idx.to(q.device)
    nt = cols // 16                       # tiles per row
    tile = lambda p: p // cols // 16 * nt + p % cols // 16
    gi = (idx + g)[None, :, :, None].expand(B, H, -1, d)
    ks, vs = k.gather(2, gi), v.gather(2, gi)  # once
    far = torch.full_like(idx, math.log(r), dtype=q.dtype)
    rest = far.new_zeros(H, 256 + g)      # tile, globals
    p = torch.arange(rows * cols, device=q.device)
    out = torch.empty_like(q)
    for t in range(rows * cols // 256):
        own = g + p[tile(p) == t]    # 256 queries of t
        far_t = far.masked_fill(tile(idx) == t, -math.inf)
        bias = torch.cat((far_t, rest), 1)[None, :, None]
        kt = torch.cat((ks, k[:, :, own], k[:, :, :g]), 2)
        vt = torch.cat((vs, v[:, :, own], v[:, :, :g]), 2)
        out[:, :, own] = sdpa(q[:, :, own], kt, vt, bias)
    out[:, :, :g] = sdpa(q[:, :, :g], k, v)  # exact
    return out
\end{lstlisting}
\end{minipage}\hfill
\begin{minipage}[t]{0.375\textwidth}\vspace{0pt}\setlength{\parskip}{5pt plus 1pt}\setlength{\emergencystretch}{2em}
\noindent\textbf{Sampler.}
\texttt{sample} takes a grid of $n_y\times n_x$ patches (\texttt{rows}, \texttt{cols}), the budget $r$, the layer, the head and the seed, and returns the sampled patch indices.
It numbers the position $(y,x)$ of a $4\times4$ cell as $4y+x$, so pair $j\in\{0,\dots,7\}$ joins position $j$ with its reflection $15-j=4(3-y)+(3-x)$.
In every cell, a generator seeded with the SHA-256 hash of the seed, layer, head and cell coordinates shuffles the eight pairs.
The order is therefore uniformly random and independent across layers, heads and cells, which gives (P2).
Keeping its first $8/r$ pairs gives (P1) and nested samples, as in \cref{sec:supp_inclusion}.
\texttt{sample} follows our code's per-cell generator and returns the same sets as our code.
Stacking its output over the heads gives the $H\times N/r$ index matrix of one layer, which the operator gathers once for all tiles.

\noindent\textbf{Operator.}
\texttt{meros\_attention} takes $\mathbf{Q}$, $\mathbf{K}$ and $\mathbf{V}$ of shape $B\times H\times(|\Glob|+N)\times d$ after the model's own position encoding, with the global tokens first and the patches in raster order.
It gathers the sampled keys and values of all heads once.
For each tile $t$, one call of PyTorch's scaled dot-product attention then evaluates \mainref{eq:meros} for the 256 queries of $t$, with the key sequence and additive bias of \cref{sec:supp_kernel}.
The global queries attend to all $N+|\Glob|$ keys exactly.
Because the output keeps the shape and token order of the input, the function can replace the attention call of a model.
It materializes one key sequence per tile, which our three kernels avoid by streaming it one block at a time.
\end{minipage}
\end{figure}

Listing~\ref{lst:supp_reference} condenses the reference code we test our kernels against and computes the same outputs.
The notes beside the code describe its two functions, the sampler that runs before inference and the operator that replaces the attention call.

\subsection{Hardware Profile}
\label{sec:supp_profile}

We profile the kernels with the NVIDIA profilers Nsight Systems and Nsight Compute on layer~6 of \mbox{DINOv3}-B, the layer of \mainref{tab:tf}.
\Cref{fig:supp_kernel} shows \meros at $r{=}2$ and $r{=}8$ and FA4 on the B300 at 4096 and 8192\,px.

\noindent\textbf{Efficiency.}
Counting only the $4d$ FLOP per query--key pair that \mainref{eq:meros} needs, the CuTe kernel sustains 51 to 54\% of the dense BF16 peak.
FA4 reaches 49 to 50\% (\cref{fig:supp_kernel}(a)), so \meros processes each key it reads at least as fast as FA4, the fastest exact kernel we compare with.
At 4096 and 8192\,px, its speedup over FA4 is accordingly close to the reduction in keys of \cref{tab:supp_keys}.
Packing the samples, the split-K kernels of the global queries and the scatter of the outputs take 1 to 16\% of the GPU time (\cref{fig:supp_kernel}(b)).
The patch-query kernel takes the rest.

\noindent\textbf{Memory traffic.}
Between the shared L2 cache and the multiprocessors, the CuTe kernel performs 252 to 256~FLOP per byte.
This matches the intensity of 256~FLOP/byte derived in \cref{sec:supp_kernel}, because the two 128-query halves of a tile share every loaded key.
Since all tiles read the same packed buffer, 90 to 98\% of these reads hit in L2.
DRAM traffic is only 1.7 to 15.6 bytes for each key a tile reads, against the 256 bytes of a key and its value at $d=64$ without reuse.
With memory throughput at 21 to 23\% of its peak, bandwidth does not limit the CuTe kernel at these two sizes.

\noindent\textbf{Bottleneck.}
At 4096\,px, most stall samples of FA4 and \meros fall into two groups (\cref{fig:supp_kernel}(c)).
They wait either on the mbarriers of the warp-specialized pipeline or on the exponentials and floating-point work of the softmax.
For \meros, each group accounts for about a third, while issuing the matrix instructions, TMEM and TMA together account for under 3\%.
As in FA4, the softmax bounds the speed, while the tensor cores and the memory access pattern do not.

\section{Efficiency}
\label{sec:supp_results}

\Cref{tab:eff} summarizes the speedups over FA2 and FA4 of the operator and of 14 backbones, the seven ViTs of \mainref{tab:seg} and seven more.
\Cref{tab:speed} follows one \mbox{DINOv3-B} layer from 1K to 8K.
On this layer, the speedup of \meros $r{=}8$ over FA2 rises from 4.04$\times$ to 24.8$\times$, and its error to FP32 exact attention drops from .180 to .027.

\begin{table}[!tp]
\begin{minipage}[t]{0.435\textwidth}%
\renewcommand{\mtarget}{\linewidth}
\setlength{\extrarowheight}{0.9pt}
\setlength{\mcolw}{\dimexpr\linewidth*1000/961\relax}
\centering
\caption{\textbf{Whole-backbone speedup at larger batch sizes.} \meros $r{=}8$ against exact attention (FA2) at an equal batch size, on one B300 with random inputs of the 1024 and 2048\,px tiles of \cref{tab:seg}. The batch-1 column repeats the speedups from that table. The last two columns give the largest batch size that we have timed and the speedup of \meros over exact attention at that batch size.}
\begin{colwrap}
\label{tab:supp_batch}
\tabsetup{7.4}
\setlength{\tabcolsep}{2.4pt}
\setlength{\tabw}{\dimexpr\linewidth-2\tabcolsep\relax}%
\begin{tabular}{p{0.34\tabw}*{4}{>{\centering\arraybackslash}p{\dimexpr(0.66\tabw-8\tabcolsep)/4\relax}}}
\toprule
\rowcolor{tabhead}[\tabcolsep][\rov] & & \bfseries Batch 1 & \multicolumn{2}{c}{\bfseries Largest batch}\\
\rowcolor{tabhead}[\tabcolsep][\rov] \bfseries Backbone & \bfseries Tile & \bfseries\boldmath Speed\rlap{$\uparrow$}\phantom{$\downarrow$} & \bfseries Size & \bfseries\boldmath Speed\rlap{$\uparrow$}\phantom{$\downarrow$}\\
\midrule
\mbox{DINOv3-B}~\cite{simeoni2025dinov3} & 1024 & 2.93$\times$ & 8 & 3.19$\times$\\
\mbox{DINOv3-B}~\cite{simeoni2025dinov3} & 2048 & 5.35$\times$ & 16 & 5.40$\times$\\
MAE-B~\cite{he2022mae} & 1024 & 1.46$\times$ & 8 & 1.87$\times$\\
MAE-B~\cite{he2022mae} & 2048 & 3.87$\times$ & 6 & 4.44$\times$\\
MAE-L~\cite{he2022mae} & 1024 & 1.65$\times$ & 8 & 1.96$\times$\\
MAE-L~\cite{he2022mae} & 2048 & 4.25$\times$ & 6 & 4.43$\times$\\
DeiT\,III-L~\cite{touvron2022deit3} & 2048 & 4.32$\times$ & 6 & 4.35$\times$\\
\bottomrule
\end{tabular}
\end{colwrap}
\end{minipage}\hfill
\begin{minipage}[t]{0.535\textwidth}%
\renewcommand{\mtarget}{\linewidth}
\setlength{\mcolw}{\dimexpr\linewidth*1000/961\relax}
\centering
\caption{\textbf{\meros across resolutions} on layer~6 of \mbox{DINOv3-B}~\cite{simeoni2025dinov3} with its real queries, keys and values. (a)~Error on DIV8K~\cite{gu2019div8k} images, mean of three sampling seeds (SD $\le$ 0.002). (b)~Speedups timed on one B300 with the operator benchmark of \cref{sec:supp_timing}. Bold marks the fastest method at each input size. All columns except 6K also appear in \cref{tab:tf}. From 1K to 8K, the error of every budget falls about sevenfold, while the speedup of $r{=}8$ rises at every step, from 4.04$\times$ to 24.8$\times$.}
\begin{colwrap}
\label{tab:speed}
\tabsetup{7.4}
\renewcommand{\arraystretch}{1.0}
\setlength{\tabcolsep}{6.55pt}
\setlength{\tabw}{\dimexpr\linewidth-2\tabcolsep\relax}%
\begin{tabular}{p{0.25\tabw}*{5}{>{\centering\arraybackslash}p{\dimexpr(0.75\tabw-10\tabcolsep)/5\relax}}}
\toprule
\rowcolor{xhead}[\tabcolsep][\rov] & \multicolumn{5}{c}{\bfseries\boldmath (a) Relative error to FP32 exact attention $\downarrow$}\\
\hcmid\cmidrule(lr){2-6}
\rowcolor{xhead}[\tabcolsep][\rov] \bfseries\boldmath Method & \bfseries\boldmath 1K & \bfseries\boldmath 2K & \bfseries\boldmath 4K & \bfseries\boldmath 6K & \bfseries\boldmath 8K\\
\midrule
\rowcolor{xmeros}[\tabcolsep][\rov] \meros $r{=}2$ & \cellcolor{xerr!55!white}.073 & \cellcolor{xerr!41!white}.041 & \cellcolor{xerr!17!white}.020 & \cellcolor{xerr!7!white}.013 & \cellcolor{xerr!2!white}.010\\
\rowcolor{xmeros}[\tabcolsep][\rov] \meros $r{=}4$ & \cellcolor{xerr!65!white}.123 & \cellcolor{xerr!51!white}.071 & \cellcolor{xerr!31!white}.035 & \cellcolor{xerr!21!white}.023 & \cellcolor{xerr!12!white}.018\\
\rowcolor{xmeros}[\tabcolsep][\rov] \meros $r{=}8$ & \cellcolor{xerr!70!white}.180 & \cellcolor{xerr!60!white}.106 & \cellcolor{xerr!46!white}.052 & \cellcolor{xerr!36!white}.036 & \cellcolor{xerr!26!white}.027\\
\midrule
\rowcolor{xhead}[\tabcolsep][\rov] & \multicolumn{5}{c}{\bfseries\boldmath (b) Speedup over FA2~\cite{dao2024flashattention2} $\uparrow$}\\
\hcmid\cmidrule(lr){2-6}
\rowcolor{xhead}[\tabcolsep][\rov] \bfseries\boldmath Method & \bfseries\boldmath 1K & \bfseries\boldmath 2K & \bfseries\boldmath 4K & \bfseries\boldmath 6K & \bfseries\boldmath 8K\\
\midrule
FA4~\cite{zadouri2026flashattention} & \cellcolor{xspd!2!white}1.26$\times$ & \cellcolor{xspd!10!white}2.61$\times$ & \cellcolor{xspd!18!white}2.97$\times$ & \cellcolor{xspd!21!white}2.98$\times$ & \cellcolor{xspd!14!white}2.94$\times$\\
\rowcolor{xmeros}[\tabcolsep][\rov] \meros $r{=}2$ & \cellcolor{xspd!6!white}2.30$\times$ & \cellcolor{xspd!33!white}4.80$\times$ & \cellcolor{xspd!41!white}6.08$\times$ & \cellcolor{xspd!45!white}6.26$\times$ & \cellcolor{xspd!37!white}6.05$\times$\\
\rowcolor{xmeros}[\tabcolsep][\rov] \meros $r{=}4$ & \cellcolor{xspd!25!white}3.42$\times$ & \cellcolor{xspd!49!white}7.89$\times$ & \cellcolor{xspd!56!white}11.7$\times$ & \cellcolor{xspd!60!white}12.5$\times$ & \cellcolor{xspd!60!white}12.5$\times$\\
\rowcolor{xmeros}[\tabcolsep][\rov] \meros $r{=}8$ & \cellcolor{xspd!29!white}\textbf{4.04}$\times$ & \cellcolor{xspd!53!white}\textbf{11.0}$\times$ & \cellcolor{xspd!64!white}\textbf{21.0}$\times$ & \cellcolor{xspd!68!white}\textbf{23.5}$\times$ & \cellcolor{xspd!72!white}\textbf{24.8}$\times$\\
\bottomrule
\end{tabular}
\end{colwrap}
\end{minipage}
\end{table}

\subsection{Whole Backbones}
\label{sec:supp_backbones}

\noindent\textbf{Setup.}
\Cref{tab:supp_backbones} covers the seven ViTs of \mainref{tab:seg} together with \mbox{DINOv3-L}, \mbox{DINOv2-B/L}, EVA-02-B/L and CLIP-B/L.
Each backbone runs on a random image at batch~1 on one B300 and is timed with the backbone benchmark of \cref{sec:supp_timing}.
The patch-14 sizes give grid sides of 80 to 592 patches, all multiples of 16, as the $16\times16$ tiles of \meros require.\PUP

\noindent\textbf{Larger batches.}
\mainref{tab:seg} and \cref{tab:supp_backbones} time one input at a time.
\Cref{tab:supp_batch} also times batches of the 1024 and 2048\,px tiles of \mainref{tab:seg}, with the same benchmark and random inputs.
In every row, the speedup of $r{=}8$ at the largest batch is at least that at batch~1.
At batch~16, \mbox{DINOv3-B} processes 135 tiles of $2048^2$\,px per second with $r{=}8$, against 25.0 with FA2.

\section{Task Results}
\label{sec:supp_tasks}

\begin{table*}[!tp]
\centering
\caption{\textbf{Integration of \meros} into the models of \cref{sec:exp}. We list each model's patch size in pixels, head dimension $d$ and number $|\Glob|$ of class and register tokens. \meros replaces all layers or the listed global-attention layers. The grid is padded on the right and bottom and cropped after the head, padded only, or resized, so each side is a multiple of 16 patches. The last column names the timed kernel. CuTe/Triton is the faster of the two kernels at each input size and budget, and Generic the generic Triton kernel.}
\label{tab:supp_models}
\tabsetup{7.4}
\setlength{\tabcolsep}{2.4pt}
\setlength{\tabw}{\dimexpr\textwidth-2\tabcolsep\relax}%
\begin{widewrap}\begin{tabular}{>{\raggedright\arraybackslash}p{0.185\tabw}>{\raggedright\arraybackslash}p{0.125\tabw}*{4}{>{\centering\arraybackslash}p{0.045\tabw}}>{\raggedright\arraybackslash}p{0.085\tabw}>{\raggedright\arraybackslash}p{0.165\tabw}>{\raggedright\arraybackslash}p{0.075\tabw}>{\raggedright\arraybackslash}p{\dimexpr0.185\tabw-18\tabcolsep\relax}}
\toprule
\rowcolor{tabhead}[\tabcolsep][\rov] \bfseries Model & \bfseries Results & \bfseries Patch & \bfseries\boldmath $d$ & \bfseries Heads & \bfseries\boldmath $|\Glob|$ & \bfseries Layers & \bfseries Position terms & \bfseries Grid & \bfseries Kernel\\
\midrule
DINOv3-B~\cite{simeoni2025dinov3} & \cref{tab:tf,tab:seg,tab:dis,tab:analysis} & 16 & 64 & 12 & 5 & All 12 & RoPE & Pad, crop & CuTe/Triton\\
MAE-B~\cite{he2022mae} & \cref{tab:tome,tab:seg} & 16 & 64 & 12 & 1 & All 12 & Absolute & Pad, crop & CuTe/Triton\\
MAE-L~\cite{he2022mae} & \cref{tab:tome,tab:seg,tab:dis,tab:analysis} & 16 & 64 & 16 & 1 & All 24 & Absolute & Pad, crop & CuTe/Triton\\
DeiT\,III-B~\cite{touvron2022deit3} & \cref{tab:seg} & 16 & 64 & 12 & 1 & All 12 & Absolute & Pad, crop & CuTe/Triton\\
DeiT\,III-L~\cite{touvron2022deit3} & \cref{tab:tome,tab:seg} & 16 & 64 & 16 & 1 & All 24 & Absolute & Pad, crop & CuTe/Triton\\
SigLIP2-B~\cite{tschannen2025siglip2} & \cref{tab:seg} & 16 & 64 & 12 & 0 & All 12 & Absolute & Pad, crop & CuTe/Triton\\
SigLIP2-L~\cite{tschannen2025siglip2} & \cref{tab:seg} & 16 & 64 & 16 & 0 & All 24 & Absolute & Pad, crop & CuTe/Triton\\
ViTMatte-B~\cite{yao2024vitmatte} & \cref{tab:matting} & 16 & 64 & 12 & 0 & 2, 5, 8, 11 & Absolute, relative bias & Pad, crop & Generic\\
SAM ViT-H~\cite{kirillov2023sam} & \cref{fig:teaser} & 16 & 80 & 16 & 0 & 7, 15, 23, 31 & Absolute, relative bias & Resize, pad & Generic\\
MoGe-3 ViT-G~\cite{kong2026moge3} & \cref{tab:depth} & 14 & 64 & 24 & 1 & All 40 & Absolute & Resize & Triton\\
Depth Anything V2-L~\cite{yang2024dav2} & \cref{tab:depth} & 14 & 64 & 16 & 1 & All 24 & Absolute & Resize & Triton\\
Qwen3-VL-4B~\cite{bai2025qwen3vl} & \cref{tab:vlm} & 16 & 64 & 16 & 0 & All 24 & Absolute, 2D RoPE & Pad & Generic\\
Qwen3-VL-8B~\cite{bai2025qwen3vl} & \cref{tab:vlm} & 16 & 72 & 16 & 0 & All 27 & Absolute, 2D RoPE & Pad & Generic\\
GLM-4.6V-Flash~\cite{glm2025glm45v} & \cref{tab:vlm} & 14 & 128 & 12 & 0 & All 24 & Absolute, 2D RoPE & Pad & Generic\\
Kimi-VL-A3B~\cite{kimi2025kimivl} & \cref{tab:vlm} & 14 & 72 & 16 & 0 & All 27 & Absolute, 2D RoPE & Resize & Generic\\
\bottomrule
\end{tabular}\end{widewrap}
\end{table*}

\noindent\textbf{How \meros enters each model.}
\Cref{tab:supp_models} lists the patch size, layers, grid handling and kernel (\cref{sec:supp_kernel}) of each model in \cref{sec:exp}.
\meros replaces every global attention layer of a model and leaves windowed layers, such as those of ViTMatte and SAM, unchanged.
It reads queries and keys after the model's own position encoding and adds relative position biases for the keys it reads (\cref{sec:supp_grid}).
The depth models run on the H200 with the Triton kernel.

\subsection{Depth Estimation}
\label{sec:supp_depth}

\noindent\textbf{Timing.}
We time the ViT encoder alone at batch~1 on one H200, whereas \cref{fig:qual}(a) times the whole depth model, with its unchanged decoder, in a separate run.
Each image gets one warm-up call and three timed calls, and its encoder time is the median of the three.
The speedup is the mean encoder time of exact attention divided by that of the method.

\begin{table}[!tp]
\centering
\caption{\textbf{Settings of the depth baselines} of \cref{tab:depth} on MoGe-3 ViT-G~\cite{kong2026moge3} and Depth Anything V2-L~\cite{yang2024dav2}. For each method, we report the setting with the lowest mean ratio of its AbsRel to that of exact attention. The mean runs over the cells of that table in which every setting of the method ran. The last column lists the other settings that we evaluated.}
\setlength{\mcolw}{\dimexpr\textwidth*1000/961\relax}\renewcommand{\mtarget}{\textwidth}
\begin{colwrap}
\label{tab:supp_depth_settings}
\tabsetup{7.4}
\renewcommand{\arraystretch}{1.08}
\setlength{\tabcolsep}{3pt}
\setlength{\tabw}{\dimexpr\linewidth-2\tabcolsep\relax}%
\begin{tabular}{>{\raggedright\arraybackslash}p{\dimexpr(\tabw-6\tabcolsep)*27/100\relax}>{\raggedright\arraybackslash}p{\dimexpr(\tabw-6\tabcolsep)*14/100\relax}>{\raggedright\arraybackslash}p{\dimexpr(\tabw-6\tabcolsep)*40/100\relax}>{\raggedright\arraybackslash}p{\dimexpr(\tabw-6\tabcolsep)*19/100\relax}}
\toprule
\rowcolor{tabhead}[\tabcolsep][\rov] \textbf{Method} & \textbf{Setting} & \textbf{What the setting does} & \textbf{Also run}\\
\midrule
\rowcolor{tabsel}[\tabcolsep][\rov] NATTEN~\cite{hassani2026gnafindings} & k48 & Each patch query attends to the $48{\times}48$ patches around it & k32, k16\\
\rowcolor{tabsel}[\tabcolsep][\rov] XAttention~\cite{xu2025xattention} & s8t0.95 & Antidiagonal block scores with stride 8, blocks kept up to a threshold of 0.95 & s16t0.8\\
\rowcolor{tabsum}[\tabcolsep][\rov] i-clustered~\cite{vyas2020clustered} & 200 & 200 query clusters, each query exact on the top 32 keys of its cluster, in FP32 & 100\\
\rowcolor{tabsum}[\tabcolsep][\rov] Expedit~\cite{liang2022expediting} & default & Official ViT-L setting, tokens clustered after layer 10 of 24 to 0.7 of the grid side, same relative depth in ViT-G & aggressive\\
\rowcolor{tabapx}[\tabcolsep][\rov] HyperAttn~\cite{han2024hyperattention} & p25 & Last 25\% of the layers replaced, 7 hash projections, blocks and samples of 256 keys & p100\\
\bottomrule
\end{tabular}
\end{colwrap}
\end{table}

\noindent\textbf{Baselines in their best setting.}
Every baseline runs its official code, attached to both models in the same way.
We evaluate two or three settings per method and report the one selected by the rule in the caption of \cref{tab:supp_depth_settings}.
In NATTEN and HyperAttention, the class token attends to all keys exactly.
Each patch query then adds the class token as one more key, merged with the method's output through the softmax normalizer that the method returns.
The other methods receive the class token and the patches as one sequence.
Because the i-clustered kernel supports only FP32, it runs in FP32 inside the attention call.
The default setting of Expedit runs out of memory on ETH3D, as marked in \mainref{tab:depth}.
Its aggressive setting fits and reaches an AbsRel of 4.53\% at 4.1$\times$ the encoder speed on MoGe-3 and 6.93\% at 3.4$\times$ on Depth Anything V2.
On both models, $r{=}8$ is more accurate and faster than this setting (\mainref{tab:depth}).
With its default parameters, the official SpargeAttn kernel stalls on these inputs and exceeds our limit of 300\,s per image.
Calling the kernel directly on saved queries, keys and values reproduces the stall, which occurs in its default skipping of value products.
\mainref{tab:depth} therefore has no SpargeAttn row.\PFILL

\noindent\textbf{All 454 ETH3D images.}
\Cref{tab:supp_eth3d_full} evaluates MoGe-3 on all 454 images of ETH3D, with the methods, settings and metrics of \mainref{tab:depth}.
Every budget of \meros stays within 0.03 points of the AbsRel of exact attention.
Its $\delta_1$ stays within 0.05 points, and its RMSE within 2\,mm.
Every baseline has a higher AbsRel than any budget of \meros, from 4.64\% for HyperAttention to 24.74\% for NATTEN.
The encoder speedups here are timed on the first 10 ETH3D images, as in \mainref{tab:depth}.\PUP

\begin{table}[!tp]
\begin{minipage}[t]{0.465\textwidth}%
\renewcommand{\mtarget}{\linewidth}
\setlength{\mcolw}{\dimexpr\linewidth*1000/961\relax}
\centering
\caption{\textbf{MoGe-3 ViT-G~\cite{kong2026moge3} on all 454 ETH3D images}~\cite{schops2017eth3d}, after scale-and-shift alignment to the ground truth, with the same baselines, settings and metrics as in \cref{tab:depth}. AbsRel and $\delta_1$ are given in \% and RMSE is in meters. The encoder speedup over exact attention (Speed) is timed on one H200 on the first 10 images. Bold marks the best value in each column apart from exact attention (FA2).}
\begin{colwrap}
\label{tab:supp_eth3d_full}
\tabsetup{7.4}
\setlength{\tabcolsep}{2.4pt}
\setlength{\tabw}{\dimexpr\linewidth-2\tabcolsep\relax}%
\begin{tabular}{p{0.36\tabw}*{4}{>{\centering\arraybackslash}p{\dimexpr(0.64\tabw-8\tabcolsep)/4\relax}}}
\toprule
\rowcolor{tabhead}[\tabcolsep][\rov] \bfseries Method & \bfseries\boldmath AbsRel$\downarrow$ & \bfseries\boldmath RMSE$\downarrow$ & \bfseries\boldmath $\delta_1\uparrow$ & \bfseries\boldmath Speed\rlap{$\uparrow$}\phantom{$\downarrow$}\\
\midrule
Exact (FA2) & 3.63 & 0.493 & 98.99 & 1.00$\times$\\
\rowcolor{tabsel}[\tabcolsep][\rov] NATTEN~\cite{hassani2026gnafindings} & 24.74 & 2.669 & 55.17 & \textbf{9.43}$\times$\\
\rowcolor{tabsel}[\tabcolsep][\rov] XAttention~\cite{xu2025xattention} & 5.11 & 0.702 & 97.26 & 1.17$\times$\\
\rowcolor{tabsum}[\tabcolsep][\rov] i-clustered~\cite{vyas2020clustered} & 8.79 & 1.125 & 90.08 & 2.79$\times$\\
\rowcolor{tabsum}[\tabcolsep][\rov] Expedit~\cite{liang2022expediting} & \multicolumn{4}{c}{\textit{out of memory}}\\
\rowcolor{tabapx}[\tabcolsep][\rov] HyperAttn~\cite{han2024hyperattention} & 4.64 & 0.596 & 97.79 & 1.29$\times$\\
\midrule
\rowcolor{tabmeros}[\tabcolsep][\rov] \meros $r{=}2$ & \textbf{3.64} & 0.493 & 98.95 & 2.39$\times$\\
\rowcolor{tabmeros}[\tabcolsep][\rov] \meros $r{=}4$ & \textbf{3.64} & \textbf{0.492} & \textbf{99.04} & 4.26$\times$\\
\rowcolor{tabmeros}[\tabcolsep][\rov] \meros $r{=}8$ & 3.66 & \textbf{0.492} & 98.97 & 6.75$\times$\\
\bottomrule
\end{tabular}
\end{colwrap}
\end{minipage}\hfill
\begin{minipage}[t]{0.505\textwidth}%
\renewcommand{\mtarget}{\linewidth}
\setlength{\mcolw}{\dimexpr\linewidth*1000/961\relax}
\centering
\caption{\textbf{Token merging at matched MACs.} We count backbone MACs analytically from the tokens in each block and give them relative to exact attention. For each backbone, we show the most accurate setting of ToMe~\cite{bolya2023tome} or PiToMe~\cite{tran2024pitome} we tested that removes at least as many MACs as \meros $r{=}8$. ToMe merges the given number of tokens per block, and PiToMe keeps the given ratio. The change in mIoU from exact attention ($\Delta$mIoU) uses the frozen heads of \cref{tab:seg}. Bold marks the smaller mIoU loss per row.}
\begin{colwrap}
\label{tab:supp_tome_macs}
\tabsetup{7.4}
\setlength{\tabcolsep}{1.5pt}
\setlength{\tabw}{\dimexpr\linewidth-2\tabcolsep\relax}%
\begin{tabular}{p{0.24\tabw}>{\centering\arraybackslash}p{0.09\tabw}>{\centering\arraybackslash}p{0.15\tabw}p{0.2\tabw}>{\centering\arraybackslash}p{0.09\tabw}>{\centering\arraybackslash}p{\dimexpr0.23\tabw-10\tabcolsep\relax}}
\toprule
\rowcolor{tabhead}[\tabcolsep][\rov] & \multicolumn{2}{c}{\bfseries\boldmath \meros $r{=}8$} & \multicolumn{3}{c}{\bfseries Best merge with fewer MACs}\\
\rowcolor{tabhead}[\tabcolsep][\rov] \bfseries Backbone & \bfseries MACs & \bfseries\boldmath $\Delta$mIoU$\uparrow$ & \bfseries Merge & \bfseries MACs & \bfseries\boldmath $\Delta$mIoU$\uparrow$\\
\midrule
\rowcolor{tabsel}[\tabcolsep][\rov] \multicolumn{6}{l}{\textit{Cityscapes~\cite{cordts2016cityscapes}, tiles of 1024\,px}}\\
MAE-B~\cite{he2022mae} & 1/1.63 & $\mathbf{-0.15}$ & ToMe 256 & 1/1.82 & $-$1.82\\
MAE-L~\cite{he2022mae} & 1/1.49 & $\mathbf{-0.56}$ & ToMe 128 & 1/1.80 & $-$1.66\\
\mbox{DINOv2-B}~\cite{oquab2024dinov2} & 1/1.95 & $\mathbf{-1.06}$ & PiToMe 0.90 & 1/2.13 & $-$3.39\\
\rowcolor{tabsel}[\tabcolsep][\rov] \multicolumn{6}{l}{\textit{UAVid~\cite{lyu2020uavid}, tiles of 2048\,px}}\\
MAE-L~\cite{he2022mae} & 1/2.68 & $\mathbf{-0.13}$ & PiToMe 0.92 & 1/3.22 & $-$10.17\\
DeiT\,III-L~\cite{touvron2022deit3} & 1/2.68 & $\mathbf{-0.05}$ & ToMe 768 & 1/2.85 & $-$8.92\\
\bottomrule
\end{tabular}
\end{colwrap}
\end{minipage}
\end{table}

\subsection{Semantic Segmentation}
\label{sec:supp_seg}

\mainref{tab:seg} evaluates frozen backbones with a linear head on four datasets.
Each setting fixes the backbone, the input mode and the head, so that only the attention method changes from one table column to the next.

\noindent\textbf{Seeds.}
Exact attention involves no sampling and runs once.
\meros runs with the sampling seeds 17, 29 and 43, each of which redraws the sample of every layer and head.
\mainref{tab:seg} reports the mean over these seeds, and \cref{tab:supp_seg_seeds} adds the standard deviation.
Over the 51 pairs of setting and budget (17 settings and three budgets), this standard deviation has a median of 0.04~mIoU and a maximum of 0.26, on DeepGlobe tiles at $r{=}8$.
This maximum is below half the median loss of the most accurate baseline in \mainref{tab:seg}, SpargeAttn at 0.61~mIoU.\PFILL

\begin{table}[!tp]
\begin{minipage}[t]{0.485\textwidth}%
\renewcommand{\mtarget}{\linewidth}
\setlength{\extrarowheight}{1.2pt}
\setlength{\mcolw}{\dimexpr\linewidth*1000/961\relax}
\centering
\caption{\textbf{Seed variability of \meros} in \cref{tab:seg}. The mIoU change $\Delta$ from exact attention, as printed in that table, is the mean over the sampling seeds 17, 29 and 43. We add the standard deviation (SD) of the mIoU over these seeds. Inputs are tiles (T), a canvas (C) or the native size (N), sized in px.}
\begin{colwrap}
\label{tab:supp_seg_seeds}
\tabsetup{7.4}
\setlength{\tabcolsep}{1.5pt}
\setlength{\tabw}{\dimexpr\linewidth-2\tabcolsep\relax}%
\begin{tabular}{p{0.255\tabw}p{0.105\tabw}*{6}{>{\centering\arraybackslash}p{\dimexpr(0.64\tabw-14\tabcolsep)/6\relax}}}
\toprule
\rowcolor{tabhead}[\tabcolsep][\rov] \multicolumn{2}{l}{\bfseries \meros} & \multicolumn{2}{c}{\bfseries\boldmath $r{=}2$} & \multicolumn{2}{c}{\bfseries\boldmath $r{=}4$} & \multicolumn{2}{c}{\bfseries\boldmath $r{=}8$}\\
\rowcolor{tabhead}[\tabcolsep][\rov] \bfseries Backbone & \bfseries Input & \bfseries\boldmath $\Delta$ & \bfseries SD & \bfseries\boldmath $\Delta$ & \bfseries SD & \bfseries\boldmath $\Delta$ & \bfseries SD\\
\midrule
\rowcolor{tabsel}[\tabcolsep][\rov] \multicolumn{8}{l}{\textit{UAVid~\cite{lyu2020uavid}}}\\
DINOv3-B~\cite{simeoni2025dinov3} & T\,2048 & $-$0.09 & .00 & $-$0.17 & .03 & $-$0.33 & .04\\
MAE-B~\cite{he2022mae} & T\,2048 & $+$0.02 & .01 & $+$0.02 & .00 & $+$0.00 & .02\\
MAE-L~\cite{he2022mae} & T\,2048 & $-$0.01 & .01 & $-$0.04 & .03 & $-$0.13 & .02\\
DeiT\,III-B~\cite{touvron2022deit3} & T\,2048 & $+$0.01 & .01 & $+$0.00 & .00 & $+$0.05 & .02\\
DeiT\,III-L~\cite{touvron2022deit3} & T\,2048 & $+$0.00 & .02 & $-$0.01 & .01 & $-$0.05 & .03\\
SigLIP2-B~\cite{tschannen2025siglip2} & T\,2048 & $-$0.14 & .12 & $-$0.31 & .18 & $-$0.36 & .08\\
SigLIP2-L~\cite{tschannen2025siglip2} & T\,2048 & $-$0.03 & .05 & $-$0.05 & .04 & $-$0.33 & .16\\
\rowcolor{tabsel}[\tabcolsep][\rov] \multicolumn{8}{l}{\textit{Cityscapes~\cite{cordts2016cityscapes}}}\\
DINOv3-B~\cite{simeoni2025dinov3} & T\,1024 & $-$0.16 & .06 & $-$0.45 & .06 & $-$1.11 & .06\\
MAE-B~\cite{he2022mae} & T\,1024 & $-$0.01 & .01 & $-$0.09 & .07 & $-$0.15 & .05\\
MAE-L~\cite{he2022mae} & T\,1024 & $-$0.09 & .02 & $-$0.27 & .04 & $-$0.56 & .03\\
\rowcolor{tabsel}[\tabcolsep][\rov] \multicolumn{8}{l}{\textit{DeepGlobe~\cite{demir2018deepglobe}}}\\
DINOv3-B~\cite{simeoni2025dinov3} & T\,2048 & $-$0.07 & .03 & $-$0.29 & .10 & $-$0.57 & .26\\
DINOv3-B~\cite{simeoni2025dinov3} & C\,2560 & $+$0.07 & .06 & $-$0.03 & .09 & $-$0.38 & .25\\
DINOv3-B~\cite{simeoni2025dinov3} & N\,2448 & $-$0.08 & .21 & $-$0.23 & .11 & $-$0.62 & .25\\
\rowcolor{tabsel}[\tabcolsep][\rov] \multicolumn{8}{l}{\textit{URUR~\cite{ji2023urur}}}\\
DINOv3-B~\cite{simeoni2025dinov3} & T\,2048 & $-$0.21 & .04 & $-$0.53 & .08 & $-$0.95 & .02\\
DINOv3-B~\cite{simeoni2025dinov3} & T\,4096 & $-$0.10 & .06 & $-$0.23 & .03 & $-$0.30 & .06\\
DINOv3-B~\cite{simeoni2025dinov3} & C\,5120 & $-$0.11 & .01 & $-$0.19 & .04 & $-$0.41 & .05\\
DINOv3-B~\cite{simeoni2025dinov3} & N\,5120 & $-$0.02 & .10 & $-$0.09 & .12 & $-$0.17 & .10\\
\midrule
\rowcolor{tabmeros}[\tabcolsep][\rov] \multicolumn{2}{l}{Median of 17 settings} & $-$0.07 & .03 & $-$0.17 & .04 & $-$0.33 & .05\\
\rowcolor{tabmeros}[\tabcolsep][\rov] \multicolumn{2}{l}{Largest SD} & & .21 & & .18 & & .26\\
\bottomrule
\end{tabular}
\end{colwrap}
\end{minipage}\hfill
\begin{minipage}[t]{0.485\textwidth}%
\renewcommand{\mtarget}{\linewidth}
\setlength{\mcolw}{\dimexpr\linewidth*1000/961\relax}
\centering
\caption{\textbf{All settings of token merging} on MAE-L and DeiT\,III-L behind \cref{tab:tome}, on UAVid~\cite{lyu2020uavid} with tiles of $2048^2$\,px. We give backbone time per tile (ms) at batch~1 on one B300 and peak memory (GB). A dash means not run. The best value apart from exact attention is bold. Every setting of ToMe and PiToMe is slower than exact attention, whereas every budget of \meros is at least 3$\times$ faster per tile than exact attention.}
\begin{colwrap}
\label{tab:supp_tome}
\tabsetup{7.4}
\setlength{\tabcolsep}{2pt}
\setlength{\tabw}{\dimexpr\linewidth-2\tabcolsep\relax}%
\begin{tabular}{p{0.27\tabw}*{6}{>{\centering\arraybackslash}p{\dimexpr(0.73\tabw-12\tabcolsep)/6\relax}}}
\toprule
\rowcolor{tabhead}[\tabcolsep][\rov] & \multicolumn{3}{c}{\bfseries\boldmath MAE-L~\cite{he2022mae}} & \multicolumn{3}{c}{\bfseries\boldmath DeiT\,III-L~\cite{touvron2022deit3}}\\
\rowcolor{tabhead}[\tabcolsep][\rov] \bfseries Method & \bfseries mIoU$\uparrow$ & \bfseries ms$\downarrow$ & \bfseries GB$\downarrow$ & \bfseries mIoU$\uparrow$ & \bfseries ms$\downarrow$ & \bfseries GB$\downarrow$\\
\midrule
Exact (FA2) & 49.11 & 87.1 & 1.02 & 55.95 & 86.4 & 1.06\\
\rowcolor{tabsum}[\tabcolsep][\rov] \multicolumn{7}{l}{\textit{PiToMe~\cite{tran2024pitome}, keep ratio per block}}\\
0.995 & 48.83 & 589 & 20.4 & 55.51 & 537 & 21.0\\
0.99 & 48.43 & 541 & 20.1 & 55.04 & 496 & 20.7\\
0.98 & 47.51 & 471 & 19.5 & 54.03 & 436 & 19.7\\
0.96 & 45.21 & 346 & 18.5 & 51.53 & 325 & 19.0\\
0.94 & 42.39 & 259 & 17.9 & 48.43 & 246 & 18.4\\
0.92 & 38.94 & 215 & 17.3 & 44.87 & 206 & 17.8\\
0.90 & 35.81 & 175 & 16.9 & 41.63 & 169 & 17.2\\
\rowcolor{tabsum}[\tabcolsep][\rov] \multicolumn{7}{l}{\textit{ToMe~\cite{bolya2023tome}, merged tokens per block}}\\
8 & 49.02 & 711 & 20.7 & 55.96 & 859 & 20.3\\
16 & 48.94 & 702 & 20.7 & 55.97 & 848 & 20.2\\
32 & 48.80 & 684 & 20.6 & 55.98 & 827 & 20.1\\
64 & 48.49 & 660 & 20.5 & 55.99 & 797 & 20.6\\
128 & 47.84 & 602 & 20.2 & \textbf{56.00} & 726 & 20.3\\
256 & 46.78 & 499 & 19.8 & 55.96 & 600 & 19.3\\
512 & 43.32 & 348 & 19.2 & 55.60 & 413 & 18.8\\
655 & 36.57 & 258 & 18.5 & -- & -- & --\\
768 & -- & -- & -- & 47.03 & 288 & 18.3\\
1024 & -- & -- & -- & 15.21 & 221 & 17.8\\
1280 & -- & -- & -- & 10.70 & 180 & 17.4\\
\midrule
\rowcolor{tabmeros}[\tabcolsep][\rov] \meros $r{=}2$ & \textbf{49.10} & 28.8 & 1.30 & 55.95 & 28.4 & 1.32\\
\rowcolor{tabmeros}[\tabcolsep][\rov] \meros $r{=}4$ & 49.06 & 22.8 & 1.27 & 55.94 & 22.3 & \textbf{1.29}\\
\rowcolor{tabmeros}[\tabcolsep][\rov] \meros $r{=}8$ & 48.97 & \textbf{20.5} & \textbf{1.24} & 55.91 & \textbf{20.0} & \textbf{1.29}\\
\bottomrule
\end{tabular}
\end{colwrap}
\end{minipage}
\end{table}

\noindent\textbf{A further backbone.}
We also run \mbox{DINOv2-B}~\cite{oquab2024dinov2}, with patches of 14\,px, on UAVid and Cityscapes with the protocol of \mainref{tab:seg}.
Its tiles of 2048 and 1024\,px are padded to 2240 and 1120\,px, which gives grids of 160 and 80 patches.
Because the tile offsets are not multiples of 14\,px, we upsample the logits of each tile to pixels before averaging overlapping tiles.
As in mode N, padded patches take part in attention and receive no label.
At $r{=}8$, \meros loses 0.34~mIoU on UAVid and 1.06 on Cityscapes, while the backbone runs 5.37$\times$ and 1.65$\times$ faster (\cref{tab:supp_seg_dinov2}).
No loss in \cref{tab:supp_seg_dinov2} exceeds the worst of the 17 settings of \mainref{tab:seg} at the same budget.
Over the three seeds, the standard deviation of the mIoU of \meros is at most 0.06 on UAVid and 0.20 on Cityscapes at every budget.

\begin{table}[!tp]
\centering
\caption{\textbf{Training-free segmentation with \mbox{DINOv2-B}}~\cite{oquab2024dinov2} under the protocol of \cref{tab:seg}. We give the mIoU of exact attention and the mIoU change of \meros, averaged over the seeds 17, 29 and 43. The backbone speedups over exact attention come from \cref{tab:supp_backbones}. For each budget, the last row repeats the worst mIoU change of the 17 settings of \cref{tab:seg}.}
\setlength{\mcolw}{\dimexpr\textwidth*1000/961\relax}\renewcommand{\mtarget}{\textwidth}
\begin{colwrap}
\label{tab:supp_seg_dinov2}
\tabsetup{7.4}
\setlength{\tabcolsep}{1.5pt}
\setlength{\tabw}{\dimexpr\linewidth-2\tabcolsep\relax}%
\begin{tabular}{p{0.24\tabw}>{\centering\arraybackslash}p{0.1\tabw}*{6}{>{\centering\arraybackslash}p{\dimexpr(0.66\tabw-14\tabcolsep)/6\relax}}}
\toprule
\rowcolor{tabhead}[\tabcolsep][\rov] & & \multicolumn{3}{c}{\bfseries\boldmath $\Delta$mIoU\,$\uparrow$} & \multicolumn{3}{c}{\bfseries\boldmath Speed\rlap{$\uparrow$}\phantom{$\downarrow$}}\\
\rowcolor{tabhead}[\tabcolsep][\rov] \bfseries Dataset & \bfseries Exact & \bfseries\boldmath $r{=}2$ & \bfseries\boldmath $r{=}4$ & \bfseries\boldmath $r{=}8$ & \bfseries\boldmath $r{=}2$ & \bfseries\boldmath $r{=}4$ & \bfseries\boldmath $r{=}8$\\
\midrule
\rowcolor{tabsel}[\tabcolsep][\rov] \multicolumn{8}{l}{\textit{\mbox{DINOv2-B}~\cite{oquab2024dinov2}}}\\
UAVid~\cite{lyu2020uavid} & 63.67 & $-$0.05 & $-$0.14 & $-$0.34 & 3.47$\times$ & 4.57$\times$ & 5.37$\times$\\
Cityscapes~\cite{cordts2016cityscapes} & 73.94 & $-$0.19 & $-$0.48 & $-$1.06 & 1.50$\times$ & 1.60$\times$ & 1.65$\times$\\
\midrule
\rowcolor{tabmeros}[\tabcolsep][\rov] \multicolumn{2}{l}{Worst of 17 settings} & $-$0.21 & $-$0.53 & $-$1.11 & 1.30$\times$ & 1.38$\times$ & 1.46$\times$\\
\bottomrule
\end{tabular}
\end{colwrap}
\end{table}

\subsection{Token Merging}
\label{sec:supp_tome}

\mainref{tab:tome} compares \meros with ToMe~\cite{bolya2023tome} and PiToMe~\cite{tran2024pitome}, which remove tokens from the sequence.
All methods run on UAVid with the 2048-px tiles and frozen heads of \mainref{tab:seg}.
ToMe merges a fixed number of tokens in every block, whereas PiToMe keeps a fixed fraction of the tokens.
We sweep ToMe from 8 to 1280 merged tokens per block, and PiToMe from a keep ratio of 0.995 down to 0.90 per block.
Both track which patches each merged token absorbs, and before the frozen head we copy every merged token back to these source patches.\PFILL

\noindent\textbf{Matching.}
\mainref{tab:tome} shows one setting per merging method, and we measure cost in multiply-accumulate operations (MACs).
For PiToMe, we take the largest keep ratio whose MACs do not exceed those of $r{=}8$.
We count MACs analytically from the number of tokens in each block.
For \meros, the MACs of the two attention products scale with the keys per query of \cref{eq:supp_keys}, and all other MACs stay unchanged.
The rule selects 0.92 on MAE-L and DeiT\,III-L. On MAE-B, the closest ratio of our sweep is 0.90, which still spends 1.3 times the MACs of $r{=}8$.
For ToMe on MAE-L and DeiT\,III-L, we take the fastest setting within 0.1~mIoU of exact attention, which merges 8 tokens per block on MAE-L and 256 on DeiT\,III-L.
These are the ToMe rows of \mainref{tab:tome}.\PFILL

\noindent\textbf{All settings.}
\Cref{tab:supp_tome} lists every setting on MAE-L and DeiT\,III-L.
Each of these 32 settings of ToMe and PiToMe takes 2.0 to 9.9 times as long as exact attention.
They also reserve 16 to 20 times the peak memory of exact attention.
On MAE-L, only ToMe with 8 merged tokens per block stays within 0.1~mIoU of exact attention.
This setting takes 35 times as long as $r{=}8$.
Merging more tokens makes ToMe faster, but its fastest setting on DeiT\,III-L, 1280 merged tokens per block, still takes 180\,ms per tile at 10.70~mIoU.\PFILL

\noindent\textbf{More settings.}
We also ran both methods on eight settings of \mainref{tab:seg} and on \mbox{DINOv2-B} on UAVid and Cityscapes.
The eight are \mbox{MAE-B} and \mbox{MAE-L} on UAVid and Cityscapes, and DeiT\,III-B, DeiT\,III-L, SigLIP2-B and SigLIP2-L on UAVid.
\Cref{tab:supp_tome_macs} lists the five settings in which at least one merge we tested removes at least as many MACs as $r{=}8$. In the other five, none does.
In the listed five, $r{=}8$ is 1.1 to 10.0~mIoU more accurate than the best such merge.
PiToMe is also slower than exact attention in five settings, which are DeiT\,III-B, SigLIP2-B and SigLIP2-L with 2048-px tiles, and \mbox{MAE-B} and \mbox{MAE-L} with 1024-px tiles.
There, every keep ratio we timed runs at 0.12 to 0.46 times the speed of FA2.

\subsection{Dichotomous Segmentation}
\label{sec:supp_dis}

\mainref{tab:dis} tests fine structures on DIS5K~\cite{qin2022dis}.
Its training split DIS-TR has 3{,}000 images and its validation split DIS-VD has 470.
Each of the four test splits, DIS-TE1--4, has 500 images of up to 65.7 megapixels (MP).

\noindent\textbf{All budgets.}
\Cref{tab:supp_dis} adds $r{=}4$ to the \meros rows of \mainref{tab:dis}.
Relative to exact attention, the largest drop in $F^{\max}_\beta$ is 0.007, for \mbox{DINOv3-B} at 2048\,px with $r{=}8$.
Across these settings, $r{=}8$ runs the backbone 2.93 to 5.35$\times$ faster than exact attention.
For \mbox{DINOv3-B} at 2048\,px, the standard deviation of $r{=}8$ over three seeds is at most 0.001 on $F^{\max}_\beta$, $F^{w}_\beta$, $M$, $S_\alpha$ and $E^m_\phi$ and 0.4 on HCE$_\gamma$.
\Cref{fig:supp_gal_seg}(a) compares exact attention at 1024\,px with $r{=}8$ at 2048\,px, at about equal latency, where the IoU of $r{=}8$ to the reference is 0.92 and 0.97, against 0.59 and 0.83.

\begin{table}[!tp]
\centering
\caption{\textbf{Dichotomous segmentation at every budget} on the 2{,}000 DIS-TE images~\cite{qin2022dis}, with the frozen head of \cref{tab:dis}. Backbone latency (ms) is measured on one B300 at the input size of each block, with its speedup over exact attention (FA2). Shaded rows are \meros at budget $r$, the best budget in each block is bold, and all metrics are the same as in \cref{tab:dis}.}
\setlength{\mcolw}{\dimexpr\textwidth*1000/961\relax}\renewcommand{\mtarget}{\textwidth}
\begin{colwrap}
\label{tab:supp_dis}
\tabsetup{6.8}
\setlength{\tabcolsep}{1pt}
\setlength{\tabw}{\dimexpr\linewidth-2\tabcolsep\relax}%
\begin{tabular}{>{\raggedright\arraybackslash}p{\dimexpr0.08\linewidth-2\tabcolsep\relax}>{\centering\arraybackslash}p{\dimexpr0.14\linewidth-2\tabcolsep\relax}>{\centering\arraybackslash}p{\dimexpr0.11\linewidth-2\tabcolsep\relax}>{\centering\arraybackslash}p{\dimexpr0.09\linewidth-2\tabcolsep\relax}>{\centering\arraybackslash}p{\dimexpr0.11\linewidth-2\tabcolsep\relax}>{\centering\arraybackslash}p{\dimexpr0.11\linewidth-2\tabcolsep\relax}>{\centering\arraybackslash}p{\dimexpr0.12\linewidth-2\tabcolsep\relax}>{\centering\arraybackslash}p{\dimexpr0.10\linewidth-2\tabcolsep\relax}>{\centering\arraybackslash}p{\dimexpr0.14\linewidth-2\tabcolsep\relax}}
\toprule
\rowcolor{tabhead}[\tabcolsep][\rov] & \bfseries\boldmath $F^{\max}_\beta{\uparrow}$ & \bfseries\boldmath $F^{w}_\beta{\uparrow}$ & \bfseries\boldmath $M{\downarrow}$ & \bfseries\boldmath $S_\alpha{\uparrow}$ & \bfseries\boldmath $E^m_\phi{\uparrow}$ & \bfseries\boldmath HCE$_\gamma{\downarrow}$ & \bfseries\boldmath ms${\downarrow}$ & \bfseries\boldmath Speed\rlap{$\uparrow$}\phantom{$\downarrow$}\\
\midrule
\rowcolor{tabsel}[\tabcolsep][\rov] \multicolumn{9}{l}{\textit{DINOv3-B~\cite{simeoni2025dinov3}, 1024\,px}}\\
Exact & .781 & .606 & .087 & .807 & .814 & 1443 & 7.2 & 1.00$\times$\\
\rowcolor{tabmeros}[\tabcolsep][\rov] $r{=}2$ & \textbf{.779} & \textbf{.605} & \textbf{.088} & \textbf{.806} & \textbf{.813} & 1441 & 3.0 & 2.43$\times$\\
\rowcolor{tabmeros}[\tabcolsep][\rov] $r{=}4$ & .778 & .603 & \textbf{.088} & .805 & .812 & 1440 & 2.6 & 2.75$\times$\\
\rowcolor{tabmeros}[\tabcolsep][\rov] $r{=}8$ & .775 & .598 & .090 & .803 & .808 & \textbf{1438} & \textbf{2.5} & \textbf{2.93}$\times$\\
\rowcolor{tabsel}[\tabcolsep][\rov] \multicolumn{9}{l}{\textit{DINOv3-B~\cite{simeoni2025dinov3}, 2048\,px}}\\
Exact & .816 & .624 & .085 & .823 & .828 & 1514 & 43.2 & 1.00$\times$\\
\rowcolor{tabmeros}[\tabcolsep][\rov] $r{=}2$ & \textbf{.815} & \textbf{.622} & \textbf{.086} & \textbf{.822} & \textbf{.827} & \textbf{1513} & 11.4 & 3.79$\times$\\
\rowcolor{tabmeros}[\tabcolsep][\rov] $r{=}4$ & .813 & .619 & .087 & .820 & .825 & 1515 & 9.1 & 4.75$\times$\\
\rowcolor{tabmeros}[\tabcolsep][\rov] $r{=}8$ & .809 & .614 & .088 & .818 & .821 & \textbf{1513} & \textbf{8.1} & \textbf{5.35}$\times$\\
\rowcolor{tabsel}[\tabcolsep][\rov] \multicolumn{9}{l}{\textit{MAE-L~\cite{he2022mae}, 2048\,px}}\\
Exact & .682 & .464 & .147 & .728 & .700 & 1556 & 87.1 & 1.00$\times$\\
\rowcolor{tabmeros}[\tabcolsep][\rov] $r{=}2$ & \textbf{.681} & \textbf{.463} & \textbf{.147} & \textbf{.728} & \textbf{.700} & \textbf{1555} & 28.7 & 3.03$\times$\\
\rowcolor{tabmeros}[\tabcolsep][\rov] $r{=}4$ & \textbf{.681} & \textbf{.463} & .148 & .727 & .699 & 1557 & 22.8 & 3.82$\times$\\
\rowcolor{tabmeros}[\tabcolsep][\rov] $r{=}8$ & .679 & .461 & .149 & .726 & .698 & \textbf{1555} & \textbf{20.5} & \textbf{4.25}$\times$\\
\bottomrule
\end{tabular}
\end{colwrap}
\end{table}

\subsection{Vision-Language Models}
\label{sec:supp_vlm}

\noindent\textbf{Timing.}
The time to the first token (TTFT) is the GPU time from the call to the first generated token.
It covers the vision encoder, the merger and the prefill of the language model, but not CPU preprocessing.
We measure it with CUDA events on one B300 at batch~1, on 20 HR-Bench 8K images at the input size of each model.
Each image runs once to warm up and three times for timing, and we report the median over images.
The speedup is the TTFT of exact attention divided by that of the method.
Exact attention (FA2 in \mainref{tab:vlm}) runs the released attention code of each model.
FA4 runs its official kernel in the encoder.\PFILL

\noindent\textbf{Where the time goes.}
\Cref{tab:supp_vlm_time} splits the TTFT of Qwen3-VL-8B into its parts.
With exact attention, the encoder takes 10.8 of the 12.6\,s.
Because token pruning after the encoder, as in FastV, PDrop, SparseVLM and VisionZip, shortens only the prefill, its gain is bounded.
\meros $r{=}8$ runs the encoder 5.75$\times$ faster.
Its encoder uses the generic Triton kernel (\cref{tab:supp_models}), not the CuTe kernel behind \cref{tab:eff}.
SparseVLM prunes tokens inside the language model, so it combines with \meros in the encoder. Adding it to $r{=}8$ cuts the prefill from 1.77 to 0.46\,s.
Window attention shortens the encoder to 2.05\,s, against 1.88\,s for $r{=}8$, but with the accuracy loss reported in \mainref{tab:vlm}.

\begin{table}[!tp]
\centering
\caption{\textbf{Where the time to the first token goes} for Qwen3-VL-8B~\cite{bai2025qwen3vl} on HR-Bench 8K~\cite{wang2025hrbench} at native resolution, median over 20 images on one B300. We split the TTFT of \cref{tab:vlm} into two parts, the time of the vision encoder, with its speedup over exact attention, and the prefill of the language model. All times are in seconds. The TTFT also includes the merger (about 2\,ms per image), so it slightly exceeds the sum of the two parts shown. The best value in each column is set in bold.}
\setlength{\mcolw}{\dimexpr\textwidth*1000/961\relax}\renewcommand{\mtarget}{\textwidth}
\begin{colwrap}
\label{tab:supp_vlm_time}
\tabsetup{7.2}
\setlength{\tabcolsep}{1.5pt}
\setlength{\tabw}{\dimexpr\linewidth-2\tabcolsep\relax}%
\begin{tabular}{>{\raggedright\arraybackslash}p{0.45\tabw}*{4}{>{\centering\arraybackslash}p{\dimexpr(0.55\tabw-8\tabcolsep)/4\relax}}}
\toprule
\rowcolor{tabhead}[\tabcolsep][\rov] & \multicolumn{2}{c}{\bfseries Encoder} & & \\
\rowcolor{tabhead}[\tabcolsep][\rov] \bfseries Method & \bfseries\boldmath s$\downarrow$ & \bfseries\boldmath Speed\rlap{$\uparrow$}\phantom{$\downarrow$} & \bfseries\boldmath Prefill$\downarrow$ & \bfseries\boldmath TTFT$\downarrow$\\
\midrule
Exact (FA2) & 10.83 & 1.00$\times$ & 1.76 & 12.60\\
FA4~\cite{zadouri2026flashattention} & 3.23 & 3.35$\times$ & 1.77 & 5.01\\
\rowcolor{tabsel}[\tabcolsep][\rov] FastV~\cite{chen2024fastv} & 10.84 & 1.00$\times$ & 0.68 & 11.52\\
\rowcolor{tabsel}[\tabcolsep][\rov] PDrop~\cite{xing2024pyramiddrop} & 10.84 & 1.00$\times$ & 0.69 & 11.53\\
\rowcolor{tabsel}[\tabcolsep][\rov] SparseVLM~\cite{zhang2025sparsevlm} & 10.83 & 1.00$\times$ & \textbf{0.46} & 11.30\\
\rowcolor{tabsel}[\tabcolsep][\rov] VisionZip~\cite{yang2025visionzip} & 11.95 & 0.91$\times$ & 0.63 & 12.58\\
\rowcolor{tabsum}[\tabcolsep][\rov] ToMe~\cite{bolya2023tome} & 15.21 & 0.71$\times$ & 1.77 & 16.98\\
\rowcolor{tabsum}[\tabcolsep][\rov] Window attention & 2.05 & 5.29$\times$ & 1.76 & 3.81\\
\midrule
\rowcolor{tabmeros}[\tabcolsep][\rov] \meros $r{=}8$ & \textbf{1.88} & \textbf{5.75}$\times$ & 1.77 & 3.66\\
\rowcolor{tabmeros}[\tabcolsep][\rov] \meros $r{=}8$ + SparseVLM & 1.89 & \textbf{5.75}$\times$ & \textbf{0.46} & \textbf{2.35}\\
\bottomrule
\end{tabular}
\end{colwrap}
\end{table}

\subsection{Image Matting}
\label{sec:supp_matting}

\noindent\textbf{Where \meros acts.}
The ViT-B backbone of ViTMatte follows ViTDet~\cite{li2022vitdet}.
Eight of its 12 blocks attend within windows, whereas blocks 2, 5, 8 and 11, counted from 0, attend globally.
\meros replaces the attention of these four global blocks and leaves the window blocks, the convolutional detail branch and the decoder unchanged.
ViTMatte has no class token or registers, so $\Glob$ is empty and each patch query reads only its 256 tile keys and the far samples.
In the four global blocks, the generic Triton kernel adds the decomposed relative position bias of \cref{sec:supp_grid} to the logit of every key it reads (\cref{sec:supp_softmax}).
The bias comes from the model's learned tables at the current grid size, so each of these keys keeps the logit it has in the original model.\PFILL

\noindent\textbf{Timing and memory.}
All timings in \mainref{tab:matting} use one B300 at batch~1.
For each image, the latency is the median of three calls timed with CUDA events after one untimed call.
Speedups are relative to exact attention in query chunks.
The GB column of \mainref{tab:matting} gives the largest per-image peak of reserved GPU memory.\PFILL

PISA~\cite{li2026pisa}, SpargeAttn~\cite{zhang2025sparge} and ClusterAttn~\cite{nordenram2026clusterattention} run the variants they have in \mainref{tab:tf}.
These are PISA~0.3 with density 0.3, SpargeAttn~0.5s with top-$k$ 0.5 and an attention sink, and ClusterAttn~1.0 with \texttt{k\_frac} 1.0.
None of the three kernels can add a bias to the logits.
ClusterAttn has no bias argument, and the bias option of PISA belongs to its own routing. SpargeAttn accepts a mask that its kernel never uses.
We therefore drop the bias inside the four global blocks for these three methods, as marked in \mainref{tab:matting}.
Dropping the bias alone raises the SAD of exact attention from 35.3 to 38.0 on PPM-100 (\cref{sec:exp_matting}).
ClusterAttn runs our Blackwell port of its kernel ($^\ddagger$ in \mainref{tab:matting}).

\noindent\textbf{Transparent-460.}
Transparent-460~\cite{cai2022transmatting} contains transparent and translucent objects such as glass, water, fire and webs.
Its test split provides 50 foregrounds with alpha mattes and trimaps, and we use these trimaps. We composite each foreground once, onto one official background, with the official composition code.
This gives 50 test images at native resolution, from 2.8 to 30 megapixels or 12{,}544 to 124{,}416 tokens.
We run ViTMatte-B as for \cref{tab:matting} and time it as above, on one B300.
Even with the 288\,GB of a B300, the original model runs out of memory on 38 of the 50 images and fits only those of at most 45{,}056 tokens.
Exact attention in query chunks and \meros never exceed 29\,GB of peak memory on any of these images.
On the two images of \cref{fig:qual}(d), \meros $r{=}8$ runs 4.6$\times$ and 5.6$\times$ faster than exact attention.
It also lowers the SAD from 176.9 to 175.8 and from 101.8 to 100.9.
Over all 50 images, \meros $r{=}8$ is 4.6$\times$ faster than exact attention in total.\PFILL

\section{Analysis}
\label{sec:supp_analysis}

The analyses of \cref{sec:method,sec:exp_ana} rest on the protocols and full results of \cref{sec:supp_anatomy,sec:supp_check,sec:supp_equal_budget,sec:supp_components}.

\subsection{Where the Attention Mass Lies}
\label{sec:supp_anatomy}

\begin{table}[!tp]
\centering
\tabsetup{7.0}
\setlength{\tabcolsep}{2.2pt}
\caption{\textbf{Where the attention mass lies} in \mbox{DINOv3-B}~\cite{simeoni2025dinov3} at 2048\,px, layers 6 and 11. (a)~Mass shares, the weight of a tile key over that of a far key, and $\neff$, as medians per image over all 16{,}384 patch queries and 12 heads. Far $\ge$ half is the share of query--head pairs whose far field holds at least half of the mass. (b)~Percentiles of the ratio of estimated to true far mass for 480 query--head pairs per layer. The selections and \meros read about one eighth of the far keys, cell centroids read one mean key per cell, and a ratio of 1 means that the far mass is estimated exactly. L stands for the index of the layer.}
\setlength{\mcolw}{\dimexpr\textwidth*1000/961\relax}\renewcommand{\mtarget}{\textwidth}
\begin{colwrap}
\label{tab:supp_anatomy}
\begin{tabular}{p{58pt}>{\centering\arraybackslash}p{11pt}*{5}{>{\centering\arraybackslash}p{\dimexpr(\linewidth-69pt-14\tabcolsep)/5\relax}}}
\toprule
\rowcolor{tabhead}[\tabcolsep][\rov] & & \multicolumn{5}{c}{\bfseries\boldmath (a) Median per image}\\
\hcmid\cmidrule(lr){3-7}
\rowcolor{tabhead}[\tabcolsep][\rov] \bfseries Quantity & \bfseries L & \bfseries Skyline & \bfseries Scaffold & \bfseries Ridge & \bfseries Village & \bfseries Cow\\
\midrule
Tile mass (\%) & 6 & 11 & 10 & 10 & 16 & 12\\
 & 11 & 7 & 7 & 6 & 11 & 7\\
Global mass (\%) & 6 & 5.9 & 3.9 & 3.2 & 3.2 & 5.5\\
 & 11 & 0.5 & 0.5 & 0.4 & 0.5 & 0.4\\
Far mass (\%) & 6 & 72 & 80 & 77 & 74 & 73\\
 & 11 & 89 & 89 & 90 & 84 & 89\\
Far $\ge$ half (\%) & 6 & 74 & 82 & 81 & 76 & 77\\
 & 11 & 86 & 90 & 87 & 85 & 87\\
Tile/far weight & 6 & 8.8 & 7.9 & 8.3 & 13.5 & 10.4\\
 & 11 & 4.9 & 4.9 & 4.3 & 7.5 & 4.6\\
$\neff$ (of 1{,}008) & 6 & 56 & 78 & 62 & 46 & 53\\
 & 11 & 129 & 171 & 147 & 98 & 118\\
\midrule
\rowcolor{tabhead}[\tabcolsep][\rov] & & \multicolumn{5}{c}{\bfseries\boldmath (b) Estimate / true far mass}\\
\hcmid\cmidrule(lr){3-7}
\rowcolor{tabhead}[\tabcolsep][\rov] \bfseries Estimate & \bfseries L & \bfseries 5th & \bfseries 25th & \bfseries 50th & \bfseries 75th & \bfseries 95th\\
\midrule
\rowcolor{tabsel}[\tabcolsep][\rov] Oracle top 1/8 & 6 & .489 & .752 & .948 & .995 & 1.000\\
\rowcolor{tabsel}[\tabcolsep][\rov] & 11 & .407 & .630 & .843 & .951 & .995\\
\rowcolor{tabsel}[\tabcolsep][\rov] Nearest keys & 6 & .120 & .364 & .596 & .853 & .976\\
\rowcolor{tabsel}[\tabcolsep][\rov] & 11 & .087 & .234 & .466 & .764 & .957\\
\rowcolor{tabsel}[\tabcolsep][\rov] Larger window & 6 & .109 & .348 & .566 & .847 & .977\\
\rowcolor{tabsel}[\tabcolsep][\rov] & 11 & .085 & .230 & .440 & .753 & .947\\
\rowcolor{tabsum}[\tabcolsep][\rov] Cell centroids & 6 & .379 & .642 & .747 & .832 & .901\\
\rowcolor{tabsum}[\tabcolsep][\rov] & 11 & .390 & .654 & .828 & .912 & .958\\
\rowcolor{tabmeros}[\tabcolsep][\rov] \meros $r{=}8$ & 6 & .859 & .968 & 1.003 & 1.038 & 1.162\\
\rowcolor{tabmeros}[\tabcolsep][\rov] & 11 & .914 & .977 & 1.001 & 1.024 & 1.104\\
\bottomrule
\end{tabular}
\end{colwrap}
\end{table}

\Cref{sec:anatomy,sec:select} report how the attention mass divides between the tile, the global tokens and the far field, and how much of the far mass selection, summarization and \meros recover.
\Cref{tab:supp_anatomy} gives the full distribution.

\noindent\textbf{Setup.}
We run \mbox{DINOv3-B}~\cite{simeoni2025dinov3} with exact attention on five photographs of a city skyline, a statue under scaffolding, a mountain ridge, a village and a cow.
Each photograph is cropped to a square and resized to 2048\,px.
Every patch query then has 256 tile keys, five global tokens and 16{,}128 far keys in 1{,}008 cells.
We take the queries and keys of layers 6 and 11 (the last) after the rotary position embedding, with FP32 weights under BF16 autocast.
For all 16{,}384 patch queries and 12 heads, we compute the shares of the total mass in the tile, the global tokens and the far field.
The same pass gives the $\neff$ of \mainref{eq:neff} for every query and head.
For the far-field estimates, we also store the full attention rows of eight queries per image, with logits and softmax computed in FP32.
We pick them from the final features of exact attention, so the choice does not depend on \meros.
Four lie on salient objects, where the features differ most from their image mean, and two on background whose features match their $7\times7$ neighborhood.
The last two are drawn at random, away from the border and the other queries.
With 12 heads, the five photographs give 480 query--head pairs per layer for the estimates below.\PFILL

\noindent\textbf{Where the mass lies.}
The tile holds 1.6\% of the keys and a median of 6 to 16\% of the mass (\cref{tab:supp_anatomy}a).
The far field, with 98.4\% of the keys, holds a median of 72 to 80\% of the mass at layer 6 and 84 to 90\% at layer 11.
For 74 to 90\% of the query--head pairs, it carries at least half of the mass.
The five global tokens take a median of 3 to 6\% at layer 6 but at most 0.5\% at layer 11.
Per key, attention peaks around the query. A tile key carries a median of 4.3 to 13.5 times the weight of a far key.
The far field still dominates the sum because it has 63 times as many keys as the tile.
Its mass is also spread out, with a median $\neff$ of 46 to 78 of the 1{,}008 far cells at layer 6 and 98 to 171 at layer 11.
The query of \cref{fig:method}(a) is one of these pairs, on the cow at layer 11.
Its tile holds 4\% of the mass, the global tokens 8\% and the far field 87\%, spread over its 1{,}008 cells.\PFILL

\noindent\textbf{Far-field estimates.}
The three selections (the oracle, nearest keys and the larger window) and the \meros sample read about one eighth of the far keys, the budget of \meros $r{=}8$.
We divide each estimate by the true far mass $Z_{\Far_i}$.
The oracle keeps the 2{,}016 far keys with the largest true weights. No selection of this size keeps more mass, but finding it requires the exact attention row.
Nearest keys and the larger window select by position, as in \cref{sec:supp_equal_budget}.
Nearest keys keep the 2{,}272 patches closest to the center of the tile, and the larger window keeps the $48\times48$ square around it.
Only their far keys enter the estimate.
Cell centroids replace each far $4\times4$ cell by its mean key, counted 16 times, as in \cref{prop:bias}(ii).
Since the logit of a mean key is the mean logit, each cell contributes 16 times the geometric mean of its masses, which the stored rows give directly.
For \meros, we use its own sample with seed 17, the first of our three sampling seeds 17, 29 and 43.
Each query reads the 2{,}016 samples outside its tile and weights them by 8.

Selection and centroids never exceed the true far mass, as \cref{prop:bias} states (\cref{tab:supp_anatomy}b).
The oracle misses a median of 5.2\% of the far mass at layer 6 and 15.7\% at layer 11, and cell centroids miss 25.3\% and 17.2\%.
Nearest keys and the larger window keep a median of only 60\% and 57\% at layer 6, and 47\% and 44\% at layer 11.
The ratio of \meros, by contrast, has a mean of 1.00 over all 960 pairs.
Its interquartile range of 0.968 to 1.038 at layer 6 and 0.977 to 1.024 at layer 11 stays within $\pm$4\%.
\meros is closer to the true far mass than the oracle for 54\% of the pairs at layer 6 and 78\% at layer 11.
Against cell centroids, it is closer for 98\% and 99\% of the pairs.
At layer 11, the far field is larger and spread over more cells than at layer 6.
There the oracle misses three times as much mass, while the median distance of the \meros ratio from 1 drops from 3.5\% to 2.3\%.\PFILL

\begin{table*}[!tp]
\centering
\tabsetup{6.9}
\setlength{\tabcolsep}{1.5pt}
\caption{\textbf{Equal key budget} (full version of \cref{tab:analysis}a). Every estimate computes the tile and global tokens exactly and reads about as many keys as \meros $r{=}8$. The Keys column counts keys per patch query of \mbox{DINOv3-B} at 2048\,px. The feature error and cosine (Cos.) compare final patch features with those of exact attention on 12 images of UAVid~\cite{lyu2020uavid}, Cityscapes~\cite{cordts2016cityscapes} and DeepGlobe~\cite{demir2018deepglobe} with a fixed field of view. The two sampled estimates are means of three seeds, whose errors have an SD of at most 0.0024. $\Delta$mIoU gives the worst over UAVid and Cityscapes for each backbone, and the mean and worst over both backbones and datasets, with the protocol of \cref{tab:seg}. Bold marks the best value in each column, and every bold value belongs to \meros.}
\label{tab:supp_equal_budget}
\begin{widewrap}\begin{tabular}{p{\dimexpr\textwidth-353.5pt-36\tabcolsep\relax}>{\centering\arraybackslash}p{18.5pt}*{6}{>{\centering\arraybackslash}p{15.5pt}}*{6}{>{\centering\arraybackslash}p{18.5pt}}*{2}{>{\centering\arraybackslash}p{46pt}}*{2}{>{\centering\arraybackslash}p{24.5pt}}}
\toprule
\rowcolor{tabhead}[\tabcolsep][\rov] & & \multicolumn{6}{c}{\bfseries\boldmath Feature error $\downarrow$} & \multicolumn{6}{c}{\bfseries\boldmath Cos.\ $\uparrow$} & \multicolumn{4}{c}{\bfseries\boldmath $\Delta$mIoU $\uparrow$}\\
\hcmid\cmidrule(lr){3-8}\cmidrule(lr){9-14}\cmidrule(lr){15-18}
\rowcolor{tabhead}[\tabcolsep][\rov] & \bfseries Keys & \multicolumn{3}{c}{\bfseries\boldmath DINOv3-B~\cite{simeoni2025dinov3}} & \multicolumn{3}{c}{\bfseries\boldmath MAE-L~\cite{he2022mae}} & \multicolumn{3}{c}{\bfseries\boldmath DINOv3-B~\cite{simeoni2025dinov3}} & \multicolumn{3}{c}{\bfseries\boldmath MAE-L~\cite{he2022mae}} & \bfseries\boldmath DINOv3-B~\cite{simeoni2025dinov3} & \bfseries\boldmath MAE-L~\cite{he2022mae} & \multicolumn{2}{c}{\bfseries\boldmath Both}\\
\hcmid\cmidrule(lr){3-5}\cmidrule(lr){6-8}\cmidrule(lr){9-11}\cmidrule(lr){12-14}\cmidrule(lr){15-15}\cmidrule(lr){16-16}\cmidrule(lr){17-18}
\rowcolor{tabhead}[\tabcolsep][\rov] \bfseries Far field & \bfseries 2K & \bfseries 1K & \bfseries 2K & \bfseries 4K & \bfseries 1K & \bfseries 2K & \bfseries 4K & \bfseries 1K & \bfseries 2K & \bfseries 4K & \bfseries 1K & \bfseries 2K & \bfseries 4K & \bfseries worst & \bfseries worst & \bfseries mean & \bfseries worst\\
\midrule
\rowcolor{tabsel}[\tabcolsep][\rov] Nearest keys & 2{,}277 & .488 & .617 & .854 & .238 & .253 & .269 & .8746 & .7993 & .6410 & .9714 & .9676 & .9629 & $-$14.66 & $-$18.13 & $-$11.30 & $-$18.13\\
\rowcolor{tabsel}[\tabcolsep][\rov] Larger window & 2{,}309 & .481 & .614 & .853 & .236 & .296 & .269 & .8787 & .8011 & .6417 & .9717 & .9558 & .9630 & $-$14.27 & $-$18.75 & $-$11.33 & $-$18.75\\
Strided subset & 2{,}309 & .418 & .491 & .871 & .198 & .192 & .143 & .9080 & .8766 & .6261 & .9802 & .9814 & .9897 & $-$9.01 & $-$8.45 & $-$5.54 & $-$9.01\\
Unweighted sample & 2{,}277 & .394 & .479 & .874 & .190 & .183 & .133 & .9178 & .8830 & .6246 & .9817 & .9831 & .9911 & $-$7.61 & $-$7.37 & $-$4.77 & $-$7.61\\
\rowcolor{tabsum}[\tabcolsep][\rov] Cell centroids & 2{,}277 & .224 & .321 & .539 & .128 & .105 & .090 & .9735 & .9442 & .8537 & .9917 & .9943 & .9957 & $-$1.36 & $-$0.82 & $-$0.70 & $-$1.36\\
Weighted grid & 2{,}277 & .189 & .181 & .194 & .078 & .043 & .023 & .9818 & .9832 & .9798 & .9969 & .9991 & .9997 & $-$2.11 & $-$1.00 & $-$0.98 & $-$2.11\\
\rowcolor{tabmeros}[\tabcolsep][\rov] \meros & 2{,}277 & \textbf{.156} & \textbf{.155} & \textbf{.186} & \textbf{.058} & \textbf{.033} & \textbf{.017} & \textbf{.9875} & \textbf{.9878} & \textbf{.9812} & \textbf{.9983} & \textbf{.9994} & \textbf{.9999} & $\mathbf{-1.11}$ & $\mathbf{-0.56}$ & $\mathbf{-0.53}$ & $\mathbf{-1.11}$\\
\bottomrule
\end{tabular}\end{widewrap}
\end{table*}

\begin{table*}[!tp]
\centering
\tabsetup{7.0}
\caption{\textbf{Component ablations} on \mbox{DINOv3-B}~\cite{simeoni2025dinov3} (full version of \cref{tab:analysis}b). (a)~One change at a time at $r{=}8$, and full \meros at the other budgets, with $\Delta$mIoU on Cityscapes. (b)~The reflected pairs of \meros against independent positions at all three budgets, both with inclusion probability $1/r$ and weight $r$. The error and the cosine compare the final patch features with those of exact attention, averaged over all tiles. Agree is the share of pixels (\%) that get the label of exact attention. All values are means over the three sampling seeds. Bold marks the best of the variants at $r{=}8$ in (a) and the better of the two samples at each budget in (b).}
\label{tab:supp_components}
\begin{minipage}[t]{0.578\textwidth}
\vspace{0pt}\centering
\setlength{\tabcolsep}{2.2pt}
\begin{widewrap}\begin{tabular}{p{\dimexpr\linewidth-179pt-16\tabcolsep\relax}*{2}{>{\centering\arraybackslash}p{21pt}>{\centering\arraybackslash}p{20pt}>{\centering\arraybackslash}p{30pt}}>{\centering\arraybackslash}p{37pt}}
\toprule
\rowcolor{tabhead}[\tabcolsep][\rov] & \multicolumn{7}{c}{\bfseries\boldmath (a) Components and budgets}\\
\hcmid\cmidrule(lr){2-8}
\rowcolor{tabhead}[\tabcolsep][\rov] & \multicolumn{3}{c}{\bfseries\boldmath UAVid~\cite{lyu2020uavid}} & \multicolumn{4}{c}{\bfseries\boldmath Cityscapes~\cite{cordts2016cityscapes}}\\
\hcmid\cmidrule(lr){2-4}\cmidrule(lr){5-8}
\rowcolor{tabhead}[\tabcolsep][\rov] \bfseries Variant & \bfseries\boldmath error$\downarrow$ & \bfseries\boldmath cos$\uparrow$ & \bfseries\boldmath Agree$\uparrow$ & \bfseries\boldmath error$\downarrow$ & \bfseries\boldmath cos$\uparrow$ & \bfseries\boldmath Agree$\uparrow$ & \bfseries\boldmath $\Delta$mIoU$\uparrow$\\
\midrule
Local tile sampled & .253 & .967 & 95.5 & .411 & .912 & 95.1 & $-$12.64\\
No weight $r$ & .439 & .903 & 91.5 & .476 & .882 & 96.1 & $-$7.61\\
$8{\times}8$ local tile & .171 & .985 & 97.5 & .216 & .976 & 98.4 & $-$1.91\\
Fixed grid & .169 & .985 & 97.2 & .201 & .979 & 98.3 & $-$2.11\\
\rowcolor{tabmeros}[\tabcolsep][\rov] Full \meros, $r{=}8$ & \textbf{.141} & \textbf{.990} & \textbf{98.0} & \textbf{.162} & \textbf{.987} & \textbf{98.9} & $\mathbf{-1.11}$\\
\midrule
\rowcolor{tabmeros}[\tabcolsep][\rov] Full \meros, $r{=}4$ & .094 & .995 & 98.7 & .109 & .994 & 99.3 & $-$0.45\\
\rowcolor{tabmeros}[\tabcolsep][\rov] Full \meros, $r{=}2$ & .055 & .998 & 99.3 & .065 & .998 & 99.6 & $-$0.16\\
\bottomrule
\end{tabular}\end{widewrap}
\end{minipage}\hfill
\begin{minipage}[t]{0.402\textwidth}
\vspace{0pt}\centering
\setlength{\tabcolsep}{2.2pt}
\begin{widewrap}\begin{tabular}{p{\dimexpr\linewidth-120pt-14\tabcolsep\relax}*{6}{>{\centering\arraybackslash}p{20pt}}}
\toprule
\rowcolor{tabhead}[\tabcolsep][\rov] & \multicolumn{6}{c}{\bfseries\boldmath (b) Pairs against independent positions}\\
\hcmid\cmidrule(lr){2-7}
\rowcolor{tabhead}[\tabcolsep][\rov] & \multicolumn{3}{c}{\bfseries\boldmath UAVid~\cite{lyu2020uavid}} & \multicolumn{3}{c}{\bfseries\boldmath Cityscapes~\cite{cordts2016cityscapes}}\\
\hcmid\cmidrule(lr){2-4}\cmidrule(lr){5-7}
\rowcolor{tabhead}[\tabcolsep][\rov] \bfseries Sample & \bfseries\boldmath $r{=}2$ & \bfseries\boldmath $r{=}4$ & \bfseries\boldmath $r{=}8$ & \bfseries\boldmath $r{=}2$ & \bfseries\boldmath $r{=}4$ & \bfseries\boldmath $r{=}8$\\
\midrule
\multicolumn{7}{l}{\textit{Error $\downarrow$}}\\
Independent & .0578 & .0981 & .1470 & .0683 & .1147 & .1714\\
\rowcolor{tabmeros}[\tabcolsep][\rov] Pairs & \textbf{.0554} & \textbf{.0939} & \textbf{.1411} & \textbf{.0647} & \textbf{.1087} & \textbf{.1618}\\
\midrule
\multicolumn{7}{l}{\textit{Agree $\uparrow$}}\\
Independent & 99.27 & 98.71 & 97.98 & 99.55 & 99.24 & 98.82\\
\rowcolor{tabmeros}[\tabcolsep][\rov] Pairs & \textbf{99.30} & \textbf{98.74} & \textbf{98.03} & \textbf{99.57} & \textbf{99.27} & \textbf{98.88}\\
\bottomrule
\end{tabular}\end{widewrap}
\end{minipage}
\end{table*}

\subsection{Checking the Theory on Real Attention}
\label{sec:supp_check}

The results of \cref{sec:supp_proofs} hold for every input.
We now compare them with measurements on the attention of pretrained ViTs in 20 settings of model, layer and size, and also check the spread condition of \cref{cor:resolution}.

\noindent\textbf{Setup.}
We take the queries, keys and values of \mbox{DINOv3}-B (layers 6 and 11), MAE-L (layers 12 and 23) and DINOv2-L with registers~\cite{darcet2024registers} (layers 12 and 23).
The inputs are four DIV8K images whose central square is resized to each size, so the field of view stays fixed.
The sizes run from 1024 to 8192\,px for \mbox{DINOv3}-B, from 1024 to 4096\,px for MAE-L and from 1120 to 4256\,px for DINOv2-L.
With two layers per model, this gives 20 settings of model, layer and size.
For 1024 random patch queries per image and all heads, we compute the exact output and every quantity of \cref{sec:supp_notation} in FP64.
We then take 256 independent draws of the sample with our sampler at each $r\in\{2,4,8\}$.
For one draw per setting and budget, this computation matches the output of our operator to a relative $\ell_2$ distance below $2\times10^{-5}$.

\noindent\textbf{Bounds.}
The measured root-mean-square error stays below the bound of \cref{thm:error} for all 1{,}024 queries and all heads of every image and setting (\cref{fig:supp_theory}(a)).
For each model and layer, the ratio of the measured error to the bound changes by less than a factor of 1.4 across input sizes.
Because this ratio varies so little, the measured error follows the scaling of the bound over the tested sizes, including the rate of \cref{cor:resolution}.
Over the 256 draws, the mean of $\hat\bo$ lies at a median distance from $\bo$ of 0.056 to 0.065 times the root-mean-square error.
Monte Carlo noise alone gives about $1/\sqrt{256}=0.0625$, so the noise of the mean accounts for this distance.
This distance stays within the gray band of the noise in \cref{fig:supp_theory}(e) for queries whose error spans four orders of magnitude.
The bias of \meros on these models is therefore below what 256 independent draws can resolve, as expected from the second-order bias bound of \cref{thm:error}.

\noindent\textbf{Exact variance.}
The formula of \cref{sec:supp_variance} matches the measurement (\cref{fig:supp_theory}(b)).
In each setting, the median ratio of the measured mean of $\|\bU\|^2$ to its exact value is between 0.99 and 1.01.
For $\delta^2$, it is between 0.98 and 1.01.
The mean squared error agrees with $\E\|\bU\|^2$ to within a median of 2.5\%.
The first-order term $\bU$ of the error identity thus accounts for nearly all the error.
Since this variance is proportional to $r-1$, the error of every query should grow by $\sqrt7$ from $r{=}2$ to $r{=}8$.
The measured median ratio is 2.64, with 80\% of the queries between 2.56 and 2.70 (\cref{fig:supp_theory}(f)).

\noindent\textbf{Reflected pairs.}
\Cref{prop:pairs} predicts that pairs reduce the error where $V^->\frac87V^+$.
This condition holds in 75 to 91\% of the cells, depending on the setting.
We compare pairs with a simple random sample that keeps the same number of keys with the same weight $r$.
The median ratio of their mean squared errors is 0.79 to 0.91 per setting.
Pairs give the lower error for 86 to over 99\% of the queries (\cref{fig:supp_theory}(c)).\PFILL

\noindent\textbf{Spread of the far mass.}
\Cref{cor:resolution} needs the far mass to spread over more cells as the resolution grows.
At a fixed field of view, a larger input moves more of each query's mass into the far field (\cref{fig:supp_theory}(g)).
At the same time, the median $\neff$ grows nearly in proportion to the number of far cells (\cref{fig:supp_theory}(d)).
Across the settings, the number of far cells runs from 240 to 16{,}368.
From 2048\,px to the largest size, each doubling of the side multiplies the root-mean-square error by 0.53 on average over models, layers, budgets and doublings (\cref{fig:supp_theory}(h)).
This factor is close to the $1/2$ of \cref{cor:resolution}, as for the operator error of one pretrained \mbox{DINOv3-B} layer in \cref{sec:exp_speed}.\PFILL

\subsection{Full Results at an Equal Key Budget}
\label{sec:supp_equal_budget}

\Cref{tab:analysis}a runs six other far-field estimates at the key budget of \meros $r{=}8$.
\Cref{tab:supp_equal_budget} adds the cosine similarity to the features of exact attention and the worst mIoU loss of each backbone on UAVid and Cityscapes.

\noindent\textbf{Budget.}
Every estimate computes the tile and the global tokens exactly, so all differences come from the far field.
A patch query of \meros $r{=}8$ reads $n_{\mathrm{p}}=256+(N-256)/8$ patch keys and the $|\Glob|$ global tokens, 2{,}277 keys for \mbox{DINOv3-B} at 2048\,px.
The other estimates read the same number of keys, with two small exceptions.
The larger window rounds its side $\sqrt{n_{\mathrm{p}}}$ to an integer, and the strided subset reads 32 more keys than the budget of \meros $r{=}8$.

\noindent\textbf{Estimates.}
Nearest keys and the larger window select by position, as windowed attention does~\cite{hassani2026gnafindings}, and give every key weight 1.
With nearest keys, each $16\times16$ tile shares the $n_{\mathrm{p}}$ patches closest to its center, with ties broken by position.
The larger window is one exact square around the tile, of side $\sqrt{n_{\mathrm{p}}}$ rounded to 48 patches at 2048\,px.
At the borders, the square is shifted inside the image.
The strided subset adds every eighth patch of the grid in raster order with weight 1, a fixed subsample that resembles the key subsampling of ToDo~\cite{smith2024todo}.
The unweighted sample reads the keys of \meros but gives them weight 1 instead of $r$, like the variant ``No weight $r$'' of \cref{sec:supp_components}.
Cell centroids here split each far $4\times4$ cell into two $2\times4$ halves, unlike those of \cref{sec:supp_anatomy}, which merge a whole cell.
Each half is replaced by its mean key and mean value, counted 8 times through a logit offset of $\log 8$, as in proportional attention~\cite{bolya2023tome}.
This gives $N/8$ centroids, the budget of $r{=}8$.
The centroids of the query's own tile are masked.
The weighted grid keeps a fixed lattice of every second row and every fourth column, two positions per cell, with weight 8.
It thus reads as many keys as \meros, which keeps one reflected pair per cell with the same weight.
Its fixed positions, however, give no key the inclusion probability $1/r$ that \cref{lem:unbiased} needs.\PFILL

\noindent\textbf{Results.}
\meros has the lowest error and the highest cosine at every input size on both backbones (\cref{tab:supp_equal_budget}), and the smallest worst loss on each backbone.
The gap to cell centroids widens with resolution, as their error rises from 1.4 to 2.9 times that of \meros on \mbox{DINOv3-B} and from 2.2 to 5.3 times on MAE-L.
Summarization errs in the same direction in every cell (\cref{prop:bias}), so its error does not average out as cells are added.
Without the weight $r$, the same sample has 2.5 to 7.8 times the error of \meros, a factor that grows with resolution on both backbones.
The weighted grid comes second in feature error.
Its worst loss is nevertheless 1.9 and 1.8 times that of \meros, with 2.11 against 1.11~mIoU on \mbox{DINOv3-B} and 1.00 against 0.56 on MAE-L.
Nearest keys and the larger window lose the most, 11.3~mIoU on average over both backbones and datasets.\PFILL

\subsection{Full Results of the Component Ablations}
\label{sec:supp_components}

In \cref{tab:analysis}b, we change one component of \meros at a time on \mbox{DINOv3-B} at $r{=}8$.
\Cref{tab:supp_components}a adds full \meros at the other two budgets, and \cref{tab:supp_components}b compares reflected pairs with independent positions at all three.

\noindent\textbf{Variants.}
Sampling the local tile drops its exact computation.
Each patch query then reads the global tokens and all $N/8$ samples with weight 8, including those in its own tile.
This also lowers the floor of \cref{prop:anchor} from $\pex$ to the mass share of the global tokens.
Dropping the weight $r$ gives the unweighted sample of \cref{sec:supp_equal_budget}, which keeps the sample of \meros with weight 1.
The $8{\times}8$ local tile computes exactly the aligned $8{\times}8$ block that contains the query and samples every cell outside it.
The fixed grid is the weighted grid of \cref{sec:supp_equal_budget}.
Independent positions draw $16/r$ of the 16 positions of each cell uniformly without replacement, with weight $r$.
They keep (P1) and (P2) and drop only the reflection.
A last variant, not in \cref{tab:supp_components}, lets the global queries attend only to the sample of \meros.

\noindent\textbf{Results.}
Full \meros has the lowest error, the highest cosine and the highest agreement among the $r{=}8$ variants (\cref{tab:supp_components}a).
Across budgets, its feature error grows by 1.7 times from $r{=}2$ to $r{=}4$ and by 2.5 times from $r{=}2$ to $r{=}8$.
These factors are close to $\sqrt{3}\approx1.73$ and $\sqrt{7}\approx2.65$.
\Cref{sec:exp_speed} reports this growth with $\sqrt{r-1}$ for the operator error.
Reflected pairs give a 4 to 6\% lower feature error than independent positions at every budget on both datasets (\cref{tab:supp_components}b).
This result matches \cref{sec:supp_check}, where pairs give the lower operator error for 86 to over 99\% of the queries.
Letting the global queries attend only to the sample of \meros raises the feature error from .141 to .148 on UAVid and from .162 to .169 on Cityscapes.
Computing them exactly costs 0.2\% of the logits at 2048\,px (\cref{sec:supp_cost}).

\section{Hardware and Timing}
\label{sec:supp_protocol}

\Cref{sec:supp_timing} describes the protocols of the two timing benchmarks behind our speed and memory measurements.

\subsection{Timing Protocol}
\label{sec:supp_timing}

\noindent\textbf{Operator benchmark.}
The operator benchmark times one attention layer, from the queries, keys and values that the layer computes for a real image to the attention output before the output projection.
Each method, variant and image size runs in its own process.
Each of three rounds makes 20 warm-up calls up to 2048\,px, 10 at 3072 and 4096\,px, and 3 above.
It then times at least 50 calls and 2\,s, or at least 5 calls and 2\,s from 6144\,px on.
CUDA events enclose each call, followed by a synchronization.
The latency of the layer is the median over all timed calls of the three rounds.
The vanilla attention of \mainref{fig:teaser}(c) is a PyTorch attention that forms the full score matrix and runs out of memory at 4K and 8K.\PUP

\noindent\textbf{Backbone benchmark.}
The backbone benchmark times a whole forward pass on a random BF16 image at batch~1.
Each model and input size runs in its own process, which times all methods in three rounds.
Every round builds each method afresh, and the order of the methods changes from round to round.
A method makes 10 warm-up calls for image sides up to 2560\,px, 5 below 6144\,px and 2 from 6144\,px on.
It then runs at least 20 calls and 2\,s, or 3 calls from 6144\,px on, each timed by CUDA events.
The latency is again the median over all calls of the three rounds.
The backbone timings of \meros in \cref{tab:tf,tab:tome,tab:seg} and in \cref{tab:supp_backbones} all come from this benchmark.

\noindent\textbf{Hopper kernels.}
In the operator benchmark, SpargeAttn and ClusterAttn run their official Hopper kernels, built for sm\_90a.
Their operator speedups are therefore timed against FA2 on the H200.
For timing, ClusterAttn runs the authors' batched eigensolver inside the CUDA graphs of its official code.
The backbone speedups of ClusterAttn in \mainref{tab:tf} and \mainref{tab:seg} are also timed against FA2 on the H200.
The same holds for the 8K backbone speedup and memory of SpargeAttn in \mainref{tab:tf}.
For matting, timed on the B300, ClusterAttn runs our Blackwell port of its Hopper-only kernel.

\noindent\textbf{\meros timing.}
\meros selects its kernel path once per model, input size and budget (\cref{sec:supp_kernel}) and executes as a captured CUDA graph covering the layer in the operator benchmark and the full forward pass in the backbone benchmark. As with kernel compilation, the graph is built once per input shape and reused across all calls of that shape. Subsequent calls skip path selection and graph construction, replaying the recorded kernels in a single fast launch.

\noindent\textbf{Task timings.}
Depth, VLM and matting timings instead follow the per-image protocols of \cref{sec:supp_depth,sec:supp_vlm,sec:supp_matting}.

\end{document}